\RequirePackage{snapshot}
\documentclass[10pt,twocolumn,letterpaper]{article}

\usepackage{cvpr}
\usepackage{times}
\usepackage{epsfig}
\usepackage{graphicx}
\usepackage{amsmath}
\usepackage{amssymb}

\usepackage[numbers]{natbib}
\usepackage{wrapfig}
\usepackage{multirow}
\usepackage{algpseudocode}
\usepackage{algorithm}
\usepackage{lipsum}
\usepackage{subfig}
\usepackage{hhline}
\usepackage{paralist}
\usepackage{array}\usepackage{setspace}

\clearpage{}\newcommand{\numimg}{M}
\newcommand{\numsam}{N}

\DeclareMathOperator*{\argmax}{arg\,max}
\DeclareMathOperator*{\argmin}{arg\,min}

\newcommand{\cutabstractup}{\vspace*{-0.3in}}
\newcommand{\cutsectionup}{\vspace*{-0.1in}}\newcommand{\cutsectiondown}{\vspace*{-0.07in}}

\newcommand{\cutsubsectionup}{\vspace*{-0.05in}} \newcommand{\cutsubsectiondown}{\vspace*{-0.05in}} 
  
\newcommand{\cutcaptionup}{\vspace*{-0.1in}}\newcommand{\cutcaptiondown}{\vspace*{-0.1in}}

\makeatletter
\newcommand{\thickhline}{    \noalign {\ifnum 0=`}\fi \hrule height 1pt
    \futurelet \reserved@a \@xhline
}
\makeatother

\makeatletter
\ifcase \@ptsize \relax  \newcommand{\miniscule}{\@setfontsize\miniscule{5}{6}}\fi
\makeatother
\clearpage{}

\usepackage[pagebackref=true,breaklinks=true,letterpaper=true,colorlinks,bookmarks=false]{hyperref}

\usepackage{nameref}
\usepackage{zref-xr}
\zxrsetup{toltxlabel}
\zexternaldocument*{supplementary}

\usepackage[toc,page,header]{appendix}
\usepackage[undotted]{minitoc}
\usepackage{tocvsec2}

\usepackage{setspace}

\cvprfinalcopy 
\def\cvprPaperID{1698}

\usepackage{color}
\definecolor{gray}{rgb}{0.5,0.5,0.5}

\iftrue  
	        \newcommand{\todo}[1]{}
        \newcommand{\outline}[1]{}
        \newcommand{\textgray}[1]{}
        \newcommand{\commenttext}[1]{}
        \newcommand{\commentfoot}[1]{}
        \newcommand{\commentselfoot}[2]{}
        \newcommand{\commentselrep}[2]{}
        \newcommand{\topic}[1]{}
\else 
	        \newcommand{\todo}[1]{{\textcolor{red}{[[TODO: {#1}]]}}}
        \newcommand{\outline}[1]{{\textcolor{blue}{[[{#1}]]}}}
        \newcommand{\textgray}[1]{\textcolor{gray}{[[{#1}]]}}
        \newcommand{\commenttext}[1]{\textcolor{red}{[[{#1}]]}}
        \newcommand{\commentfoot}[1]{\footnote{\textcolor{red}{\textit{#1}}}}
        \newcommand{\commentselfoot}[2]{{\textcolor{blue}{#1}}\commenttext{#2}}
        \newcommand{\commentselrep}[2] {{\textcolor{blue}{#1}} {\textcolor{green}{[[\textit{#2}]]}}}
        \newcommand{\topic}[1]{\textcolor{gray}{\textbf{(#1.)}}}
\fi

\clearpage{}\newcommand{\supprefp}[1]{{Appendix~#1}}
\newcommand{\suppreft}[1]{{Appendix~#1}}
\newcommand{\suppref}[1]{{Appendix~#1}}
\newcommand{\suppname}[0]{the appendices}

\clearpage{}

\begin{document}

\doparttoc[n] \faketableofcontents 
\title{Improving Object Detection with Deep Convolutional Networks\\ via Bayesian Optimization and Structured Prediction}

\author{\smallskip{}Yuting Zhang\textsuperscript{*$\dagger$}, Kihyuk Sohn\textsuperscript{$\dagger$}, Ruben Villegas\textsuperscript{$\dagger$}, Gang Pan\textsuperscript{*}, Honglak Lee\textsuperscript{$\dagger$}\\
\and 
\normalsize \textsuperscript{*} Department of Computer Science, \\
\normalsize  Zhejiang University, Hangzhou, Zhejiang, China, \\
{\tt\small \{zyt,gpan\}@zju.edu.cn}
\and
\normalsize \textsuperscript{$\dagger$} Department of Electrical Engineering and Computer Science, \\
\normalsize University of Michigan, Ann Arbor, MI, USA, \\
{\tt\small \{yutingzh,kihyuks,rubville,honglak\}@umich.edu}
}
 
\maketitle

\cutabstractup
\begin{abstract}
\vspace*{-0.1in}
Object detection systems based on the deep convolutional neural network (CNN) have recently made ground-breaking advances on several object detection benchmarks.
While the features learned by these high-capacity neural networks are discriminative for categorization, inaccurate localization is still a major source of error for detection. 
Building upon high-capacity CNN architectures, we address the localization problem by 1) using a search algorithm based on Bayesian optimization that sequentially proposes candidate regions for an object bounding box, and 2) training the CNN with a structured loss that explicitly penalizes the localization inaccuracy. 
In experiments, we demonstrate that each of the proposed methods improves the detection performance over the baseline method on PASCAL VOC 2007 and 2012 datasets.
Furthermore, two methods are complementary and significantly outperform the previous state-of-the-art when combined.
\end{abstract}
\vspace*{-0.1in}
 \cutsectionup
\section{Introduction}
\label{sec-intro}
\cutsectiondown

Object detection is one of the long-standing and important problems in computer vision.
Motivated by the recent success of deep learning~\cite{lecun1989backpropagation,hinton:science,bengio2007greedy,boureau2008sparse,lee2011unsupervised,bengio:replearn,schmidhuber} on visual object recognition tasks~\cite{alex-imagenet,overfeat,zeiler2014visualizing,vggnet,szegedy2014going}, significant improvements have been made in the object detection problem ~\cite{dnn-det-mask,dnn-scalable-cvpr2014,he2014spatial}.
Most notably, \citet{rcnn-cvpr} proposed the ``regions with convolutional neural network'' (R-CNN) framework for object detection and demonstrated state-of-the-art performance on standard detection benchmarks (e.g., PASCAL VOC~\cite{pascal-voc-2007,pascal-voc-2010}, ILSVRC~\cite{ILSVRCarxiv14}) with a large margin over the previous arts, which are mostly based on deformable part model (DPM)~\cite{dpm-pami}.

There are two major keys to the success of the R-CNN.
First, features matter~\citep{rcnn-cvpr}.
In the R-CNN, the low-level image features (e.g., HOG~\cite{hog}) are replaced with the CNN features, which are arguably more discriminative representations. One drawback of CNN features, however, is that they are expensive to compute.
The R-CNN overcomes this issue by proposing a few hundreds or thousands candidate bounding boxes via the selective search algorithm~\citep{selective-search} to effectively reduce the computational cost required to evaluate the detection scores at all regions of an image.

\commenttext{RUBEN: removed "an" from "caused by an inaccurate localization", "caused by inaccurate localization" is enough}
Despite the success of R-CNN, it has been pointed out through an error analysis~\citep{voc-error} that inaccurate localization causes the most egregious errors in the R-CNN framework~\cite{rcnn-cvpr}.
For example, if there is no bounding box in the close proximity of ground truth among those proposed by selective search, no matter what we have for the features or classifiers, there is no way to detect the correct bounding box of the object.
Indeed, there are many applications that require accurate localization of an object bounding box, such as detecting moving objects (e.g., car, pedestrian, bicycles) for autonomous driving~\cite{KITTI}, detecting objects for robotic grasping or manipulation in robotic surgery or manufacturing~\cite{rss2013deepgrasp}, and many others.

In this work, we address the localization difficulty of the R-CNN detection framework with two ideas.
First, we develop a fine-grained search algorithm to expand an initial set of bounding boxes by proposing new bounding boxes with scores that are likely to be higher than the initial ones.
By doing so, even if the initial region proposals were poor, the algorithm can find a region that is getting closer to the ground truth after a few iterations. We build our algorithm in the Bayesian optimization framework~\cite{mockus1978application,snoek2012practical}, where evaluation of the complex detection function is replaced with queries from a probabilistic distribution of the function values defined with a computationally efficient surrogate model. 
Second, we train a CNN classifier with a structured SVM objective that aims at classification and localization simultaneously. \commenttext{RUBEN: this entire sentence does not sound right so I changed it. Here is the original one if you guys want it back:
We define the structured SVM layer of CNN whose objective function is defined with a hinge loss that balances between classification (i.e., determines whether an object exists) and localization (i.e., determines how much it overlaps with the ground truth), and replace the last layer of the CNN, which is usually a softmax or linear SVM layer, with structured SVM layer.
}
We define the structured SVM objective function with a hinge loss that balances between classification (i.e., determines whether an object exists) and localization (i.e., determines how much it overlaps with the ground truth) to be used as the last layer of the CNN. 

In experiments, we evaluated our methods on PASCAL VOC 2007 and 2012 detection tasks and compared to other competing methods.
We demonstrated significantly improved performance over the state-of-the-art at different levels of intersection over union (IoU) criteria. 
In particular, our proposed method outperforms the previous arts with a large margin at higher IoU criteria (e.g., IoU = $0.7$), which highlights the good localization ability of our method.

\commenttext{Ruben: changed "In sum" to "overall"}
Overall, the contributions of this paper are as follows: 1)~we develop a Bayesian optimization framework that can find more accurate object bounding boxes without significantly increasing the number of bounding box proposals, 2)~we develop a structured SVM framework to train a CNN classifier for accurate localization, 3)~the aforementioned methods are complementary and can be easily adopted to various CNN models, and finally, 4)~we demonstrate significant improvement in detection performance over the R-CNN on both PASCAL VOC 2007 and 2012 benchmarks.
 \vspace*{-0.1in}
\cutsectionup
\section{Related work}
\label{sec-related}
\cutsectiondown
The DPM~\cite{dpm-pami} and its variants~\cite{oc-dpm,seg-dpm} have been the dominating methods for object detection tasks for years.
These methods use image descriptors such as HOG~\cite{hog}, SIFT~\cite{lowe2004distinctive}, and LBP~\cite{lbp} as features and densely sweep through the entire image to find a maximum response region.
With the notable success of CNN on large scale object recognition~\cite{alex-imagenet}, several detection methods based on CNNs have been proposed~\citep{overfeat,sermanet2013pedestrian,dnn-det-mask,dnn-scalable-cvpr2014,rcnn-cvpr}. Following the traditional sliding window method for region proposal, \citet{overfeat} proposed to search exhaustively over an entire image using CNNs, but made it efficient by conducting a convolution on the entire image at once at multiple scales. Apart from the sliding window method, \citet{dnn-det-mask} used CNNs to regress the bounding boxes of objects in the image and used another CNN classifier to verify whether the predicted boxes contain objects.
\citet{rcnn-cvpr} proposed the R-CNN following the ``recognition using regions'' paradigm~\cite{rec-with-region}, which also inspired several previous state-of-the-art methods~\cite{selective-search,regionlets}.
In this framework, a few hundreds or thousands of regions are proposed for an image via the selective search algorithm~\citep{selective-search} and the CNN is finetuned with these region proposals.
Our method is built upon the R-CNN framework using the CNN proposed in~\citep{vggnet}, but with 1) a novel method to propose extra bounding boxes in the case of poor localization, and 2) a classifier with improved localization sensitivity.

The structured SVM objective function in our work is inspired by \citet{struct-localization}, where they trained a kernelized structured SVM on low-level visual features (i.e., HoG~\citep{hog}) to predict the object location. 
\citet{struct-cnn} integrated a structured objective with the deep neural network for object detection, but they adopted the branch-and-bound strategy for training as in \citep{struct-localization}.
In our work, we formulate the linear structured objective upon high-level features learned by deep CNN architectures, but our negative mining step is very efficient thanks to the region-based detection framework. 
We also present a gradient-based optimization method for training our architecture.

\commenttext{Ruben: Changed "works" for "work"}
There have been several other related work for accurate object localization.
\citet{seg-dpm} incorporated the geometric consistency of bounding boxes with bottom-up segmentation as auxiliary features into the DPM. 
\citet{localize-detected} used the structured SVM with color and edge features to refine the bounding box coordinates in DPM framework.
\citet{accurate-cvpr2014} used the height prior of an object.
These auxiliary features to aid object localization can be injected into our framework without modifications.

Localization refinement can be also taken as a CNN regression problem. 
\citet{rcnn-cvpr} extracted the middle layer features and linearly regressed the initially proposed regions to better locations. 
\citet{overfeat} refined bounding boxes from a grid layout to flexible locations and sizes using the higher layers of the deep CNN architecture. 
\citet{dnn-scalable-cvpr2014} jointly conducted classification and regression in a single architecture. 
Our method is different in that 1) it uses the information from multiple existing regions instead of a single bounding box for predicting a new candidate region, and 2) it focuses only on maximizing the localization ability of the CNN classifier instead of doing any regression from one bounding box to another.
 
\begin{figure*}[t]
\centering
\includegraphics[width=1\textwidth]{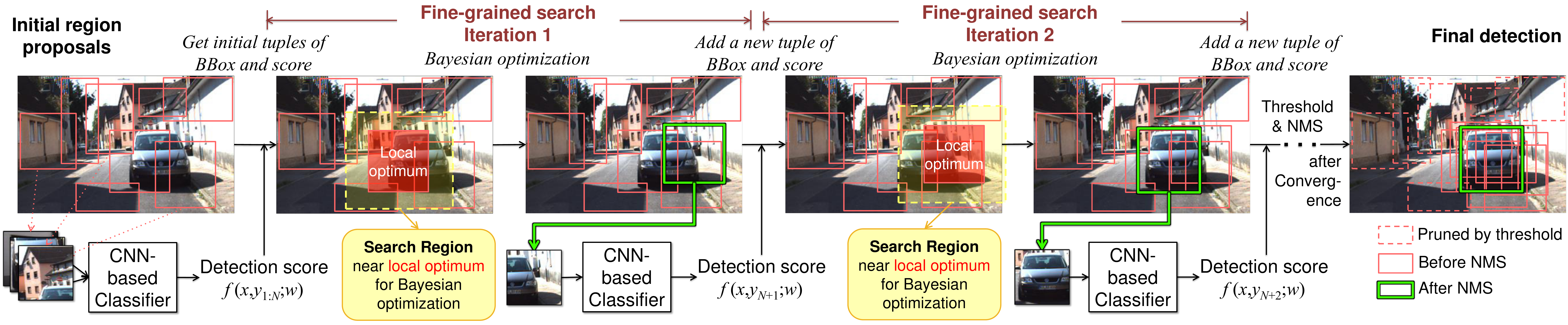}
\cutcaptionup
\vspace{-0.1in}
\caption{Pipeline of our method. 
1) Initial bounding boxes are given by methods such as the selective search~\citep{selective-search} and their detection scores are obtained from the CNN-based classifier trained with structured SVM objective. 2) The box(es) with optimal score(s) in the local regions are found by greedy NMS~\cite{rcnn-cvpr} (shown as ``local optimum'' boxes), and Bayesian optimization takes the neighborhood of each local optimum to propose a new box with a high chance of getting a better
detection score. 3) We evaluate the detection score of the new box, take it as an observation for the next iteration of the Bayesian optimization until convergence. Note that the local optimum and the search region may change in each iteration. 4) All the bounding boxes are fed into the standard post-processing stage (e.g., thresholding and NMS, etc.).}
\label{fig:structured-rcnn} 
\cutcaptiondown
\vspace{-0.1in}
\end{figure*}

\cutsectionup
\section{Fine-grained search for bounding box via Bayesian optimization}
\label{sec-gp}
\cutsectiondown
Let $f(x,y)$ denote a detection score of an image $x$ at the region with the box coordinates $y=(u_{1}, v_{1}, u_{2}, v_{2})\in{\mathcal Y}$. 
The object detection problem deals with finding the local maximum of $f(x,y)$ with respect to $y$ of an unseen image $x$.\footnote{When multiple (including zero) objects exist, it involves finding the local maxima that exceed a certain threshold.}
As it requires an evaluation of the score function at many possible regions, it is crucial to have an efficient algorithm to search for the candidate bounding boxes.

A sliding window method has been used as a dominant search algorithm~\cite{hog,dpm-pami}, which exhaustively searches over an entire image with fixed-sized windows at different scales to find a bounding box with a maximum score.
However, evaluating the score function at all regions determined by the sliding window approach is prohibitively expensive when the CNN features are used as the image region descriptor. The problem becomes more severe when flexible aspect ratios are needed for handling object shape variations. 
Alternatively, the ``recognition using regions''~\cite{rec-with-region,rcnn-cvpr} method has been proposed, which requires to evaluate significantly fewer number of regions (e.g., few hundreds or thousands) with different scales and aspect ratios, and it can use the state-of-the-art image features with high computational complexity, such as the CNN features~\cite{decaf}.
One potential issue of object detection pipelines based on region proposal is that the correct detection will not happen when there is no region proposed in the proximity of the ground truth bounding box.\footnote{We refer to selective search as a representative method for region proposal.}
To resolve this issue, one can propose more bounding boxes to cover the entire image more densely, but this would significantly increase the computational cost. 
In this section, we develop a fine-grained search (FGS) algorithm based on Bayesian optimization that sequentially proposes a new bounding box with a higher \emph{expected} detection score than previously proposed bounding boxes without significantly increasing the number of region proposals.
We first present the general Bayesian optimization framework (Section~\ref{sec-general-bayesian-framework}) and describe the FGS algorithm using Gaussian process as the prior for the score function (Section~\ref{sec-gp-regression}).
We then present the local FGS algorithm that searches over multiple local regions instead of a single global region (Section~\ref{sec-l-FGS}), and discuss the hyperparameter learning of our FGS algorithm (Section~\ref{sec-gp-param}).

\cutsubsectionup
\subsection{General Bayesian optimization framework}
\label{sec-general-bayesian-framework}
\cutsubsectiondown
Let $\{y_{1},\cdots,y_{\numsam}\}$ be the set of solutions (e.g., bounding boxes).
In the Bayesian optimization framework, $f = f(x,y)$ is assumed to be drawn from a probabilistic model: \begin{equation}
p\left(f|\mathcal{D}_{\numsam}\right)\propto p\left(\mathcal{D}_{\numsam}|f\right) p(f),
\end{equation}
where $\mathcal{D}_{\numsam}$ $=$ $\{(y_{j},f_{j})\}_{j=1}^{\numsam}$ and $f_{j}$ $=$ $f(x, y_{j})$.
Here, the goal is to find a new solution $y_{\numsam+1}$ that maximizes the chance of improving the detection score $f_{\numsam+1}$, where the chance is often defined as an acquisition function $a(y_{\numsam+1}|\mathcal{D}_{\numsam})$. Then, the algorithm proceeds by recursively sampling a new solution $y_{\numsam+t}$ from $\mathcal{D}_{\numsam+(t-1)}$, and update the set $\mathcal{D}_{\numsam+t} = \mathcal{D}_{\numsam+(t-1)}\cup\{y_{\numsam+t}\}$ to draw a new sample solution $y_{\numsam+(t+1)}$ with an updated observation.

Bayesian optimization is efficient in terms of the number of function evaluation~\cite{jones2001taxonomy}, and is particularly effective when $f$ is computationally expensive.
When $a(y_{\numsam+1}|\mathcal{D}_{\numsam})$ is much less expensive than $f$ to evaluate, and the computation for $\argmax_{y_{\numsam+1}} a(y_{\numsam+1}|\mathcal{D}_{\numsam})$ requires only a few function evaluations, we can efficiently find a solution that is getting closer to the ground truth. 
\commenttext{What do you mean by manageable? Can efficiently be replaced with within few iterations?}

\cutsubsectionup
\subsection{Efficient region proposal via GP regression}
\label{sec-gp-regression}
\cutsubsectiondown
A Gaussian process (GP) defines a prior distribution $p(f)$ over the function $f:\mathcal{Y}\rightarrow\mathbb{R}$. Due to this property, a distribution over $f$ is fully specified by a mean function $m:\mathcal{Y}\rightarrow\mathbb{R}$ and a positive definite covariance kernel $k:\mathcal{Y}\times\mathcal{Y}\rightarrow\mathbb{R}$, i.e., $f$ $\sim$ $\mathcal{GP}(m(\cdot),k(\cdot,\cdot))$. 
Specifically, for a finite set $\{y_{j}\}_{j=1}^{\numsam}\subset\mathcal{Y}$, the random vector $[f_{j}]_{1\leq j \leq \numsam}$ follows a multivariate Gaussian distribution $\mathcal{\numsam}\left([m(y_{j})]_{1\leq j \leq \numsam},[k(y_{i},y_{j})]_{1\leq i,j\leq \numsam}\right)$.
A random Gaussian noise with precision $\beta$ is usually added to each $f_j$ independently in practice. 
Here, we used the constant mean function $m(y)$ $=$ $m_{0}$ and the squared exponential covariance kernel with automatic relevance determination (SEard) as follows: $k_{\text{SEard}}(y_{i},y_{j};z)=$
\begin{equation}
\eta\exp\left(-\frac{1}{2}\left(\Psi_{z}(y_{i})-\Psi_{z}(y_{j})\right)^{\top}\Lambda\left(\Psi_{z}(y_{i})-\Psi_{z}(y_{j})\right)\right),\nonumber\end{equation}
where $\Lambda$ is a $4\times 4$ diagonal matrix whose diagonal entries are $\lambda_{i}^2, i= 1,\cdots,4$. 
These form a $7$-dimensional GP hyperparameter $\theta = (\beta,m_{0},\eta,\lambda_{1}^2,\lambda_{2}^2,\lambda_{3}^2,\lambda_{4}^2)$ to be learned from the training data.
$\Psi_{z}:\mathcal{Y}\rightarrow\mathbb{R}^{4}$ transforms the bounding box coordinates $y$ into a new form:
\begin{equation}
\Psi_{z}(y) = \left[\frac{\bar{u}}{\exp(z)}\; ;\;\frac{\bar{v}}{\exp(z)}\; ;\;\log{w}\; ;\; \log{h} \right],
\end{equation}
where $\bar{u}=\frac{u_{1}+u_{2}}{2}$ and $\bar{v}=\frac{v_{1}+v_{2}}{2}$ denote the center coordinates, $w=u_{2}-u_{1}$ denotes the width, and $h=v_{2}-v_{1}$ denotes the height of a bounding box.
We introduce a latent variable $z$ to make the covariance kernel scale-invariant.\footnote{If the image and the bounding boxes $y_{i}, y_{j}$ are scaled down by a certain factor, we can keep $k_{\text{SEard}}(y_{i},y_{j};z)$ invariant by properly setting $z$.}
We determine $z$ in a data-driven manner by maximizing the marginal likelihood of $\mathcal{D}_{N}$, or 
\begin{equation}
\hat{z}=\argmax_{z}p(\{f_j\}_{j=1}^{N}|\{y_j\}_{j=1}^{N};\theta). \label{eq:gp-latent-max}
\end{equation}

The GP regression (GPR) problem tries to find a new argument $y_{\numsam+1}$ given $\numsam$ observations $\mathcal{D}_{\numsam}$ that maximizes the value of acquisition function, which, in our case, is defined with the expected improvement (EI) as: $a_{EI}(y_{\numsam+1}|\mathcal{D}_{\numsam}) =$
\begin{equation}
 \int_{\hat{f}_{\numsam}}^{\infty}(f-\hat{f}_{\numsam})\cdot p(f|y_{\numsam+1},\mathcal{D}_{\numsam};\theta)df\label{eq:gp-acq}
\end{equation}
where $\hat{f}_{\numsam} = \max_{1\leq j \leq \numsam} f_{j}$.
The posterior of $f_{\numsam + 1}$ given $(y_{\numsam+1},\mathcal{D}_{\numsam})$ follows Gaussian distribution: $p(f_{\numsam+1}|y_{\numsam+1},\mathcal{D}_{\numsam};\theta)=$
\begin{align}
\mathcal{N}\big(f_{\numsam+1};\mu(y_{\numsam+1}|\mathcal{D}_{\numsam}),\sigma^{2}(y_{\numsam+1}|\mathcal{D}_{\numsam})\big),\label{eq:gp-regression}
\end{align}
with the following mean function and covariance kernels:
\begin{align*}
\mu(y_{\numsam+1}|\mathcal{D}_{\numsam}) & = m_{0}+\mathbf{k}_{\numsam+1}^{\top}\mathbf{K}_{\numsam}^{-1}\left(\big[f_{j} - m_{0}\big]_{1\leq j\leq \numsam}\right), \\
\sigma^{2}(y_{\numsam+1}|\mathcal{D}_{\numsam}) & =k_{N+1}-\mathbf{k}_{\numsam+1}^{\top}\mathbf{K}_{\numsam}^{-1}\mathbf{k}_{\numsam+1}, \\
k_{N+1} & =\beta^{-1}+ k(y_{\numsam+1},y_{\numsam+1}),\\
\mathbf{k}_{\numsam+1} & =\big[k(y_{\numsam+1},y_{j})\big]_{1\leq j\leq \numsam},\\
\mathbf{K}_{\numsam} & =\big[k(y_{i},y_{j})\big]_{1\leq i,j\leq \numsam}+\beta^{-1}\mathbf{I}\;.
\end{align*}
We refer~\cite{GPML} for detailed derivation.
By plugging~\eqref{eq:gp-regression} in~\eqref{eq:gp-acq}, \begin{align}
&a_{EI}(y_{\numsam+1}|\mathcal{D}_{\numsam})=\sigma\left(y_{\numsam+1}|\mathcal{D}_{\numsam}\right)\times\nonumber\\
&\big(\gamma(y_{\numsam+1})F(\gamma(y_{\numsam+1}))+\mathcal{N}(\gamma(y_{\numsam+1});0,1)\big)\label{eq:gp-acq-closed}
\end{align}
where $\gamma(y_{\numsam+1})=\frac{\mu(y_{\numsam+1}|\mathcal{D}_{\numsam})-\hat{f}_{\numsam}}{\sigma\left(y_{\numsam+1}|\mathcal{D}_{\numsam}\right)}$. 
$F(\cdot)$ is the cumulative distribution function of standard normal distribution $\mathcal{N}(\cdot)$.

\cutsubsectionup
\subsection{Local fine-grained search}
\label{sec-l-FGS}
\cutsubsectiondown

\begin{algorithm}[t]
\caption{Local fine-grained search (FGS) \label{alg:FGS}}
\begin{algorithmic}[1]
\small {
\Require  Image $x$, classifier $f_{\text{CNN}}$, a set of structured labels and classification scores $\mathcal{D}_{\numsam}$ $=$ $\{(y_{j}, f_{j})_{j=1}^{\numsam}\}$, GP hyperparameter $\theta$, maximum number of GP iterations $t_{\text{max}}$, a threshold $f_{\text{prune}}$ to prune out the bounding boxes, different levels of IoU $\rho_{r}, r=1,\ldots,R$ determining the size of local regions.
\Ensure  A set of structured labels and classification scores $\mathcal{D}$.
\State  $\mathcal{D}\leftarrow\mathcal{D}_{\numsam}$
\For{$t=1,\cdots,t_{\text{max}}$}
\State  $\mathcal{D}_{\text{proposal}} = \varnothing$
\State  $\mathcal{D}_{\text{prune}} = \{(y,f)\in\mathcal{D}:f>f_{\text{prune}}\}$
\State  $\mathcal{D}_{\text{NMS}} = \text{NMS}(\mathcal{D}_{\text{prune}})$
\For{ \textbf{each} $(y_{\text{best}},f_{\text{best}})\in\mathcal{D}_{\text{NMS}}$}
\For{ $r=1,\cdots,R$ }
\State  $\mathcal{D}_{\text{local}} = \{(y,f)\in\mathcal{D}:\operatorname{IoU}(y,y_{\text{best}})>\rho_{r}\}$
\State  $\hat{z} = {\arg\max}_{z}\, p(\mathcal{D}_{\text{local}};\theta)$ \label{alg-ln:latent}
\hfill (Equation~\eqref{eq:gp-latent-max})
\State $\hat{y} = {\arg\max}_{y}\, a_{EI}(y|\mathcal{D}_{\text{local}};\theta,\hat{z})$ \label{alg-ln:best}
\hfill (Equation~\eqref{eq:gp-acq-closed})
\State  $\hat{f} = f_{\text{CNN}}(x,\hat{y})$
\State  $\mathcal{D}_{\text{proposal}}\leftarrow\mathcal{D}_{\text{proposal}}\cup\{(\hat{y},\hat{f})\}$
\EndFor
\EndFor
\State  $\mathcal{D}\leftarrow\mathcal{D}\cup\mathcal{D}_{\text{proposal}}$
\EndFor
}
\end{algorithmic}
\end{algorithm}

In this section, we extend the GPR-based algorithm for global maximum search to local fine-grained search (FGS).  

The local FGS steps are described in Figure~\ref{fig:structured-rcnn}.
We perform the FGS by pruning out easy negatives with low classification scores from the set of regions proposed by the selective search algorithm and sorting out a few bounding boxes with the maximum scores in local regions.
Then, for each local optimum $y_{\text{best}}$ (red boxes in Figure~\ref{fig:structured-rcnn}), we propose a new candidate bounding box (green boxes in Figure~\ref{fig:structured-rcnn}).
Specifically, we initialize a set of local observations $\mathcal{D}_{\text{local}}$ for $y_{\text{best}}$ from the set given by the selective search algorithm, whose localness is measured by an IoU between $y_{\text{best}}$ and region proposals (yellow boxes\footnote{In practice, the local search region associated with $\mathcal{D}_{\text{local}}$ is not a rectangular region around local optimum since we use IoU to determine it.} in Figure~\ref{fig:structured-rcnn}).
$\mathcal{D}_{\text{local}}$ is used to fit a GP model, and the procedure is iterated for each local optimum at different levels of IoU until there is no more acceptable proposal.
We provide a pseudocode of local FGS in Algorithm~\ref{alg:FGS}, where the parameters are set as: $t_{\text{max}}=8$, $(\rho_r)_{r=1}^{R=3}=(0.3,0.5,0.7)$. 

In addition to the capability of handling multiple objects in a single image, 
better computational efficiency is another factor making local FGS preferable to global search. 
As a kernel method, the computational complexity of GPR increases cubically to the number of observations.
By restricting the observation set to the nearby region of a local optimum, the GP fitting and proposal process can be performed efficiently.
In practice, FGS introduces only $<20\%$ computational overhead compared to the original R-CNN.   
Please see \suppname, which are also available in our technical report~\citep{fgs-tech}, for more details on its practical efficiency (\suppref{\ref{sup:FGS-efficiency}}).

\cutsubsectionup
\subsection{Learning GP hyperparameter}
\label{sec-gp-param}
\cutsubsectiondown

As we locally perform the FGS, the GP hyperparameter $\theta$ also needs to be trained with observations in the vicinity of ground truth objects.
To this end, for an annotated object in the training set, we form a set of observations with the structured labels and corresponding classification scores of the bounding boxes that are close to the ground truth bounding box. Such an observation set is composed of the bounding boxes (given by selective search and random selection) whose IoU with the ground truth exceed a certain threshold.
Finally, we fit a GP model by maximizing the joint likelihood of such observations:
\begin{equation*}
\hat{\theta} = \argmax_{\theta}\sum_{i\in I_{\text{pos}}}\log p(\{(y,f):y\in\widetilde{\mathcal{Y}}_{i},f=f(x_i,y)\};\theta),
\end{equation*}
where $I_{\text{pos}}$ is the index set for positive training samples (i.e., with ground truth object annotations), and $y_{i}$ is a ground truth annotation of an image $x_i$.\footnote{We assumed one object per image. See Section~\ref{sec-finetuning} for handling multiple objects in training.} We set $\widetilde{\mathcal{Y}}_{i} = \{y=y_{i}\operatorname{or}y\in\mathcal{Y}_{i}\operatorname{or}y\in\bar{\mathcal{Y}}_{i}: \operatorname{IoU}(y,y_{i}) > \rho\}\}$, where $\mathcal{Y}_i$ consists of the bounding boxes given by selective search on $x_i$, $\bar{\mathcal{Y}}_{i}$ is a random subset of $\mathcal{Y}$, and $\rho$ is the overlap threshold.
\commenttext{Generally, we are abusing $\hat{}$ or $\bar{}$ notations.. Such as $\hat{y}=g(x,w)$ (it's used as a max of two different functions in Sec 3.2.1 and Sec 4) $\bar{y}$, $\hat{\mathcal{Y}}$, $\bar{\mathcal{Y}}$... We need to clean up notations..}
The optimal solution $\hat{\theta}$ can be obtained via L-BFGS. Our implementation relies on the GPML toolbox~\cite{GPML}.

\cutsectionup
\section{Learning R-CNN with structured loss}
\label{sec-structured_loss}
\cutsectiondown

This section describes a training algorithm of R-CNN for object detection using structured loss.
We first revisit the object detection framework with structured output regression introduced by~\citet{struct-localization} in Section~\ref{sec-str_detect}, and extend it to R-CNN pipeline that allows training the network with structured hinge loss in Section~\ref{sec-finetuning}.

\cutsubsectionup
\subsection{Structured output regression for detection}
\label{sec-str_detect}
\cutsubsectiondown

Let $\{x_{1},x_{2},\ldots,x_{\numimg}\}$ be the set of training images and $\{y_{1},y_{2},\ldots,y_{\numimg}\}$ be the set of corresponding structured labels.
The structured label $y\in{\mathcal Y}$ is composed of 5 elements $(l, u_{1}, v_{1}, u_{2}, v_{2})$; when $l=1$, $(u_{1},v_{1})$ and $(u_{2},v_{2})$ denote the top-left and bottom-right coordinates of the object, respectively, and when $l=-1$, it implies that there is no object in $x_i$, and there is no meaning on coordinate elements $(u_{1},v_{1},u_{2},v_{2})$.
Note that the definition of $y_{i}$ is extended from Section~\ref{sec-gp} to indicate the presence of an object $(l)$ as well as its location $(u_{1}, v_{1}, u_{2}, v_{2})$ when exists. When there are multiple objects in an image, we crop an image into multiple positive ($l=1$) images, each of which contains a single object, and a negative image ($l=-1$) that doesn't contain any object.\footnote{We also perform the same procedure for images with a single object during the training.}
Let $\phi(x,y)$ represent the feature extracted from an image $x$ for a label $y$ with $l=1$. 
In our case, $\phi(x,y)$ denotes the top-layer representations of the CNN (excluding the classification layer) at location specified by $y$,\footnote{Following~\cite{rcnn-cvpr}, we crop and warp the image patch of $x$ at location given by $y$ to a fixed size (e.g., 224$\times$224) to compute the CNN features.} which are fed into the classification layer. The detection problem is to find a structured label $y\in{\mathcal Y}$ that has the highest score:
\begin{align}
g(x;w) = \argmax_{y\in\mathcal{Y}} f(x,y;w)\label{eq:detector}
\end{align}
where
\begin{align}
f(x,y;w) & = w^{T}\widetilde{\phi}(x,y),\label{eq:score}\\
\widetilde{\phi}(x,y) & = \begin{cases}
\phi(x,y) & ,\;l=+1\;, \\
\mathbf{0} & ,\;l=-1\;.
\end{cases}\label{eq:feature}
\end{align}
Note that \eqref{eq:feature} includes a trick for setting the detection threshold to $0$. The model parameter $w$ is trained to minimize the structured loss $\Delta(\cdot,\cdot)$ between the predicted label $g(x_{i};w)$ and the ground-truth label $y_{i}$:
\begin{equation}
\hat{w} = \argmin_{w} \sum_{i=1}^{\numimg} \Delta(g(x_{i}; w), y_{i}) \label{eq:param_est}
\end{equation}
For the detection problem, the structured loss $\Delta(y, y_i)$ is defined in terms of intersection over union (IoU) of two bounding boxes defined by $y$ and $y_i$ as follows:\begin{equation}
\Delta(y,y_{i})=\begin{cases}
1-\text{IoU}(y,y_{i}) & ,\;\operatorname{if}l=l_{i}=1,\\
0 & ,\;\operatorname{if}l=l_{i}=-1,\\
1 & ,\;\operatorname{if}l\neq l_{i}
\end{cases}\label{eq:strloss}
\end{equation}
where $\text{IoU}(y,y_i) = \frac{\text{Area}(y\cap y_i)}{\text{Area}(y\cup y_i)}$.
In general, the optimization problem~\eqref{eq:param_est} is difficult to solve due to the complicated form of structured loss.
Instead, we formulate the surrogate objective in structured SVM framework~\citep{tsochantaridis2005large} as follows:
\begin{align}
&\min_{w} \;\; \frac{1}{2}\|w\|^{2}+\frac{C}{\numimg}\sum_{i=1}^{\numimg}\xi_{i}\;\;\;\text{, subject to} \label{eq:struct-svm}\\ & \;\; w^{\top}\widetilde{\phi}(x_{i},y_{i})\geq w^{\top}\widetilde{\phi}(x_{i},y)+\Delta(y,y_{i})-\xi_{i},\forall y,\forall i \label{eq:stsvm_const}\\
 & \;\; \xi_{i}\geq\;0,\,\forall i
\end{align}
Using~\eqref{eq:feature} and~\eqref{eq:strloss}, the constraint~\eqref{eq:stsvm_const} is written as follows:
\begin{align}
w^{\top}{\phi}(x_{i},y_{i}) \geq &\; w^{\top}{\phi}(x_{i},y)+\Delta^{\text{loc}}(y,y_{i})-\xi_{i},\label{eq:stsvm_const_reform_pos_pos}\\
& \; \forall y\in{\mathcal Y}^{(l=1)}, \forall i \in I_{\text{pos}}\nonumber,\\
w^{\top}{\phi}(x_{i},y_{i}) \geq &\; 1-\xi_{i},\;\forall i \in I_{\text{pos}},\label{eq:stsvm_const_reform_pos_neg}\\
w^{\top}{\phi}(x_{i},y) \leq &\; -1 + \xi_{i}, \;\forall y\in{\mathcal Y}^{(l=1)}, \forall i \in I_{\text{neg}},\label{eq:stsvm_const_reform_neg_pos}
\end{align}
where $\Delta^{\text{loc}}(y,y_{i}) = 1-\text{IoU}(y,y_{i})$, $I_{\text{pos}}$ and $I_{\text{neg}}$ denote the set of indices for positive and negative training examples, respectively, and $\mathcal{Y}^{(l=1)} = \{y\in\mathcal{Y}:l=1\}$.

\cutsubsectionup
\subsection{Gradient-based learning of R-CNN with structured SVM objective}
\label{sec-finetuning}
\cutsubsectiondown
To learn the R-CNN detector with structured loss, we propose to make several modifications to the original structured SVM formulation.
First, we restrict the output space ${\mathcal Y}_{i}\subset{\mathcal Y}$ of $i$th example to regions proposed via selective search.
This results in a change in notation for every ${\mathcal Y}$ in~\eqref{eq:stsvm_const_reform_pos_pos} and~\eqref{eq:stsvm_const_reform_neg_pos} of $i$th example to ${\mathcal Y}_i$.
Second, the constraints (\ref{eq:stsvm_const_reform_pos_pos},~\ref{eq:stsvm_const_reform_pos_neg},~\ref{eq:stsvm_const_reform_neg_pos}) should be transformed into hinge loss to backpropagate the gradient to lower layers of CNN.
Specifically, the objective function~\eqref{eq:struct-svm} is reformulated as follows:
\begin{equation}
\min_{w}\;\; \frac{1}{2}\|w\|^{2}+\frac{1}{\numimg}\big(C_{1}\sum_{i\in I_{\text{pos}}}h_{\text{pos},i} + C_{2}\sum_{i\in I_{\text{neg}}} h_{\text{neg},i}\big)\label{eq:struct-svm-hinge}
\end{equation}
where $h_{\text{pos},i} = h_{\text{pos},i}(w)$, $h_{\text{neg},i} = h_{\text{neg},i}(w)$ are given as:
\begin{align}
&h_{\text{pos},i}(w) = \max\Big\{ 0, 1-w^{\top}\phi(x_{i},y_{i}), \label{eq:hinge-pos}\\
&\quad\max_{y\in\mathcal{Y}_{i}^{(l=1)}}\left(w^{\top}\left(\phi(x_{i},y) - \phi(x_{i},y_{i})\right) + \Delta^{\text{loc}}(y,y_{i})\right)\Big\} \nonumber
\end{align}
\begin{align}
&h_{\text{neg},i}(w) = \max\Big\{ 0, \max_{y\in\mathcal{Y}_{i}^{(l=1)}}(1+w^{\top}\phi(x_{i},y))\Big\} \label{eq:hinge-neg}
\end{align}
Note that we use different $C$ values for positive and negative examples. In experiments, $C_1=2$ and $C_2=1$.

Structured SVM objective may cause a slow convergence in parameter estimation since it utilizes at most one instance $y$ among a large number of instances in the (restricted) output space $\mathcal{Y}_i$, whose size varies from few hundreds to thousands.
To overcome this issue, we alternately perform a gradient-based parameter estimation and hard negative data mining that effectively adapts the number of training examples to be evaluated for updating the parameters (\suppreft{\ref{sup:hard-mining}}). 

\commenttext{Ruben: I don't think realize is the correct word here (realize means to become fully aware of (something) as a fact; understand clearly and also to cause (something desired or anticipated) to happen, but the latter is usually used when a dream happens and not for a process like a learning process). I changed it to perform (perform means to carry out, accomplish, or fulfill)}
For model parameter estimation, we use L-BFGS to first learn parameters of the classification layer only.
We found that this already resulted in a good detection performance. 
Then, we optionally use stochastic gradient descent to finetune the whole CNN classifiers (\suppreft{\ref{sup:finetuning}}).

\cutsectionup
\section{Experimental results}
\label{sec-exp}
\cutsectiondown

We applied our proposed methods to standard visual object detection tasks on PASCAL VOC 2007~\cite{pascal-voc-2007} and 2012~\cite{pascal-voc-2012}.
In all experiments, we consider R-CNNs~\citep{rcnn-cvpr} as baseline models. 
Following \citep{rcnn-cvpr}, we used the CNN models pretrained on ImageNet database~\citep{imagenet_cvpr09} with $1,000$ object categories~\citep{alex-imagenet,vggnet}, and finetuned the whole network using the target database by replacing the existing softmax classification layer to a new one with a different number of classes (e.g., $20$ classes for VOC 2007 and 2012). 
We provide the learning details in \supprefp{\ref{sup:learning-details}}. 
Our implementation is based on the Caffe toolbox \citep{caffe}.

Setting the R-CNN as a baseline method, we compared the detection performance of our proposed methods, such as R-CNN with FGS (R-CNN + FGS), R-CNN trained with structured SVM objective (R-CNN + StructObj), and their combination (R-CNN + StructObj + FGS). 
Since our goal is to localize the bounding boxes more accurately at the object regions, we also consider the IoU of $0.7$ for an evaluation criterion, which only counts the detection results as correct when the overlap between the predicted bounding box and the ground truth is greater than $70\%$.
This is more challenging than common practices (e.g., IoU $\geq 0.5$), but will be a good indicator for a better localization of an object bounding box if successful.

\begin{figure}[t]
\centering
\vspace{-0.1in}
\includegraphics[width=0.85\columnwidth]{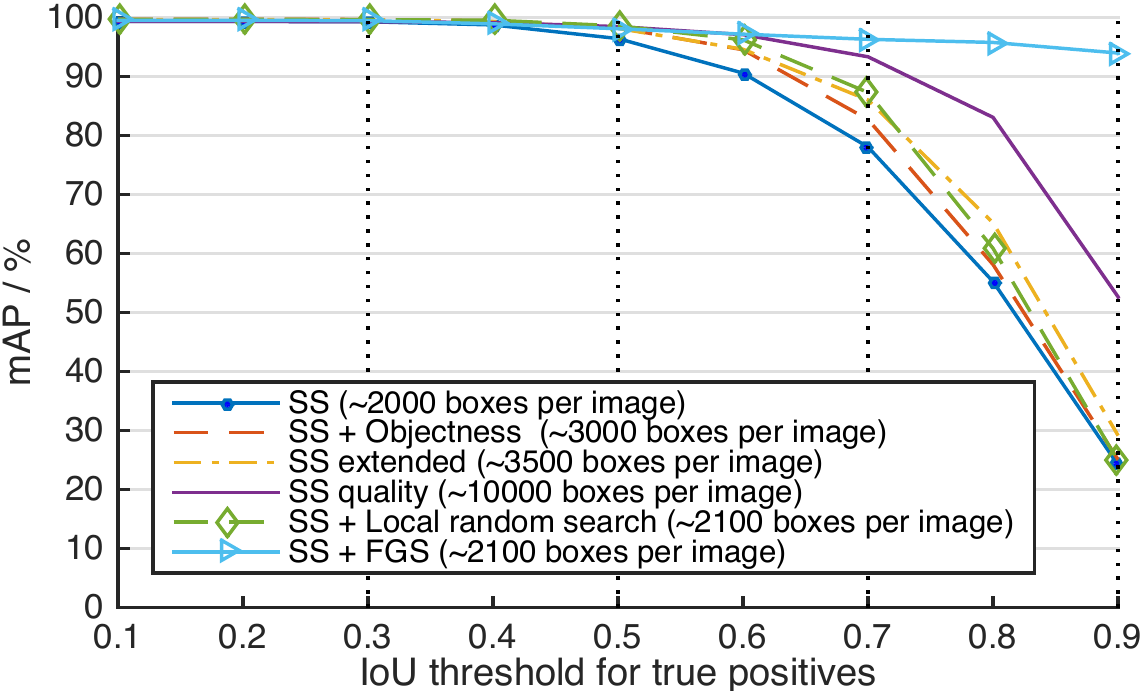}
\cutcaptionup
\protect\caption{\normalsize{mAP at different IoU criteria on PASCAL VOC 2007 test set using an oracle detector. We used different numbers of bounding boxes proposed by selective search (SS)~\cite{selective-search}, Objectness~\cite{objectness}, local random search, and our proposed FTS methods. The average numbers of bounding boxes used for evaluation are specified.}}
\label{fig:oracle-gp}
\cutcaptiondown
\vspace*{-0.15in}
\end{figure}

\begin{table*}[t]
\centering
{\setlength{\extrarowheight}{2pt}\scriptsize
{\begin{tabular}{l <{\hspace{-0.3em}}|>{\hspace{-0.5em}} c <{\hspace{-0.5em}}|>{\hspace{-0.3em}} c >{\hspace{-1.1em}}c >{\hspace{-1.1em}}c >{\hspace{-1.1em}}c >{\hspace{-1.1em}}c >{\hspace{-1.1em}}c >{\hspace{-1.1em}}c >{\hspace{-1.1em}}c >{\hspace{-1.1em}}c >{\hspace{-1.1em}}c >{\hspace{-1.1em}}c >{\hspace{-1.1em}}c >{\hspace{-1.1em}}c >{\hspace{-1.1em}}c >{\hspace{-1.1em}}c >{\hspace{-1.1em}}c >{\hspace{-1.1em}}c >{\hspace{-1.1em}}c >{\hspace{-1.1em}}c >{\hspace{-1.1em}}c <{\hspace{-0.3em}}| >{\hspace{-0.3em}}c}
\hline
Model & BBoxReg & aero & bike & bird & boat & bottle & bus & car & cat & chair & cow & table & dog & horse & mbike & person & plant & sheep & sofa & train & tv & mAP\\
\hline
R-CNN (AlexNet) & No & 64.2 & 69.7 & 50.0 & 41.9 & 32.0 & 62.6 & 71.0 & 60.7 & 32.7 & 58.5 & 46.5 & 56.1 & 60.6 & 66.8 & 54.2 & 31.5 & 52.8 & 48.9 & 57.9 & 64.7 & 54.2 \\
R-CNN (VGG) & No & 68.5 & 74.5 & 61.0 & 37.9 & 40.6 & 69.2 & 73.7 & 69.9 & 37.2 & 68.6 & 56.8 & 70.6 & 69.0 & 67.1 & 59.6 & 33.4 & 63.9 & 58.9 & 62.6 & 68.5 & 60.6 \\
+ StructObj & No & 68.7 & 73.5 & 62.6 & 40.6 & 41.5 & 69.6 & 73.5 & 71.1 & 39.9 & 69.6 & 58.1 & 70.0 & 67.5 & 69.8 & 59.8 & 35.9 & 63.6 & 59.0 & 62.6 & 67.7 & 61.2 \\
+ StructObj-FT & No & 69.3 & 75.2 & 62.2 & 39.4 & 42.3 & 70.7 & 74.5 & 74.3 & 40.4 & 71.3 & 59.8 & 72.0 & 69.8 & 69.4 & 60.3 & 35.3 & 64.5 & \bf{62.0} & 63.7 & 69.8 & 62.3 \\
+ FGS & No & 70.6 & 78.4 & 65.7 & 46.2 & 48.8 & \bf{74.6} & 77.0 & 74.3 & 42.7 & 70.8 & 60.9 & 75.1 & 75.8 & 70.7 & 66.3 & 37.1 & 66.3 & 57.6 & 66.6 & 71.0 & 64.8 \\
+ StructObj + FGS & No & \bf{73.4} & \bf{80.9} & 64.5 & \bf{46.7} & 49.1 & 73.9 & 78.2 & 76.8 & 44.8 & \bf{75.3} & \bf{63.0} & 75.3 & 74.2 & 72.7 & \bf{68.5} & 37.0 & 67.5 & 58.1 & \bf{66.9} & 70.5 & 65.9 \\
+ StructObj-FT + FGS & No & 72.5 & 78.8 & \bf{67.0} & 45.2 & \bf{51.0} & 73.8 & \bf{78.7} & \bf{78.3} & \bf{46.7} & 73.8 & 61.5 & \bf{77.1} & \bf{76.4} & \bf{73.9} & 66.5 & \bf{39.2} & \bf{69.7} & 59.4 & 66.8 & \bf{72.9} & \bf{66.5} \\
\hline
R-CNN (AlexNet) & Yes & 68.1 & 72.8 & 56.8 & 43.0 & 36.8 & 66.3 & 74.2 & 67.6 & 34.4 & 63.5 & 54.5 & 61.2 & 69.1 & 68.6 & 58.7 & 33.4 & 62.9 & 51.1 & 62.5 & 64.8 & 58.5 \\
R-CNN (VGG) & Yes & 70.8 & 77.1 & \bf{69.4} & 45.8 & 48.4 & 74.0 & 77.0 & 75.0 & 42.2 & 72.5 & 61.5 & 75.6 & 77.7 & 66.6 & 65.3 & 39.1 & 65.8 & 64.2 & 68.6 & 71.5 & 65.4 \\
+ StructObj & Yes & 73.1 & 77.5 & 69.2 & 47.6 & 47.6 & 74.5 & 78.2 & 75.4 & 44.5 & 76.3 & 64.9 & 76.7 & 76.3 & 69.9 & 68.1 & 39.4 & 67.0 & \bf{65.6} & 68.7 & 70.9 & 66.6 \\
+ StructObj-FT & Yes & 72.6 & 79.4 & \bf{69.4} & 45.2 & 47.8 & 74.4 & 77.8 & 76.5 & 45.4 & 76.3 & 61.4 & \bf{80.2} & 77.1 & 73.8 & 66.8 & 41.1 & 67.8 & 64.7 & 67.9 & 72.3 & 66.9 \\
+ FGS & Yes & \bf{74.2} & 78.9 & 67.8 & \bf{51.6} & 52.3 & 75.7 & 78.7 & 76.6 & 45.4 & 72.4 & 63.1 & 76.6 & \bf{79.3} & 70.7 & 68.0 & 40.3 & 67.8 & 61.8 & \bf{70.2} & 71.6 & 67.2 \\
+ StructObj + FGS & Yes & 74.1 & \bf{83.2} & 67.0 & 50.8 & 51.6 & \bf{76.2} & \bf{81.4} & 77.2 & 48.1 & \bf{78.9} & \bf{65.6} & 77.3 & 78.4 & \bf{75.1} & \bf{70.1} & 41.4 & 69.6 & 60.8 & \bf{70.2} & \bf{73.7} & \bf{68.5} \\
+ StructObj-FT + FGS & Yes & 71.3 & 80.5 & 69.3 & 49.6 & \bf{54.2} & 75.4 & 80.7 & \bf{79.4} & \bf{49.1} & 76.0 & 65.2 & 79.4 & 78.4 & 75.0 & 68.4 & \bf{41.6} & \bf{71.3} & 61.2 & 68.2 & 73.3 & 68.4 \\
\hline
\end{tabular}}}
\cutcaptionup
\caption{\normalsize{Test set mAP of VOC 2007 with IoU $> 0.5$. The entries with the best APs for each object category are bold-faced.\label{tab-voc2007-iou5}}}
\cutcaptiondown
\end{table*}

\begin{table*}[t]
\centering
{\setlength{\extrarowheight}{2pt}\scriptsize
{\begin{tabular}{l <{\hspace{-0.3em}}|>{\hspace{-0.5em}} c <{\hspace{-0.5em}}|>{\hspace{-0.3em}} c >{\hspace{-1.1em}}c >{\hspace{-1.1em}}c >{\hspace{-1.1em}}c >{\hspace{-1.1em}}c >{\hspace{-1.1em}}c >{\hspace{-1.1em}}c >{\hspace{-1.1em}}c >{\hspace{-1.1em}}c >{\hspace{-1.1em}}c >{\hspace{-1.1em}}c >{\hspace{-1.1em}}c >{\hspace{-1.1em}}c >{\hspace{-1.1em}}c >{\hspace{-1.1em}}c >{\hspace{-1.1em}}c >{\hspace{-1.1em}}c >{\hspace{-1.1em}}c >{\hspace{-1.1em}}c >{\hspace{-1.1em}}c <{\hspace{-0.3em}}| >{\hspace{-0.3em}}c}
\hline
Model & BBoxReg & aero & bike & bird & boat & bottle & bus & car & cat & chair & cow & table & dog & horse & mbike & person & plant & sheep & sofa & train & tv & mAP\\
\hline
R-CNN (AlexNet) & No & 32.9 & 40.1 & 19.7 & 18.7 & 11.1 & 39.4 & 40.5 & 26.5 & 14.8 & 29.8 & 24.5 & 26.4 & 23.7 & 31.9 & 18.5 & \bf{13.3} & 27.6 & 25.8 & 26.6 & 39.5 & 26.6 \\
R-CNN (VGG) & No & 40.2 & 43.3 & 23.4 & 14.4 & 13.3 & 48.2 & 44.5 & 36.4 & 17.1 & 34.0 & 27.9 & 36.3 & 26.8 & 28.2 & 21.2 & 10.3 & 33.7 & 36.6 & 31.6 & 48.9 & 30.8 \\
+ StructObj & No & 42.5 & 44.4 & 24.5 & 17.8 & 15.3 & 46.8 & 46.4 & 37.9 & 17.6 & 33.4 & 26.6 & 36.8 & 24.3 & 31.5 & 21.3 & 10.4 & 30.0 & 36.1 & 30.6 & 46.3 & 31.0 \\
+ StructObj-FT & No & 44.1 & 47.1 & 23.4 & 16.6 & 16.4 & 50.1 & 48.7 & 39.7 & 18.4 & 39.4 & 28.6 & 38.6 & 27.5 & 32.4 & 23.6 & 11.1 & 33.1 & \bf{41.0} & 34.3 & 49.6 & 33.2 \\
+ FGS & No & 44.3 & 55.5 & 28.9 & 19.1 & 22.9 & 56.9 & 57.6 & 37.8 & 19.6 & 35.7 & \bf{31.9} & 38.1 & \bf{43.0} & 42.7 & 30.3 & 9.8 & \bf{42.3} & 33.3 & 43.4 & 55.4 & 37.4 \\
+ StructObj + FGS & No & 43.5 & 56.1 & 30.9 & 18.7 & 24.9 & 55.2 & 57.6 & 38.9 & 20.7 & 38.6 & 28.4 & 37.7 & 38.7 & \bf{46.3} & 30.9 & 8.4 & 37.6 & 37.0 & 42.2 & 51.3 & 37.2 \\
+ StructObj-FT + FGS & No & \bf{46.3} & \bf{58.1} & \bf{31.1} & \bf{21.6} & \bf{25.8} & \bf{57.1} & \bf{58.2} & \bf{43.5} & \bf{23.0} & \bf{46.4} & 29.0 & \bf{40.7} & 40.6 & \bf{46.3} & \bf{33.4} & 10.6 & 41.3 & 40.9 & \bf{45.8} & \bf{56.3} & \bf{39.8} \\
\hline
R-CNN (AlexNet) & Yes & 47.6 & 48.7 & 25.3 & \bf{25.0} & 17.3 & 53.4 & 54.6 & 36.8 & 16.7 & 42.3 & 31.6 & 35.8 & 38.0 & 41.8 & 24.5 & 14.3 & 38.8 & 28.9 & 34.0 & 49.0 & 35.2 \\
R-CNN (VGG) & Yes & 45.1 & 48.6 & 26.0 & 18.2 & 21.2 & 57.2 & 52.4 & 37.3 & 20.1 & 33.7 & 31.9 & 38.8 & 39.6 & 36.3 & 26.5 & 9.2 & 37.8 & 33.4 & 39.4 & 50.7 & 35.2 \\
+ StructObj & Yes & 49.4 & 56.5 & \bf{36.5} & 21.3 & 23.3 & 61.0 & 58.1 & 44.3 & 20.8 & 47.4 & 33.3 & 39.8 & 40.7 & 45.9 & 31.0 & \bf{14.7} & 39.6 & 42.9 & 45.7 & 56.9 & 40.5 \\
+ StructObj-FT & Yes & 49.3 & 58.1 & 35.4 & 23.3 & 24.4 & 62.3 & 60.1 & 45.8 & 21.8 & 48.7 & 32.4 & 41.8 & 43.2 & 45.7 & 32.0 & 14.4 & 44.6 & \bf{45.1} & 48.6 & 59.8 & 41.8 \\
+ FGS & Yes & 50.9 & 59.8 & 34.4 & 20.9 & \bf{31.6} & \bf{66.1} & 62.3 & 44.9 & 22.0 & 46.5 & \bf{36.8} & 42.5 & 51.4 & 46.8 & 34.1 & 13.5 & \bf{44.7} & 39.1 & 48.9 & 57.7 & 42.7 \\
+ StructObj + FGS & Yes & \bf{53.6} & 60.7 & 32.1 & 19.9 & 31.3 & 63.2 & 63.2 & 46.4 & 23.6 & \bf{53.0} & 34.9 & 40.4 & \bf{53.6} & \bf{49.9} & 34.6 & 10.2 & 42.2 & 40.1 & 48.3 & 58.3 & 43.0 \\
+ StructObj-FT + FGS & Yes & 47.1 & \bf{61.8} & 35.2 & 18.1 & 29.7 & 66.0 & \bf{64.7} & \bf{48.0} & \bf{25.3} & 50.4 & 34.9 & \bf{43.7} & 50.8 & 49.4 & \bf{36.8} & 13.7 & \bf{44.7} & 43.6 & \bf{49.8} & \bf{60.5} & \bf{43.7} \\
\hline
\end{tabular}}}
\cutcaptionup
\caption{\normalsize{Test set mAP of VOC 2007 with IoU $> 0.7$. The entries with the best APs for each object category are bold-faced.\label{tab-voc2007-iou7}}}
\cutcaptiondown
\vspace*{-0.05in}
\end{table*}

\begin{table*}[t]
\centering
{\setlength{\extrarowheight}{2pt}\scriptsize
{\begin{tabular}{l <{\hspace{-0.3em}}|>{\hspace{-0.5em}} c <{\hspace{-0.5em}}|>{\hspace{-0.3em}} c >{\hspace{-1em}}c >{\hspace{-1em}}c >{\hspace{-1em}}c >{\hspace{-1em}}c >{\hspace{-1em}}c >{\hspace{-1em}}c >{\hspace{-1em}}c >{\hspace{-1em}}c >{\hspace{-1em}}c >{\hspace{-1em}}c >{\hspace{-1em}}c >{\hspace{-1em}}c >{\hspace{-1em}}c >{\hspace{-1em}}c >{\hspace{-1em}}c >{\hspace{-1em}}c >{\hspace{-1em}}c >{\hspace{-1em}}c >{\hspace{-1em}}c <{\hspace{-0.3em}}| >{\hspace{-0.3em}}c}
\hline
Model & BBoxReg & aero & bike & bird & boat & bottle & bus & car & cat & chair & cow & table & dog & horse & mbike & person & plant & sheep & sofa & train & tv & mAP\\
\hline
R-CNN (AlexNet) & No & 68.1 & 63.8 & 46.1 & 29.4 & 27.9 & 56.6 & 57.0 & 65.9 & 26.5 & 48.7 & 39.5 & 66.2 & 57.3 & 65.4 & 53.2 & 26.2 & 54.5 & 38.1 & 50.6 & 51.6 & 49.6\\
R-CNN (VGGNet) & No & 76.3 & 69.8 & 57.9 & 40.2 & 37.2 & 64.0 & 63.7 & 80.2 & 36.1 & 63.6 & 47.3 & 81.1 & 71.2 & 73.8 & 59.5 & 30.9 & 64.2 & 52.2 & 62.4 & 58.7 & 59.5\\
\hline
R-CNN (AlexNet) & Yes & 71.8 & 65.8 & 52.0 & 34.1 & 32.6 & 59.6 & 60.0 & 69.8 & 27.6 & 52.0 & 41.7 & 69.6 & 61.3 & 68.3 & 57.8 & 29.6 & 57.8 & 40.9 & 59.3 & 54.1 & 53.3 \\
R-CNN (VGGNet) & Yes & 79.2 & 72.3 & 62.9 & 43.7 & 45.1 & 67.7 & 66.7 & 83.0 & 39.3 & 66.2 & 51.7 & 82.2 & 73.2 & 76.5 & 64.2 & 33.7 & 66.7 & 56.1 & 68.3 & 61.0 & 63.0 \\
+ StructObj & Yes & 80.9 & 74.8 & 62.7 & 42.6 & 46.2 & 70.2 & 68.6 & 84.0 & 42.2 & 68.2 & 54.1 & 82.2 & 74.2 & 79.8 & 66.6 & 39.3 & 67.6 & \bf{61.0} & 71.3 & 65.2 & 65.1 \\
+ FGS & Yes & 80.5 & 73.5 & \bf{64.1} & \bf{45.3} & 48.7 & 66.5 & 68.3 & 82.8 & 39.8 & 68.2 & 52.7 & 82.1 & 75.1 & 76.6 & 66.3 & 35.5 & 66.9 & 56.8 & 68.7 & 61.6 & 64.0 \\
+ StructObj + FGS & Yes & \bf{82.9} & \bf{76.1} & \bf{64.1} & 44.6 & \bf{49.4} & \bf{70.3} & \bf{71.2} & \bf{84.6} & \bf{42.7} & 68.6 & \bf{55.8} & \bf{82.7} & \bf{77.1} & \bf{79.9} & \bf{68.7} & \bf{41.4} & \bf{69.0} & 60.0 & \bf{72.0} & \bf{66.2} & \bf{66.4} \\
\hline
NIN~\citep{DBLP:journals/corr/LinCY13} & - & 80.2 & 73.8 & 61.9 & 43.7 & 43.0 & \bf{70.3} & 67.6 & 80.7 & 41.9 & \bf{69.7} & 51.7 & 78.2 & 75.2 & 76.9 & 65.1 & 38.6 & 68.3 & 58.0 & 68.7 & 63.3 & 63.8 \\
\hline
\end{tabular}}}
\cutcaptionup
\vspace*{0.03in}
\caption{\normalsize{Test set mAP of VOC 2012 with IoU $> 0.5$. The entries with the best APs for each object category are bold-faced.\label{tab-voc2012-iou5}}}
\cutcaptiondown
\vspace*{0.03in}
\end{table*}

\begin{figure*}[t]
\begin{center}
{\renewcommand{\arraystretch}{0.3} \begin{tabular}{c <{\hspace{-0.5em}} >{\hspace{-0.5em}}c c <{\hspace{-0.5em}} >{\hspace{-0.5em}}c c <{\hspace{-0.5em}} >{\hspace{-0.5em}}c c <{\hspace{-0.5em}} >{\hspace{-0.5em}}c c <{\hspace{-0.5em}} >{\hspace{-0.5em}}c}
\includegraphics[width = 0.55in]{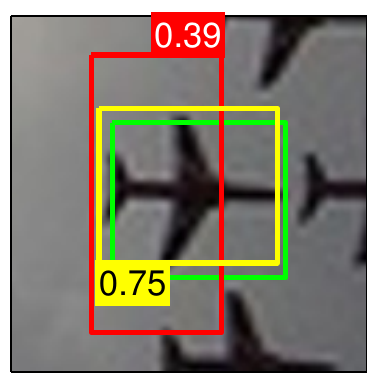} &
\includegraphics[width = 0.55in]{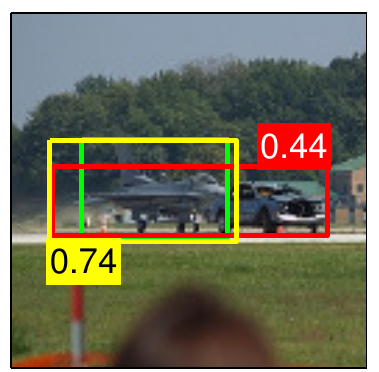} &
\includegraphics[width = 0.55in]{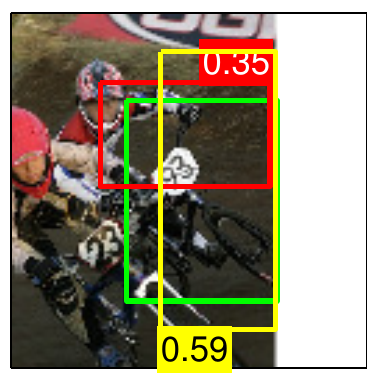} &
\includegraphics[width = 0.55in]{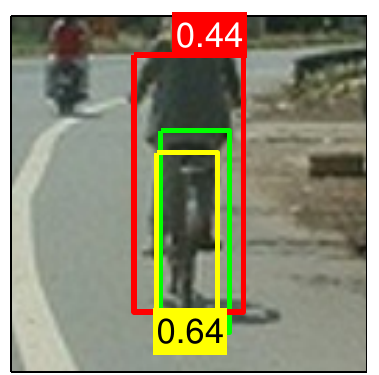} &
\includegraphics[width = 0.55in]{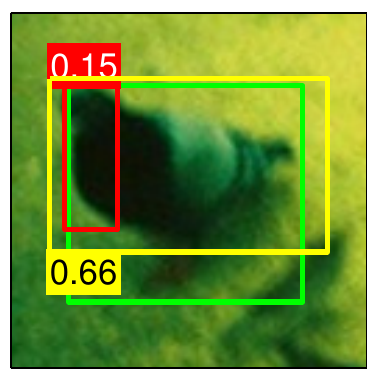} &
\includegraphics[width = 0.55in]{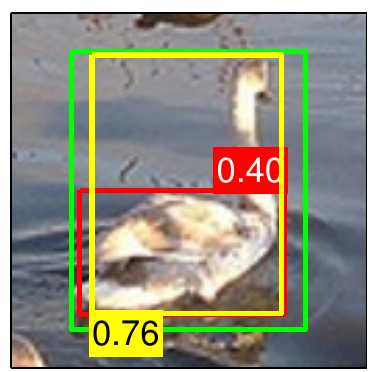} & 
\includegraphics[width = 0.55in]{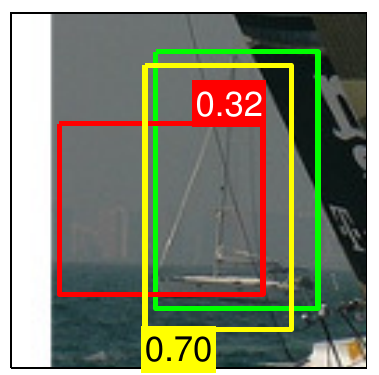} &
\includegraphics[width = 0.55in]{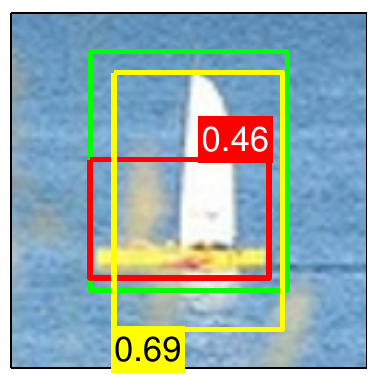} &
\includegraphics[width = 0.55in]{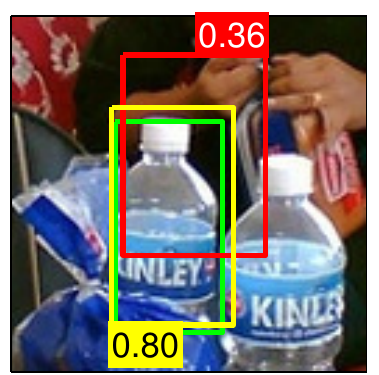} &
\includegraphics[width = 0.55in]{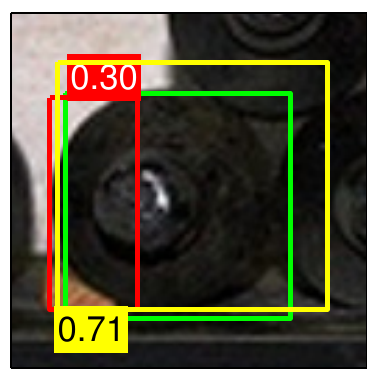} \\
\multicolumn{2}{c}{\footnotesize{aeroplane}} & \multicolumn{2}{c}{\footnotesize{bicycle}} & \multicolumn{2}{c}{\footnotesize{bird}} & \multicolumn{2}{c}{\footnotesize{boat}} & \multicolumn{2}{c}{\footnotesize{bottle}} \\
\includegraphics[width = 0.55in]{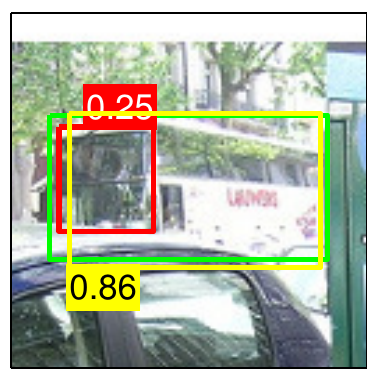} &
\includegraphics[width = 0.55in]{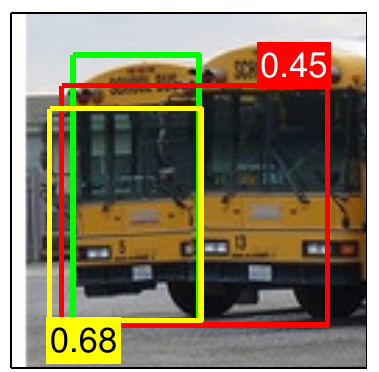} &
\includegraphics[width = 0.55in]{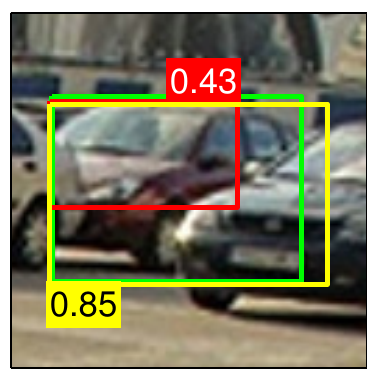} &
\includegraphics[width = 0.55in]{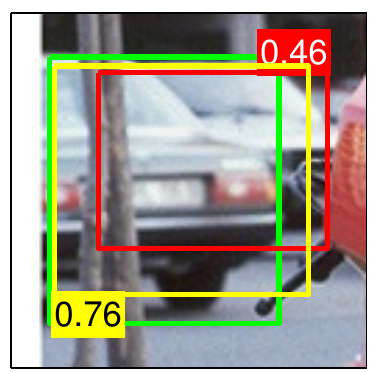} &
\includegraphics[width = 0.55in]{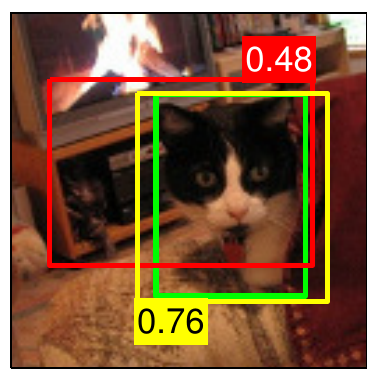} &
\includegraphics[width = 0.55in]{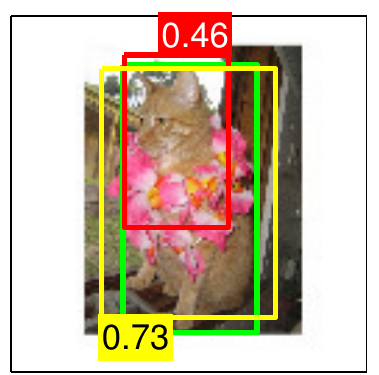} &
\includegraphics[width = 0.55in]{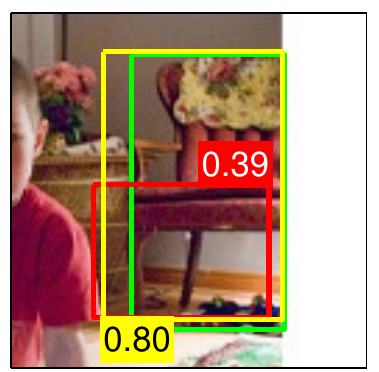} &
\includegraphics[width = 0.55in]{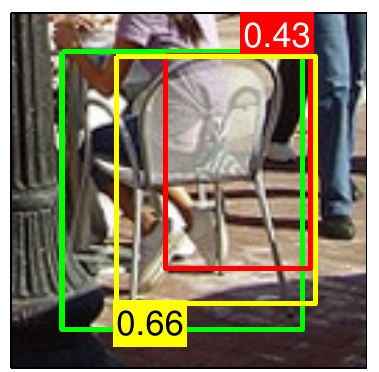} &
\includegraphics[width = 0.55in]{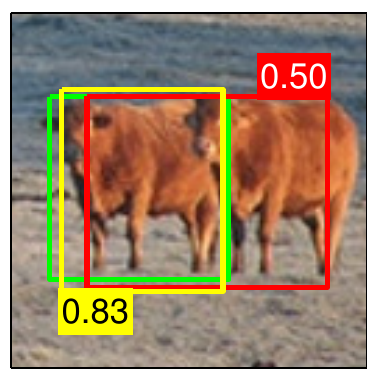} &
\includegraphics[width = 0.55in]{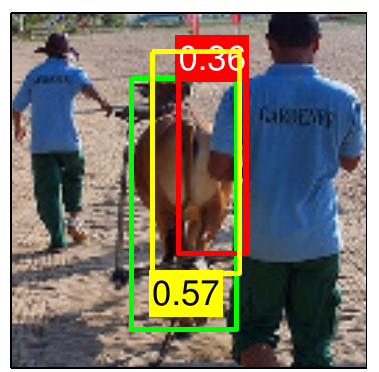} \\
\multicolumn{2}{c}{\footnotesize{bus}} & \multicolumn{2}{c}{\footnotesize{car}} & \multicolumn{2}{c}{\footnotesize{cat}} & \multicolumn{2}{c}{\footnotesize{chair}} & \multicolumn{2}{c}{\footnotesize{cow}} \\
\includegraphics[width = 0.55in]{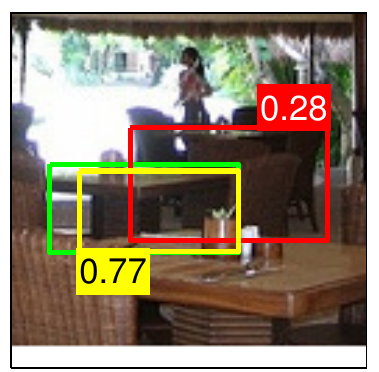} &
\includegraphics[width = 0.55in]{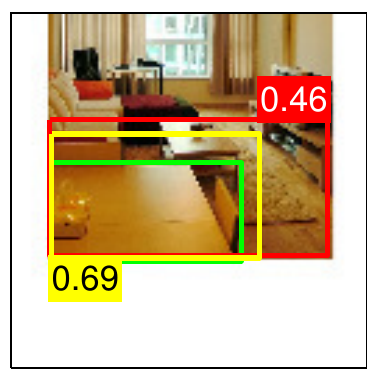} &
\includegraphics[width = 0.55in]{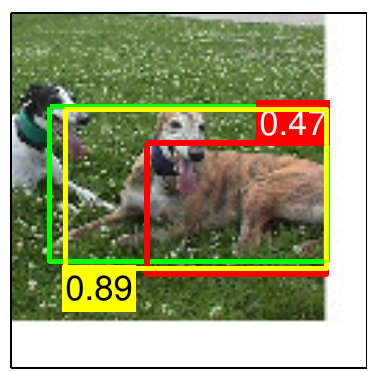} &
\includegraphics[width = 0.55in]{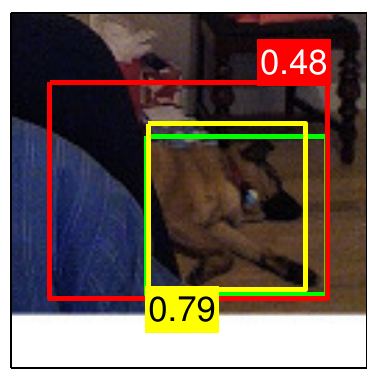} &
\includegraphics[width = 0.55in]{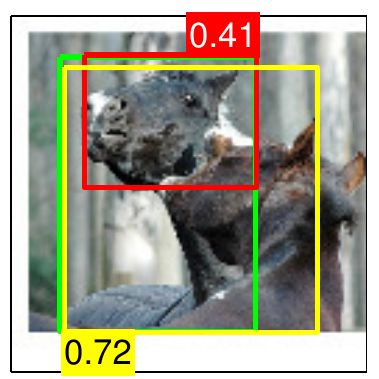} &
\includegraphics[width = 0.55in]{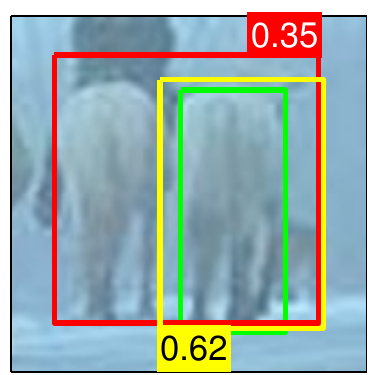} & 
\includegraphics[width = 0.55in]{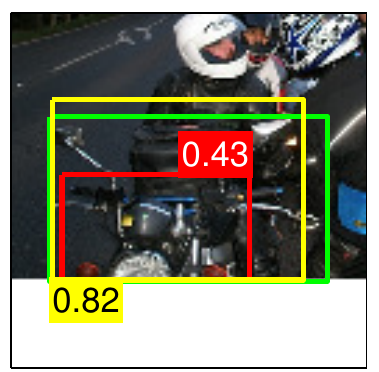} &
\includegraphics[width = 0.55in]{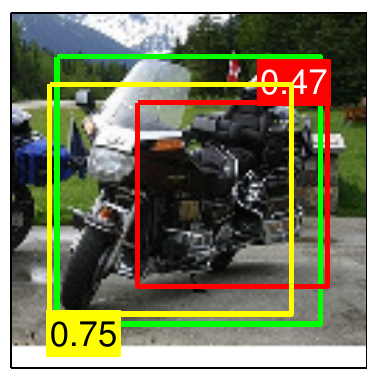} &
\includegraphics[width = 0.55in]{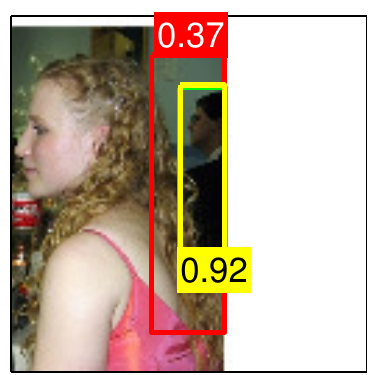} &
\includegraphics[width = 0.55in]{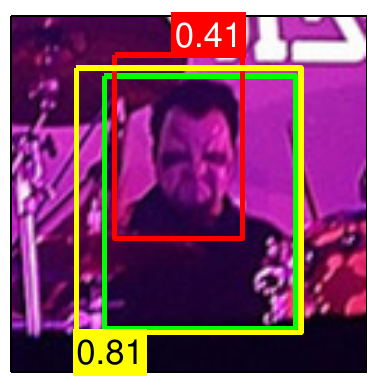} \\
\multicolumn{2}{c}{\footnotesize{diningtable}} & \multicolumn{2}{c}{\footnotesize{dog}} & \multicolumn{2}{c}{\footnotesize{horse}} & \multicolumn{2}{c}{\footnotesize{motorbike}} & \multicolumn{2}{c}{\footnotesize{person}} \\
\includegraphics[width = 0.55in]{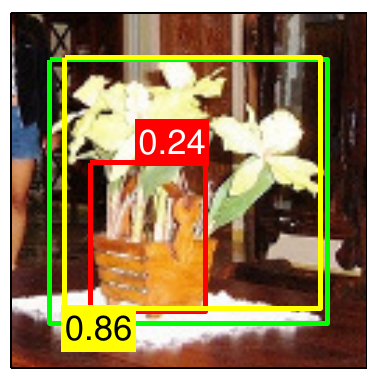} &
\includegraphics[width = 0.55in]{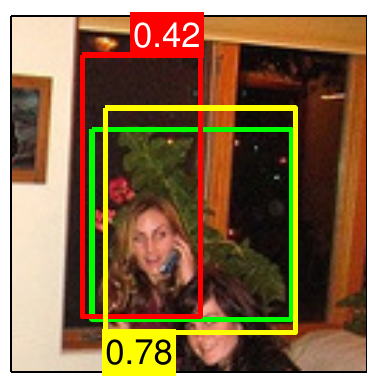} &
\includegraphics[width = 0.55in]{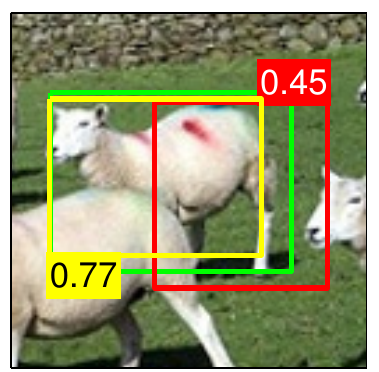} &
\includegraphics[width = 0.55in]{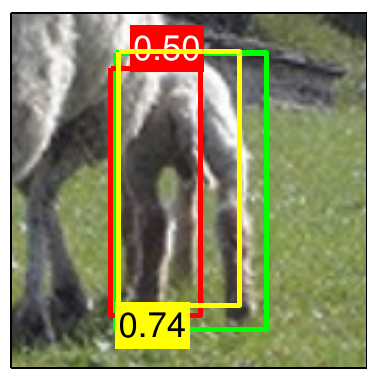} &
\includegraphics[width = 0.55in]{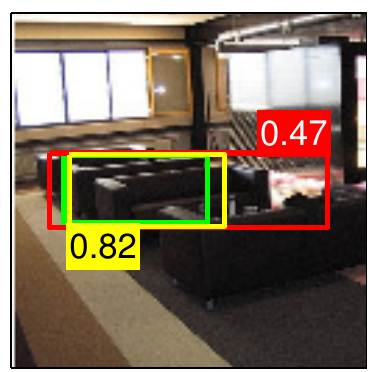} &
\includegraphics[width = 0.55in]{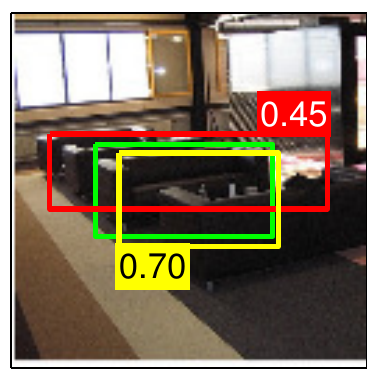} &
\includegraphics[width = 0.55in]{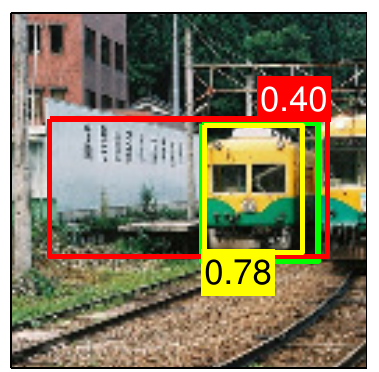} &
\includegraphics[width = 0.55in]{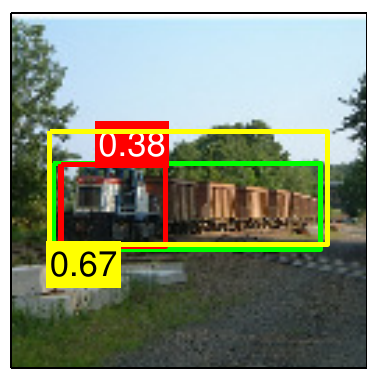} &
\includegraphics[width = 0.55in]{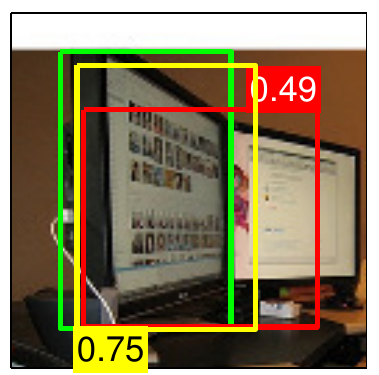} &
\includegraphics[width = 0.55in]{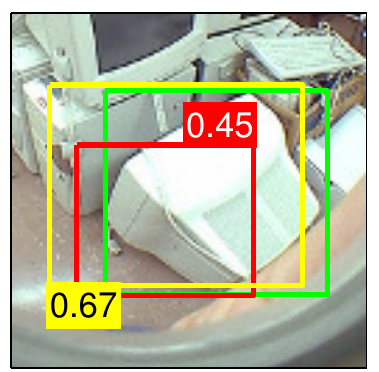} \\
\multicolumn{2}{c}{\footnotesize{pottedplant}} & \multicolumn{2}{c}{\footnotesize{sheep}} & \multicolumn{2}{c}{\footnotesize{sofa}} & \multicolumn{2}{c}{\footnotesize{train}} & \multicolumn{2}{c}{\footnotesize{tvmonitor}} \\
\end{tabular}}
\normalsize
\cutcaptionup
\vspace*{0.03in}
\caption{\normalsize{Detection examples from PASCAL VOC 2007 test set. Two examples from 20 object categories are shown, with the ground truth bounding boxes (green), the boxes obtained by baseline R-CNN (VGGNet) (red), and those obtained by the proposed R-CNN + StructObj + FGS (yellow). The numbers near the bounding boxes denote the IoU with the ground truth.}}
\label{fig-2007examples}
\cutcaptiondown
\vspace*{-0.15in}
\end{center}
\end{figure*}

\cutsubsectionup
\subsection{FGS efficacy test with oracle detector}
\label{exp-oracle}
\cutsubsectiondown

Before reporting the performance of the proposed methods in R-CNN framework, we demonstrate the efficacy of FGS algorithm using an oracle detector.
We design a hypothetical oracle detector whose score function is defined as $f_{\text{ideal}}(x_{i},y)=\operatorname{IoU}(y,y_{i})$, where $y_{i}$ is a ground truth annotation for an image $x_{i}$.
The score function is ideal in the sense that it outputs high scores for bounding boxes with high overlap with the ground truth and vice versa, overall achieving 100\% mAP.

We summarize the results in Figure~\ref{fig:oracle-gp}. We report the performance on the VOC 2007 test set at different levels of IoU criteria ($0.1,\cdots,0.9$) for the baseline selective search (SS; ``fast mode'' in \citep{selective-search}), selective search with objectness~\citep{objectness} (SS + Objectness), selective search with extended super-pixel similarity measurements (SS extended)~\citep{selective-search}, ``quality mode'' of selective search (SS quality)~\cite{selective-search}, local random search,\footnote{Like local FGS, local random search first determine the local search regions by NMS. However, it randomly choose a fixed number of bounding box in those regions rather than sequentially proposing new boxes based on some informed method.} and the proposed FGS method with the baseline selective search.

For low values of IoU ($\leq 0.3$), all methods using the oracle detectors performed almost perfectly due to the ideal score function.
However, we found that the detection performance with different region proposal schemes other than our proposed FGS algorithm start to break down at high values of IoU. 
For example, the performance of SS, SS + Objectness, SS extended, and local random search methods, which used around $2,000$ $\sim$ $3,500$ bounding boxes per image in average, significantly dropped at IoU $\geq 0.5$.
SS quality method kept pace with the FGS method until IoU of $0.6$, but again, the performance started to drop at IoU $\geq 0.7$.

On the other hand, the performance of FGS dropped $5\%$ in mAP at IoU of $0.9$ by only introducing approximately $100$ new bounding boxes per image.
Given that SS quality requires $10,000$ region proposals per image, our proposed FGS method is much more computationally efficient ($\sim 80\%$ less bounding boxes) while localizing the bounding boxes much more accurately.
This provides an insight that, if the detector is accurate, our Bayesian optimization framework would limit the number of bounding boxes to a manageable number (e.g., few thousands per image on average) to achieve almost perfect detection results.

We also report similar experimental analysis for the real detector trained with the proposed structured objective in \supprefp{\ref{sup:mAP-diff-pr}}.

\cutsubsectionup
\subsection{PASCAL VOC 2007}
\label{exp-voc2007}
\cutsubsectiondown
In this section, we demonstrate the performance of our proposed methods on PASCAL VOC 2007~\cite{pascal-voc-2007} detection task (comp4), a standard benchmark for object detection problem.
Similarly to the training pipeline of R-CNN~\citep{rcnn-cvpr}, we finetuned the CNN models (with softmax classification layer) pretrained on ImageNet database using images from both train and validation sets of VOC 2007 and further trained the network with linear SVM (baseline) or the proposed structured SVM objective.
We evaluated on the test set using the proposed FGS algorithm.
For post-processing, we performed NMS and bounding box regression~\citep{rcnn-cvpr}.

Figure~\ref{fig-2007examples} shows representative examples of successful detection using our method. 
For these cases, our method can localize objects accurately even if the initial bounding box proposals don't have good overlaps with the ground truth. 
We show more examples (including the failure cases) in \supprefp{\ref{sup:example-improvement}, \ref{sup:example-fp}, \ref{sup:example-rand}}.

The summary results are in Table~\ref{tab-voc2007-iou5} with IoU criteria of $0.5$ and Table~\ref{tab-voc2007-iou7} with $0.7$.
We report the performance with the AlexNet~\citep{alex-imagenet} and the VGGNet (16 layers)~\citep{vggnet}, a deeper CNN model than AlexNet that showed a significantly better recognition performance and achieved the best performance on object localization task in ILSVRC 2014.\footnote{The 16-layer VGGNet can be downloaded from: \url{https://gist.github.com/ksimonyan/211839e770f7b538e2d8}.}
First of all, we observed the significant performance improvement by simply having a better CNN model.
Building upon the VGGNet, the FGS improved the performance by $4.2\%$ and $1.8\%$ in mAP without and with bounding box regression (Table~\ref{tab-voc2007-iou5}).
It becomes much more significant when we consider IoU criteria of $0.7$ (Table~\ref{tab-voc2007-iou7}), improving upon the baseline model by $6.6\%$ and $7.5\%$ in mAP without and with bounding box regression.
The results demonstrate that our FGS algorithm is effective in accurately localizing the bounding box of an object.

Further improvement has been made by training a classifier with structured SVM objective; we obtained $\mathbf{68.5}\%$ mAP in IoU criteria of $0.5$, which, to our knowledge, is higher than the best published results, and $\mathbf{43.0}\%$ mAP in IoU criteria of $0.7$ with FGS and bounding box regression by training the classification layer only (``StructObj").
By finetuning the whole CNN classifiers (``StructObj-FT''), we observed extra improvement for most cases; for example, we obtained $\mathbf{43.7}\%$ mAP in IOU criteria of $0.7$, which improves by $0.7\%$ in mAP over the method without finetuning.
However, for IoU$>$0.5 criterion, the overall improvement due to finetuning was relatively small, especially when using bounding box regression. 
In this case, considering the high computational cost for finetuning, we found that training only the classification layer is practically a sufficient way to learn a good localization-aware classifier.

We provide in-depth analysis of our proposed methods in \suppname{}. 
Specifically, we report the precision-recall curves of different combinations of the proposed methods (\suppref{\ref{sup:pr-curves}}), the performance of FGS with different GP iterations (\suppref{\ref{sup:per-iter-FGS}}), the analysis of localization accuracy (\suppref{\ref{sup:loc-distr}}), and more detection examples.

\cutsubsectionup
\subsection{PASCAL VOC 2012}
\label{exp-voc2012}
\cutsubsectiondown
We also evaluate the performance of the proposed methods on PASCAL VOC 2012~\cite{pascal-voc-2012}.
As the data statistics are similar to VOC 2007, we used the same hyperparameters as described in Section~\ref{exp-voc2007} for this experiment.
We report the test set mAP over 20 object categories in Table~\ref{tab-voc2012-iou5}.
Our proposed method shows improvement by $2.1\%$ with R-CNN + StructObj and $1.0\%$ with R-CNN + FGS over baseline R-CNN using VGGNet.
Finally, we obtained $\mathbf{66.4}\%$ mAP by combining the two methods, which significantly improved upon the baseline R-CNN model and the previously published results on the leaderboard.

\cutsectionup
\section{Conclusion}
\label{sec-conc}
\cutsectiondown

In this work, we proposed two complementary methods to improve the performance of object detection in R-CNN framework with 1) fine-grained search algorithm in a Bayesian optimization framework to refine the region proposals and 2) a CNN classifier trained with structured SVM objective to improve localization.
We demonstrated the state-of-the-art detection performance on PASCAL VOC 2007 and 2012 benchmarks under standard localization requirements.
Our methods showed more significant improvement with higher IoU evaluation criteria (e.g., IoU $=0.7$), and hold promise for mission-critical applications that require highly precise localization, such as autonomous driving, robotic surgery and manipulation.

\cutsectionup
\section*{Acknowledgments}
\cutsectiondown

This work was supported by Samsung Digital Media and Communication Lab, Google Faculty Research Award, ONR grant N00014-13-1-0762, China Scholarship Council, and Rackham Merit Fellowship. We also acknowledge NVIDIA for the donation of GPUs. Finally, we thank Scott Reed, Brian Wang, Junhyuk Oh, Xinchen Yan, Ye Liu, Wenling Shang, and Roni Mittelman for helpful discussions.

\bibliographystyle{abbrvnat}
\begingroup
{
\small
\setstretch{1}
\setlength{\bibsep}{0pt}
\bibliography{cnn-det-abbrv}
}
\endgroup

\clearpage
\onecolumn

\appendix

\setsecnumdepth{none}
\part{Appendices}
\resetsecnumdepth

\vfill

{
\Large
Contents of Appendices
}
\medskip{}

\begingroup
\hypersetup{pagebackref=true,breaklinks=true,colorlinks,letterpaper=true,bookmarks=false,linkcolor=black}

\makeatletter
\renewcommand*{\l@section}{\medskip{}\@dottedtocline{1}{2.5em}{2.8em}}
\makeatother

\onehalfspacing

\renewcommand{\ptcSfont}{\normalsize\rm}

\parttoc{}

\endgroup

\renewcommand{\thefigure}{A-\arabic{figure}}
\renewcommand{\thetable}{A-\arabic{table}}
\renewcommand{\theequation}{A-\arabic{equation}}
\renewcommand{\thealgorithm}{A-\arabic{algorithm}}
\renewcommand{\thesection}{A\arabic{section}}

\setcounter{figure}{0}
\setcounter{table}{0}
\setcounter{equation}{0}
\setcounter{algorithm}{0}
\setcounter{section}{0}

\vfill

\section{Parameter estimation for finetuning with structured SVM objective}
\label{sup:finetuning}
The model parameters are updated via gradient descent.
The gradient, for example, with respect to the CNN parameters $\Theta$ $(\neq w)$ for positive examples is given as follows: \begin{equation}
\frac{\partial h_{\text{pos}, i}}{\partial \Theta} = \begin{cases}
0 & ,\; h_{\text{pos}, i} = 0\\
-w^{\top}\frac{\partial\phi(x_i,y_i)}{\partial \Theta} & ,\; h_{\text{pos}, i} = \;\;1-w^{\top}\phi(x_i, y_i)\\
w^{\top}\frac{\partial(\phi(x_i,\hat{y}_{i}) - \phi(x_i,y_i))}{\partial \Theta} & ,\; \text{otherwise}
\end{cases}\end{equation}
where $\hat{y}_{i} = \argmax_{y\in \mathcal{Y}_i^{(l=1)}} \big(w^{\top}\phi(x_i,y) + \Delta(y,y_i)\big)$.
Similarly, the gradient for negative examples can be computed as follows: \begin{equation}
\frac{\partial h_{\text{neg}, i}}{\partial \Theta} = 
\begin{cases}
0 & ,\; h_{\text{neg}, i} = 0\\
w^{\top}\frac{\partial\phi(x_i,\hat{y}_i)}{\partial \Theta} & ,\; h_{\text{neg}, i} = 1+w^{\top}\phi(x_i, \hat{y}_i)
\end{cases} \end{equation}
where $\hat{y}_{i} = \argmax_{y\in \mathcal{Y}_i^{(l=1)}} w^{\top}\phi(x_i,y)$. 
The gradient with respect to the parameters of all layers of CNN can be computed efficiently using backpropagation.
When finetuning the entire network, the parameter updated in the hard mining procedure illustrated by Algorithm~\ref{alg:hard-mining} is done by replacing $\hat{w}$ with the CNN parameters.

\vfill

\clearpage

\section{Details on hard negative data mining}
\label{sup:hard-mining}

The active set consisting of the hard training instances are updated in two steps during the iterative learning process.
First, we include instances $y\in\mathcal{Y}_i$ to the active set when they are likely to be active, i.e., affect the gradient:
\begin{align}
w^{\top}\left(\phi(x_{i},y) - \phi(x_{i},y_{i})\right) + \Delta^{\text{loc}}(y,y_{i}) &\geq \max\big\{0, 1-w^{\top}\phi(x_i,y_i)\big\} - \epsilon_{1}, \forall i\in I_{\text{pos}}\label{eq:hard-thresh-1}\\
1+w^{\top}\phi(x_i,y) &\geq -\epsilon_{1}, \forall i \in I_{\text{neg}} \label{eq:hard-thresh-2}
\end{align}
Second, once new parameters are estimated, we exclude instances from the current active set when they are likely to be inactive, i.e., have no effect on the gradient:
\begin{align}
w^{\top}\left(\phi(x_{i},y) - \phi(x_{i},y_{i})\right) + \Delta^{\text{loc}}(y,y_{i}) &\leq \min\big\{0, 1-w^{\top}\phi(x_i,y_i)\big\} - \epsilon_{2}, \forall i\in I_{\text{pos}}\label{eq:evict-thresh-1}\\
1+w^{\top}\phi(x_i,y) &\leq -\epsilon_{2}, \forall i \in I_{\text{neg}}  \label{eq:evict-thresh-2}
\end{align}
In our experiments, we used $\epsilon_{1} = 0.0001$ and $\epsilon_{2} = 0.2$.
The values of $\epsilon_{1},\epsilon_{2}$ are the same as those for the SVM training in R-CNN~\citep{rcnn-cvpr}. We did not observe a noticeable performance fluctuation due to different $\epsilon_{1},\epsilon_{2}$ values. 
Algorithm~\ref{alg:hard-mining} summarizes the hard-mining procedure.  

\begin{algorithm}[h]
\caption{Parameter estimation with hard mining \label{alg:hard-mining}}
\begin{algorithmic}[1]
{
\Require  Initial parameters $w_0$, maximum epoch number $\mathtt{epoch}_{\text{max}}$, training images $\{x_i,y_i\}_{i=1}^\numimg$, positive and negative index $I_{\text{pos}},I_{\text{neg}}$
\Ensure  Final parameters $\hat{w}$.
\State The active set $\mathcal{A},\leftarrow \{(x_i,y_i):i \in I_{\text{pos}} \}$
\State $\mathcal{A},\mathcal{A}_{\text{inc}}\leftarrow \varnothing$, $\hat{w}\leftarrow w_0$
\For{$\mathtt{epoch}=1,\ldots,\mathtt{epoch}_{\text{max}}$}
\For{$i=1,\ldots,\numimg$}
\State $\mathcal{A}_{\text{inc}}\leftarrow \mathcal{A}_{\text{inc}} \cup \{(y,x_i):y$ s.t.\ \eqref{eq:hard-thresh-1}, \eqref{eq:hard-thresh-2} for $x_i,\hat{w}\}$
\If{$|\mathcal{A}_{\text{inc}}| \geq \mathtt{update\_threshold}$ or $i==\numimg$}
\State $\mathcal{A} \leftarrow \mathcal{A} \cup \mathcal{A}_{\text{inc}}$, $\mathcal{A}_{\text{inc}} \leftarrow \varnothing$ 
\State update the classifier/network parameters $\hat{w}$ on $\mathcal{A}$
\State $\mathcal{A}\leftarrow \mathcal{A} \backslash \{(y,x)\in\mathcal{A}:y$ s.t.\ \eqref{eq:evict-thresh-1}, \eqref{eq:evict-thresh-2} for $x,\hat{w}\}$
\EndIf
\EndFor
\EndFor
}
\end{algorithmic}
\end{algorithm}

\section{Implementation details on model parameter estimation}
\label{sup:learning-details}

For our experiments on PASCAL VOC 2007 and VOC 2012, we first finetune the CNN pretrained on ImageNet by stochastic gradient descent with a 21-way softmax classification layer, where 20 ways are for the 20 object categories of interest, and the rest 1 way is for the background.   
In this step, the SGD learning rate starts at 0.0003, and decreases by 0.3 every 15000 iterations with a mini-batch size of 48. 
We set the momentum to 0.9, and the weight decay to 0.0005 for all the layers.  

After that, we replace the softmax loss layer with a 20-way structured loss layer, where each way is a binary classifier, and the hinge loss for different category are simply summed up. 

For classification layer only learning, L-BFGS is adopted, as batch gradient descent for a single layer. 
Each category has an associated active set for hard negative mining. 
The classifier update happens independently for each category when $5000$ ($\mathtt{update\_threshold}$ in Algorithm~\ref{alg:hard-mining}) new hard examples are added to the active set. 
It is worth mentioning that, in the beginning of the hard negative mining, significantly more positive images are present than the negative images, resulting in serious unbalance of the active training samples. 
As a heuristic to avoid this problem, we limit the number of positive image to the number of the active negative images when classifier update happens in the first epoch. 
We run the hard negative mining for 2 epochs in total. The first epoch is for initializing the active set with the above heuristic, and the rest is for learning with the all the training data.  
Compared to the linear SVM training in R-CNN \citep{rcnn-cvpr}, our L-BFGS based solution to the structured objective costs $8\sim10$x longer time. 
However, it turns out to be significantly more efficient than SVM\textsuperscript{struct}~\citep{svm-struct}.

For the entire network finetuning, we initialize the structured loss layer with the weights obtained bythe classification-layer-only learning. 
The whole network is then finetuned by backpropagating the gradient from the top layer with a fixed SGD learning rate of $10^{-6}$.
For implementation simplicity, we keep updating the active sets until the end of an epoch, and update the classifiers per epoch (i.e., $\mathtt{update\_threshold}=+\infty$ in Algorithm~\ref{alg:hard-mining}). 
Like before, each category still has one particular active set. 
However, the network parameters (except for the classifier) are shared across all the category so that the  feature extraction time is not scaled up with the number of categories.  
In practice, we found one epoch was enough for both hard negative mining and SGD in the entire network finetuning case. 
Running more epochs did not make noticeable improvement on the final detection performance on PASCAL VOC 2007 test set, but cost a significantly larger amount of training time.

\section{Efficiency of fine-grained search (FGS)}
\label{sup:FGS-efficiency}

In this section, we provide more details on the local FGS presented in Algorithm~\ref{alg:FGS} of the main text. 

\textbf{GPR practical efficiency:} For the initial proposals given by selective search, $|\mathcal{D}_{\text{local}}|$ usually turns out to be 20 to 100, and line~\ref{alg-ln:latent},\ref{alg-ln:best} can be efficiently solved in around 9 and 6 L-BFGS~\citep{minfunc} iterations for StructObj, respectively.

\textbf{GPU parallelism for CNN:} One image can have multiple search regions (e.g., line 6), and 20 object categories together yield more regions.  FGS proposes one extra box per iteration for every search region. These boxes are fed into the CNN together to utilize GPU parallelism. For VGGNet, we use the batch size of 8 for computing the CNN features within the FGS procedure.

\begin{figure}[h]
\centering
\includegraphics[width=0.5\columnwidth]{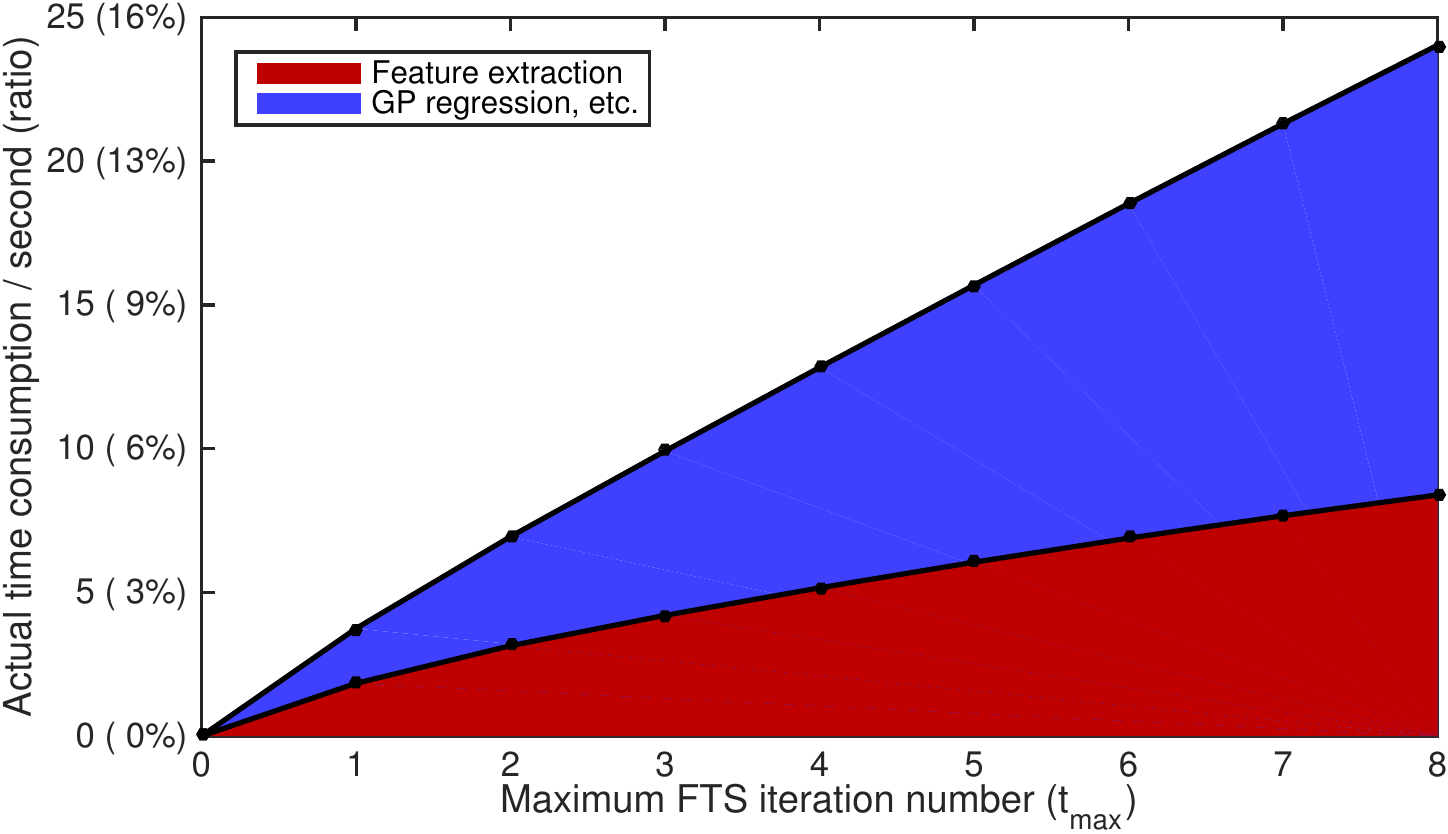}
\protect\caption{Time cost with different maximum number of iterations. The ``ratio'' is with respect to the time cost of extracting features for the initial bounding boxes.}
\label{fig:gp-iter-time}
\end{figure}

\textbf{Time overhead:} For PASCAL VOC 2007, FGS ($t_{\text{max}}=8$) induced only $\sim 15\%$ total overhead compared to initial time cost, which mostly consists of CNN feature extraction from bounding boxes proposed by selective search (SS). Specifically, $1/3$ of the overhead is caused by CNN feature extraction from the newly proposed boxes (line 11); the rest is caused by GPR (line9, 10), NMS (line 5), and pruning (line 12). Each GP iteration (line 2-16) counts for $\sim 2\%$ with respect to the initial time cost, and $\leq 8$ GP iterations were sufficient for convergence. Figure~\ref{fig:gp-iter-time} shows the trends of the accumulated time overhead introduced by FGS per iteration. 
The time overhead due to FGS may vary with different datasets (e.g., VOC 2012), but in general, it is $\le20\%$ compared to initial time cost. 

\section{Step-wise performance of fine-grained search (FGS)}
\label{sup:per-iter-FGS}
We evaluated the mAP at each GP iteration using R-CNN(VGG)+StructObj+FGS+BBoxReg. The mAPs from 0 to 8 GP iterations are reported in Table~\ref{tab:gp-stepwise}. mAP increases rapidly in the first 4 iterations, and becomes stable in the following iterations. 

\begin{table}[htbp]
\centering
\begin{tabular}{l|c|c|c|c|c|c|c|c|c}
\hline
\# GP iter & 0 & 1 & 2 & 3 & 4 & 5 & 6 & 7 & 8\\
\hline
mAP & 66.6 & 67.5 & 67.8 & 68.2 & 68.3 & 68.6 & 68.4 & 68.6 & 68.5 \\
\hline
\end{tabular}
\caption{Test set mAPs on PASCAL VOC 2007 for ``R-CNN(VGG)+StructObj+FGS+BBoxReg'' with different number of GP iterations \label{tab:gp-stepwise}}
\end{table}

\begin{table}
\end{table}

\section{Test set mAP on PASCAL VOC 2007 using VGGNet with different region proposal methods}
\label{sup:mAP-diff-pr}

In Figure 2 of the paper, we report and compare the test set mAPs on PASCAL VOC 2007 with different region proposal algorithms (e.g., selective search (SS)~\cite{selective-search} at different modes, Objectness~\cite{objectness}) using an oracle detector.
In this section, we performed similar experiments using a real detector trained with structured SVM objective based on VGGNet features.
The summary results are given in Table~\ref{tab:map_voc2007_vggnet} and ~\ref{tab:map_voc2007_vggnet_bbox}.

\begin{table}[htbp]
\centering
\begin{tabular}{l|r|r|r|r|r|r|r|r}
\hline 
Region proposal methods $\backslash$ IoU threshold & $0.1$ & $0.2$ & $0.3$ & $0.4$ & $0.5$ & $0.6$ & $0.7$ & $0.8$\\
\hline 
SS ($\sim$2000 boxes per image) & $74.3$  & $73.8$  & $72.5$  & $69.6$  & $61.2$  & $47.4$  & $31.2$  & $15.4$  \\
SS + Objectness ($\sim$3000 boxes per image) & $73.4$ & $72.9$ & $71.7$ & $68.9$ & $60.6$ & $47.0$ & $31.1$ & $15.3$ \\
SS extended ($\sim$3500 boxes per image) & $74.3$  & $73.9$  & $72.8$  & $70.1$  & $62.8$  & $48.5$  & $32.3$  & $15.9$  \\
SS quality ($\sim$10000 boxes per image) & $74.1$  & $73.7$  & $72.7$  & $70.0$  & $63.7$  & $51.4$  & $35.7$  & $\mathbf{17.9}$ \\
SS + FGS ($\sim$2150 boxes per image) & $\mathbf{76.6}$ & $\mathbf{76.1}$ & $\mathbf{75.0}$ & $\mathbf{72.4}$ & $\mathbf{65.8}$ & $\mathbf{54.1}$ & $\mathbf{37.2}$ & ${17.4}$ \\
\hline 
\end{tabular}
\caption{Test set mAPs on PASCAL VOC 2007 with different region proposal methods at varying IoU thresholds from $0.1$ to $0.8$ without bounding box regression.\label{tab:map_voc2007_vggnet}}
\end{table}

\begin{table}[htbp]
\centering
\begin{tabular}{l|r|r|r|r|r|r|r|r}
\hline 
Region proposal methods $\backslash$ IoU threshold & $0.1$ & $0.2$ & $0.3$ & $0.4$ & $0.5$ & $0.6$ & $0.7$ & $0.8$ \\
\hline 
SS ($\sim$2000 boxes per image) & $77.4$  & $76.9$  & $75.9$  & $73.4$  & $66.5$  & $55.8$  & $40.9$  & $19.4$  \\
SS + Objectness ($\sim$3000 boxes per image) & $76.9$ & $76.5$ & $75.5$ & $73.0$ & $66.0$ & $55.2$ & $40.4$ & $19.3$ \\
SS extended ($\sim$3500 boxes per image) & $77.6$  & $77.2$  & $76.2$  & $73.8$  & $67.6$  & $57.0$  & $41.8$  & $19.7$  \\
SS quality ($\sim$10000 boxes per image) & $77.1$  & $76.7$  & $75.8$  & $73.3$  & $67.1$  & $57.0$  & $42.3$  & $19.4$  \\
SS + FGS ($\sim$2150 boxes per image) & $\mathbf{78.3}$ & $\mathbf{77.8}$ & $\mathbf{76.8}$ & $\mathbf{74.4}$ & $\mathbf{68.5}$ & $\mathbf{57.9}$ & $\mathbf{43.1}$ & $\mathbf{20.4}$ \\
\hline 
\end{tabular}
\caption{Test set mAPs on PASCAL VOC 2007 with different region proposal methods at varying IoU thresholds from $0.1$ to $0.8$ with bounding box regression.\label{tab:map_voc2007_vggnet_bbox}}
\end{table}

For both cases (with and without bounding box regression), the FGS showed improved performance over other region proposal methods using smaller number of region proposals.
In particular, the SS + FGS method (row 5 in Table~\ref{tab:map_voc2007_vggnet} and~\ref{tab:map_voc2007_vggnet_bbox}) even outperformed the SS ``quality'' mode~\cite{selective-search}, which requires $\sim$ $5\times$ more computational expenses than our proposed method to compute CNN-based detection scores for bounding box proposals.

Although the current state-of-the-art CNN-based detector outperforms other object detection methods~\cite{dpm-pami,oc-dpm,struct-localization} by a large margin, there still remains a significant gap with that of the hypothetical oracle detector.
This motivates us to further research on improving the quality of the CNNs for better visual object recognition performance.

\vfill

\section{Precision-recall curves on PASCAL VOC 2007}
\label{sup:pr-curves}

\begin{figure}[H]
\centering

\includegraphics[width=0.24\textwidth]{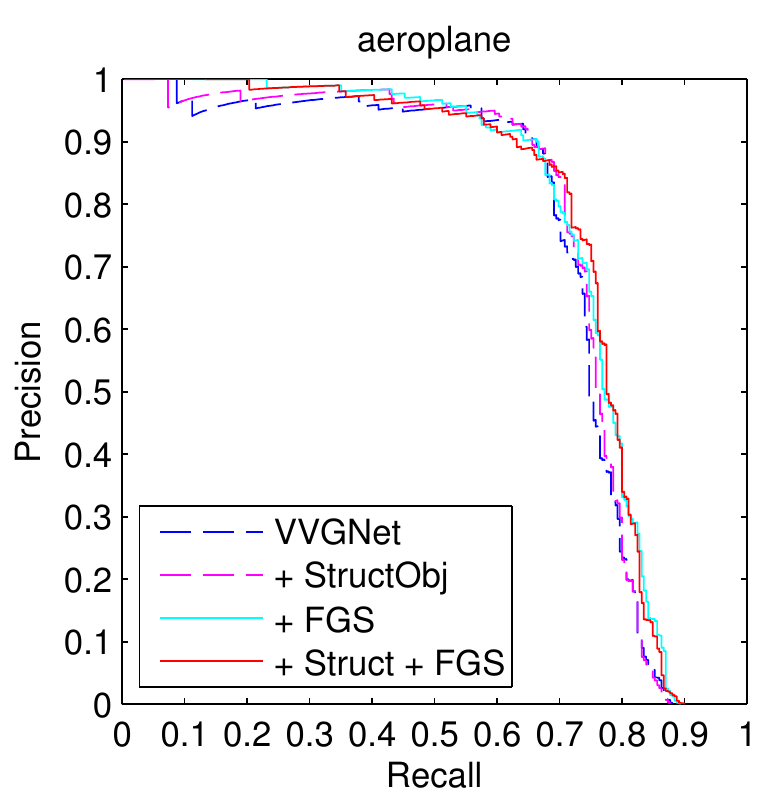}\includegraphics[width=0.24\textwidth]{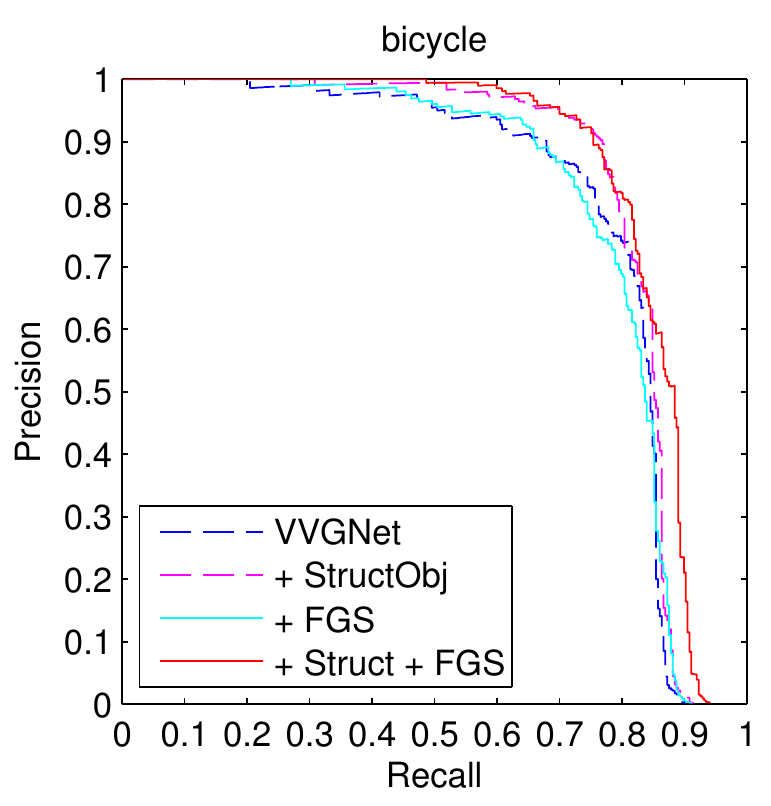}\includegraphics[width=0.24\textwidth]{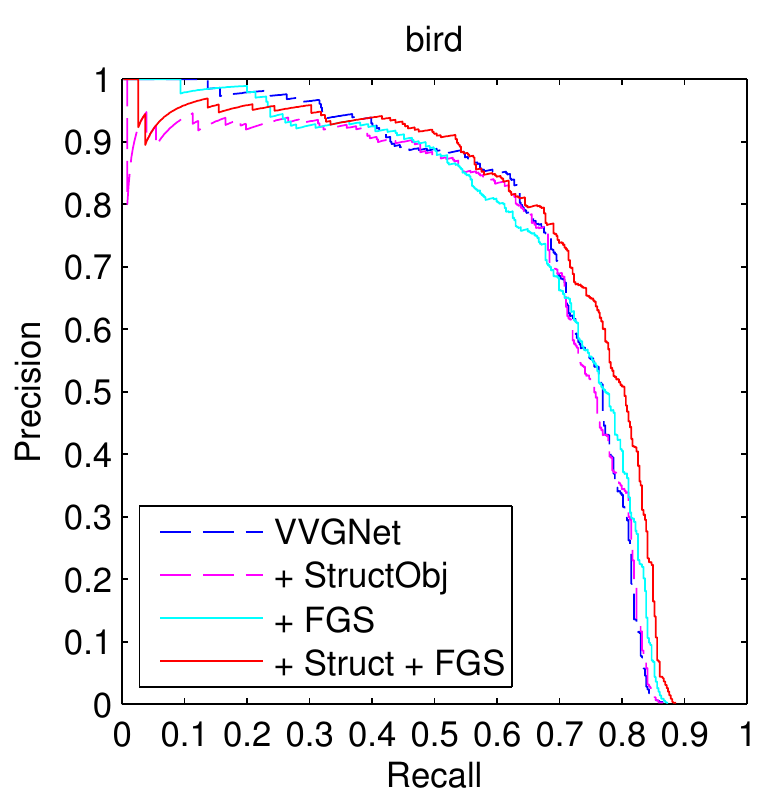}\includegraphics[width=0.24\textwidth]{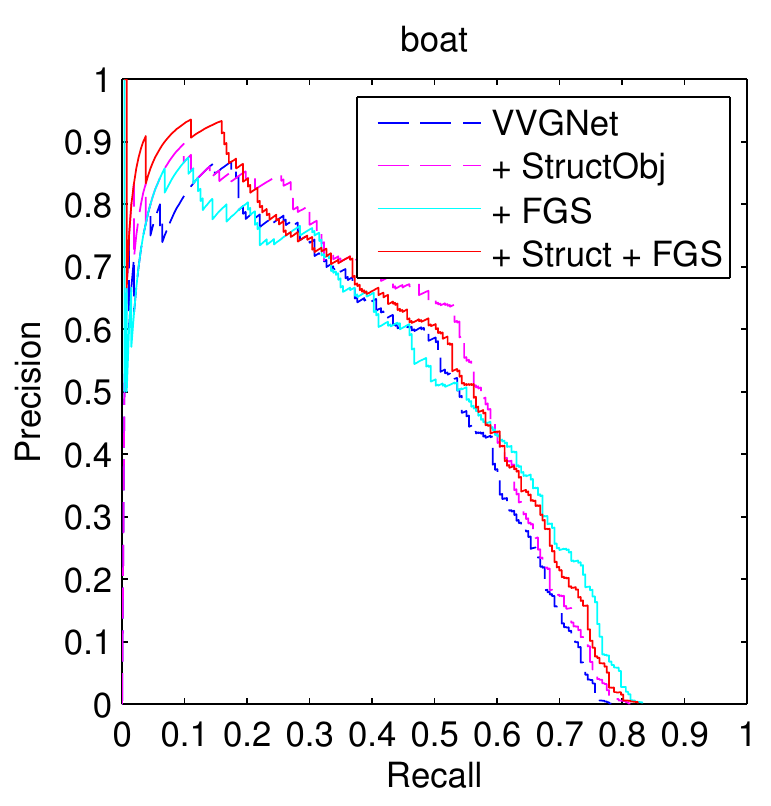}\\

\includegraphics[width=0.24\textwidth]{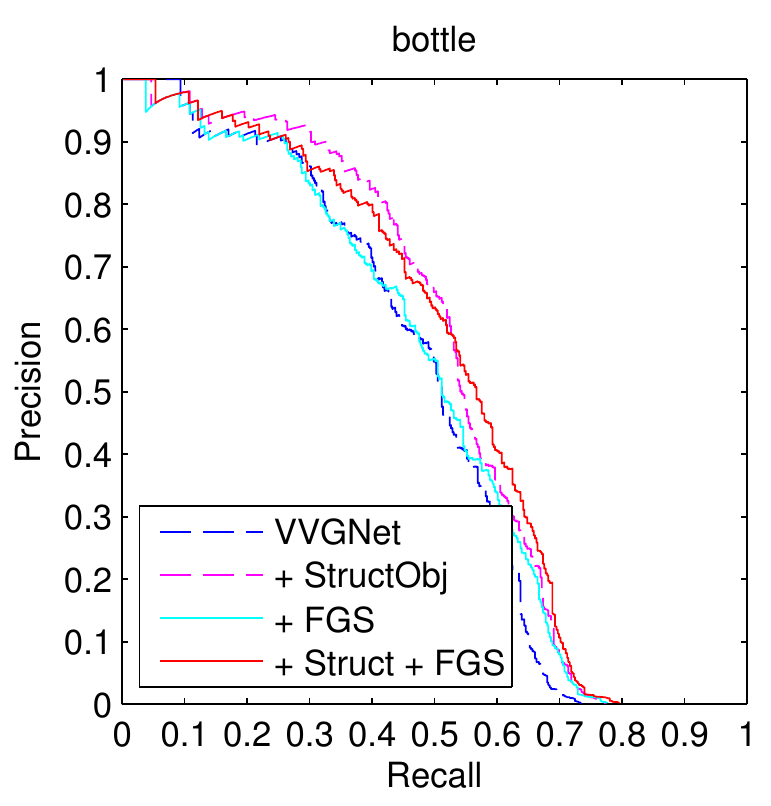}\includegraphics[width=0.24\textwidth]{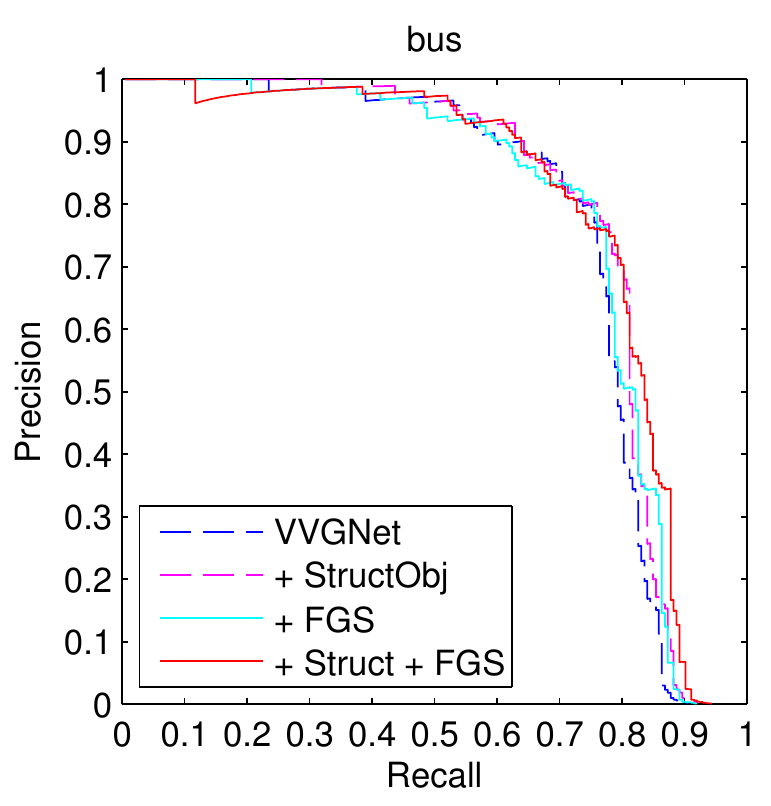}\includegraphics[width=0.24\textwidth]{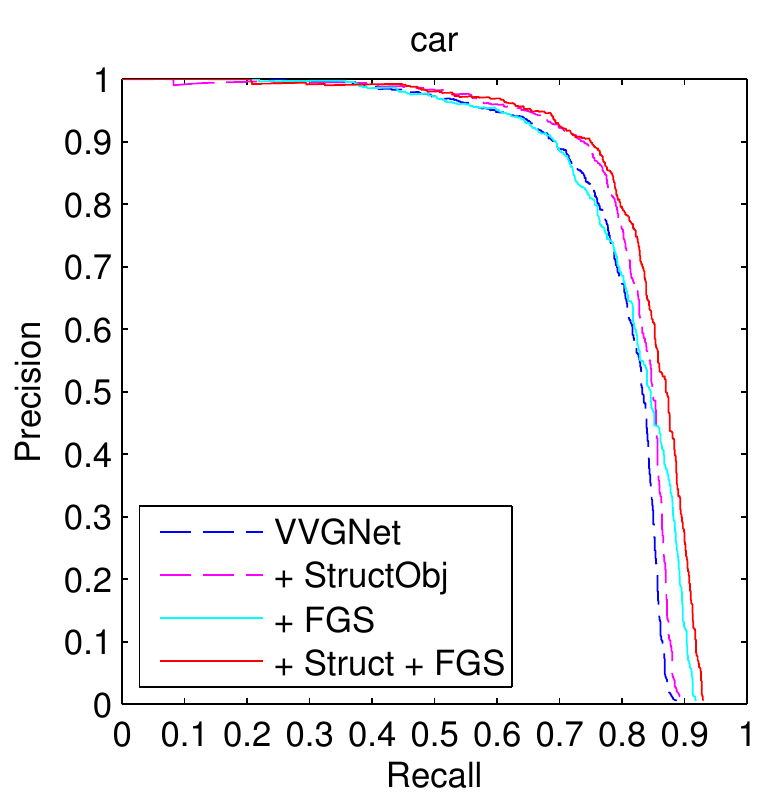}\includegraphics[width=0.24\textwidth]{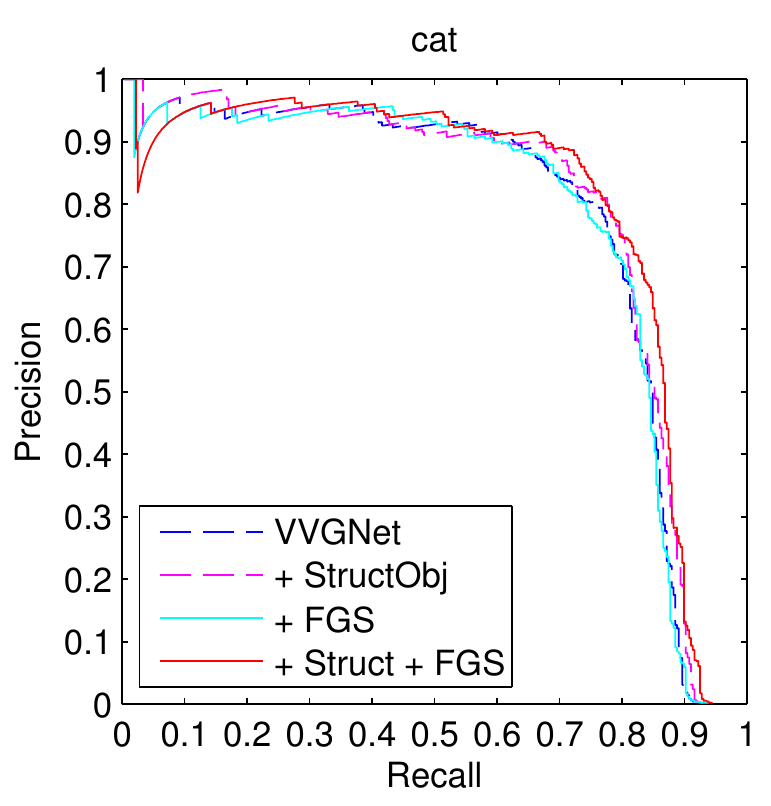}\\

\includegraphics[width=0.24\textwidth]{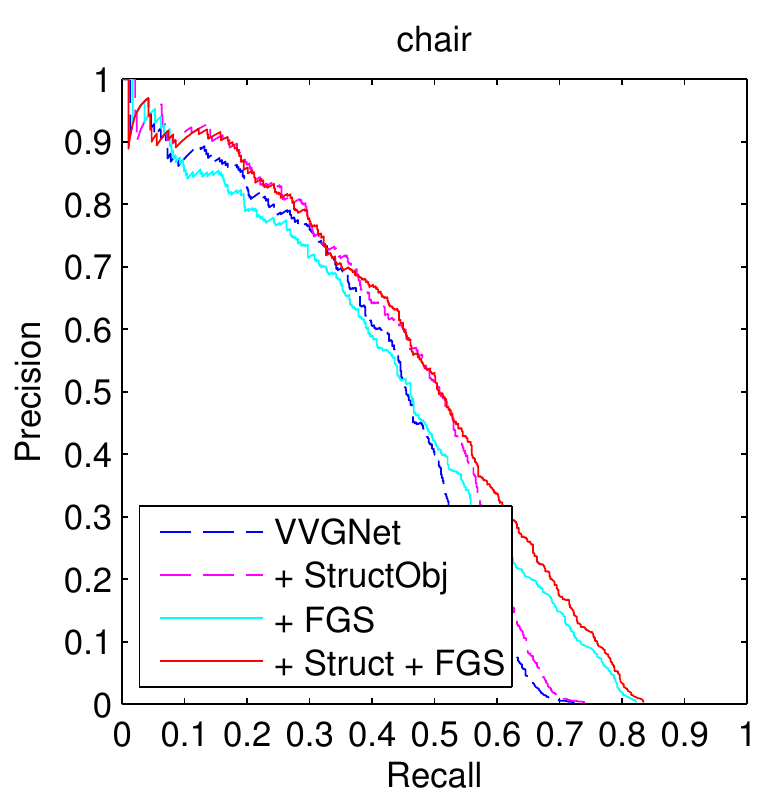}\includegraphics[width=0.24\textwidth]{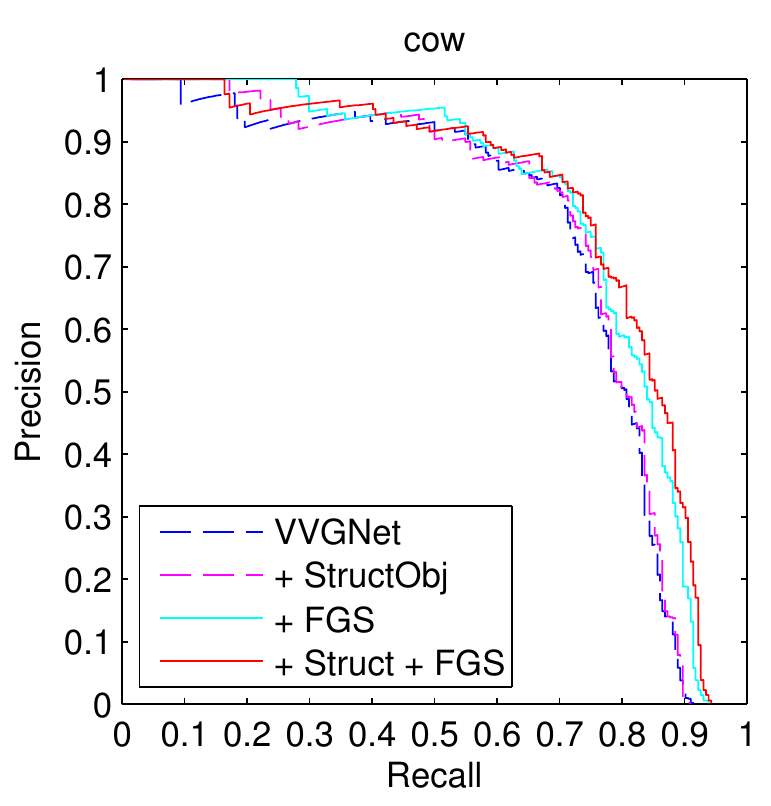}\includegraphics[width=0.24\textwidth]{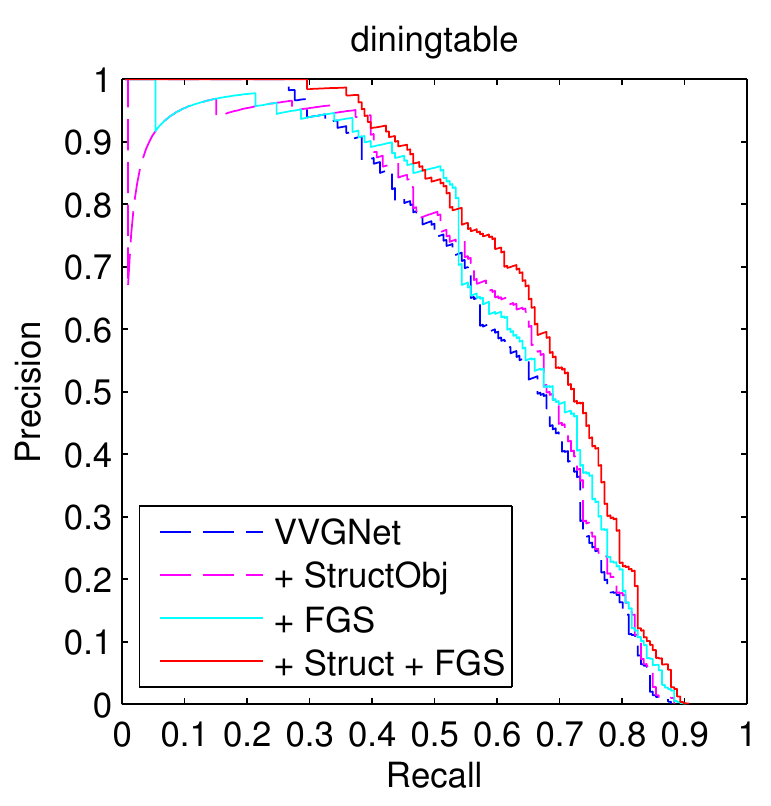}\includegraphics[width=0.24\textwidth]{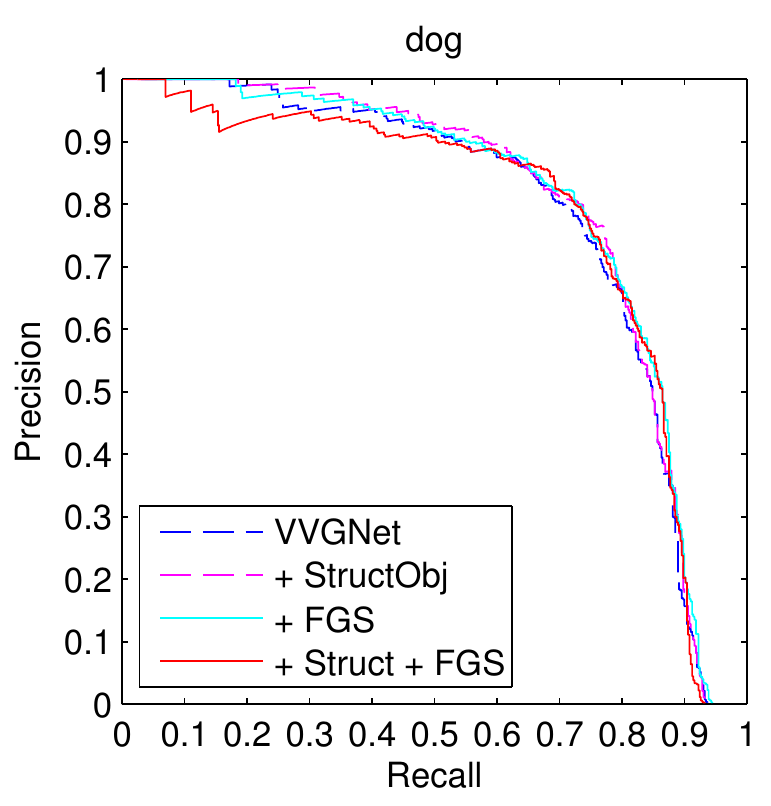}\\

\includegraphics[width=0.24\textwidth]{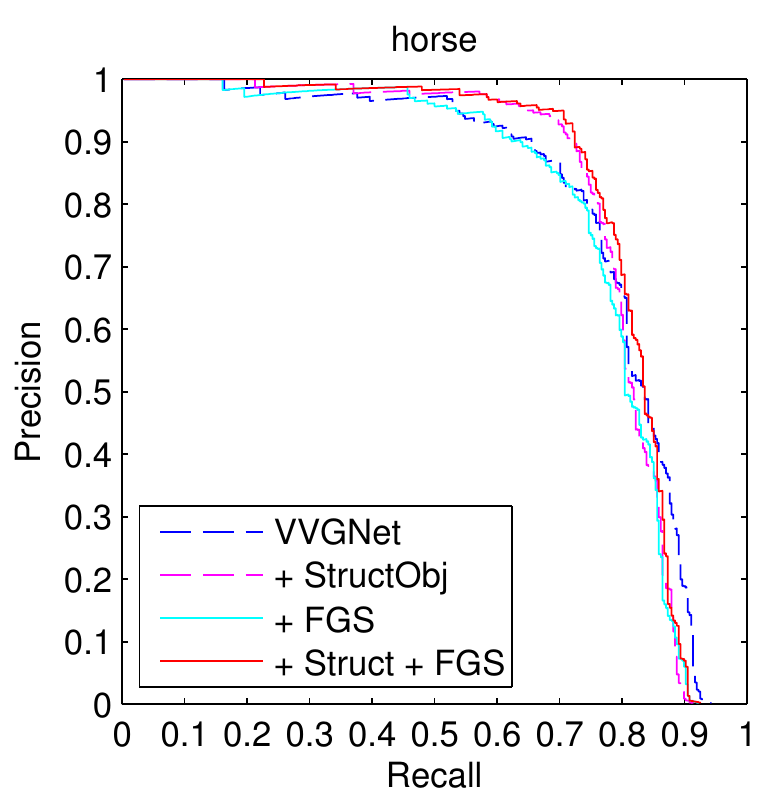}\includegraphics[width=0.24\textwidth]{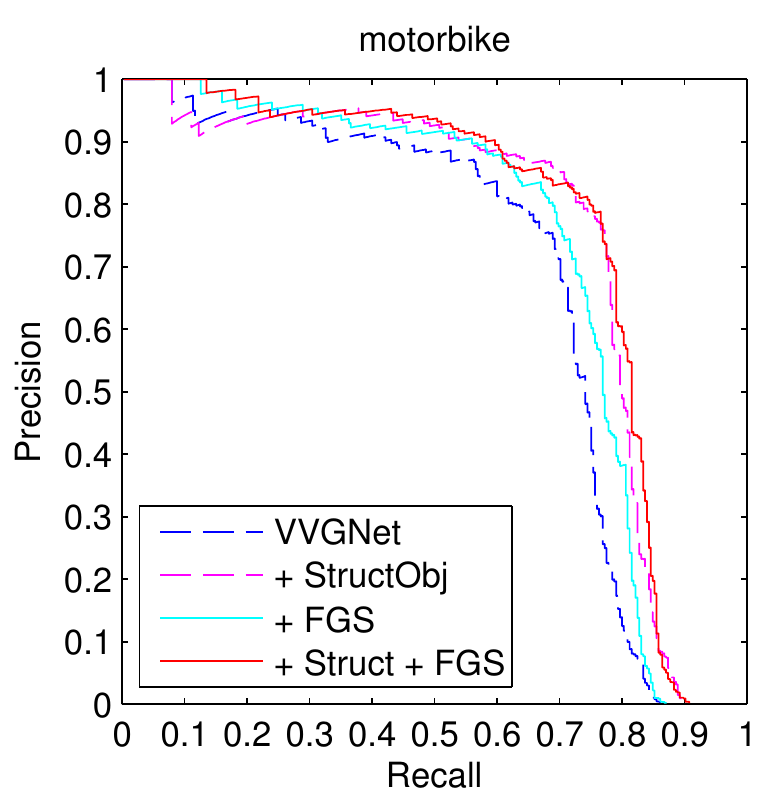}\includegraphics[width=0.24\textwidth]{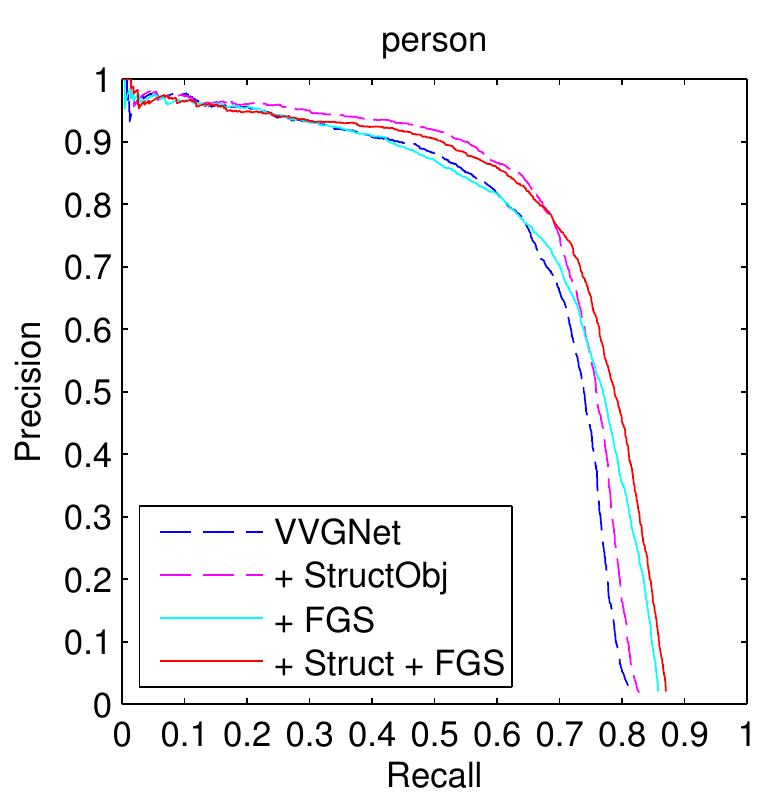}\includegraphics[width=0.24\textwidth]{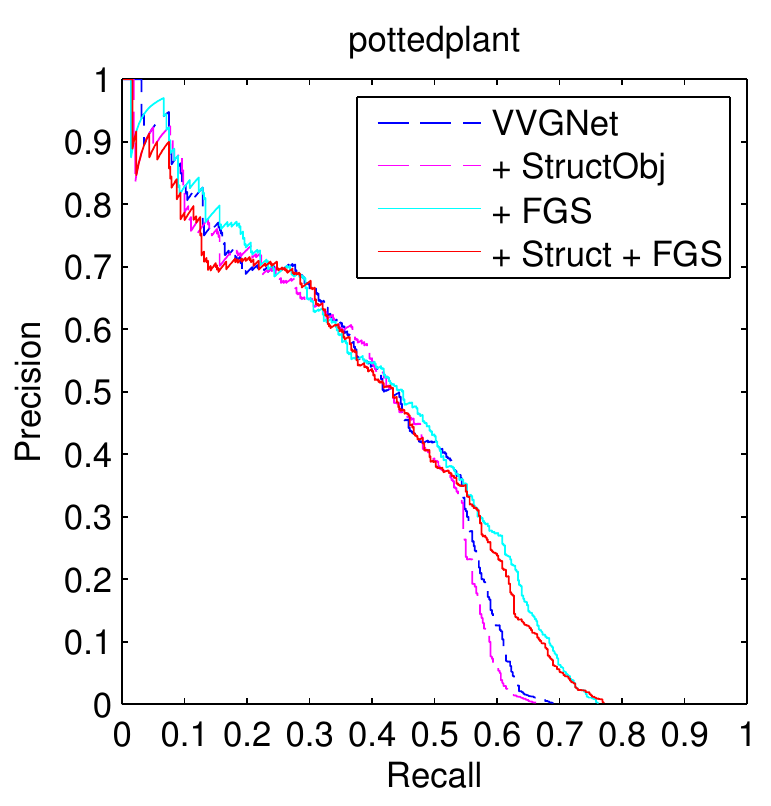}\\

\includegraphics[width=0.24\textwidth]{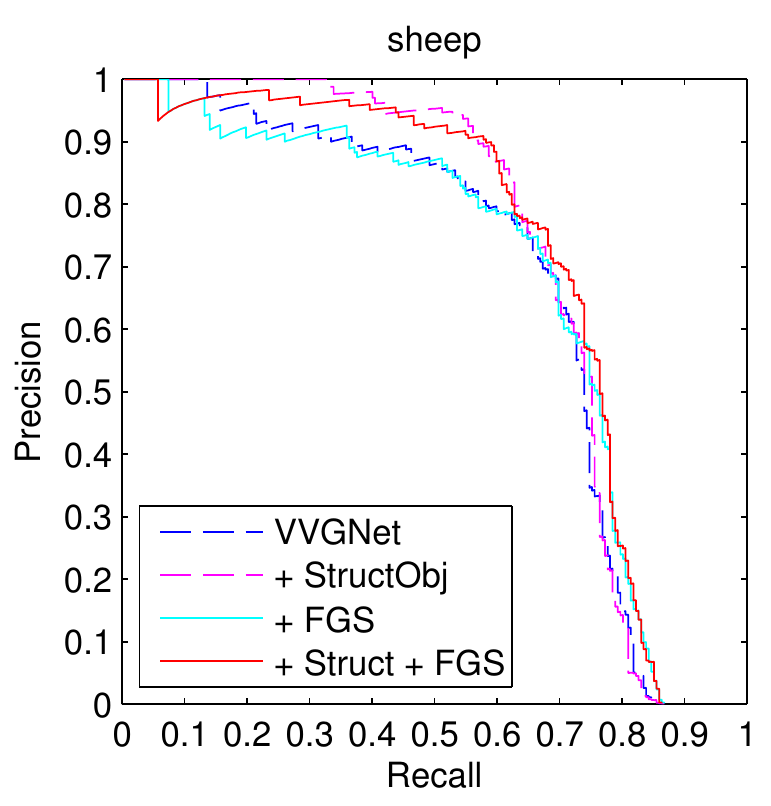}\includegraphics[width=0.24\textwidth]{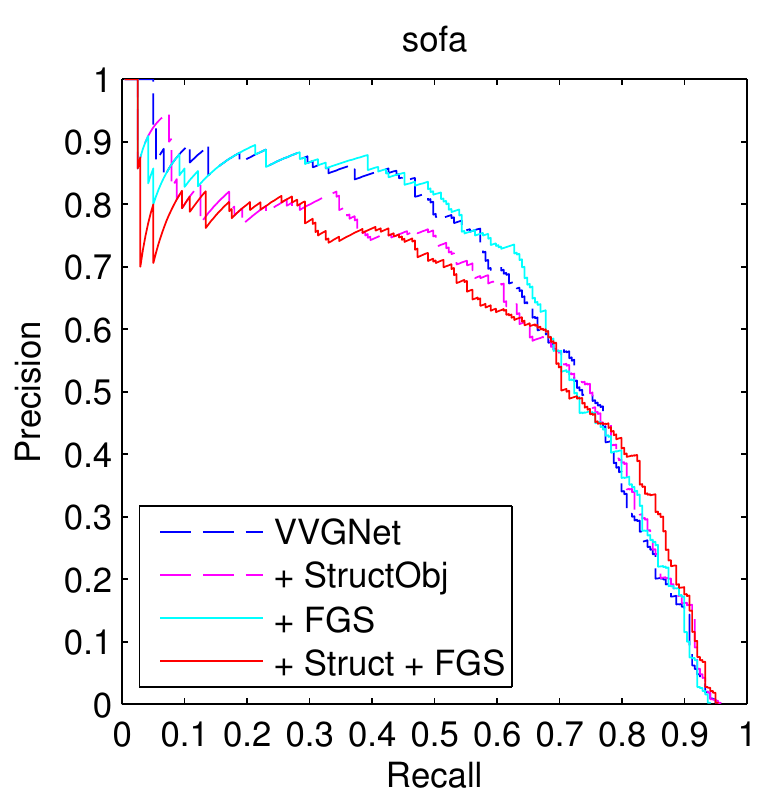}\includegraphics[width=0.24\textwidth]{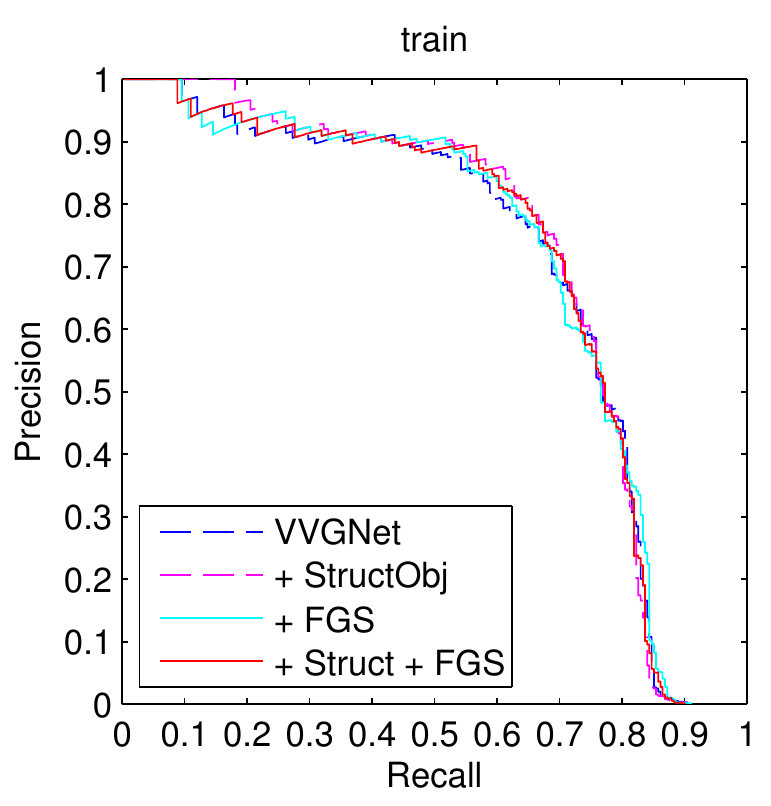}\includegraphics[width=0.24\textwidth]{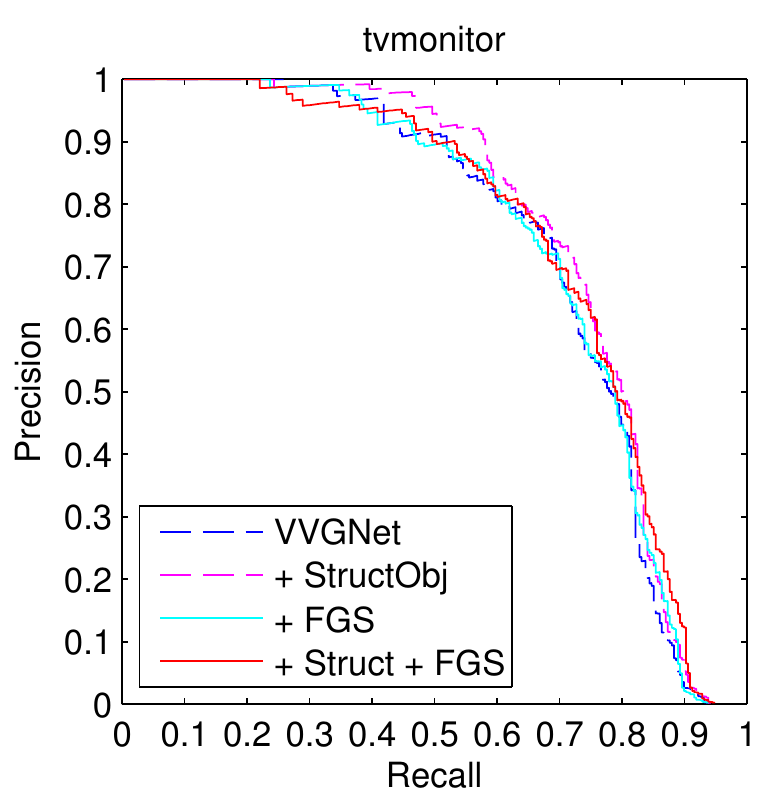} 

\caption{Category-wise precision-recall curves for four different models (VGGNet, + StructObj, + FGS, + StructObj + FGS) on PASCAL VOC 2007 database. \label{fig:pr-voc2007}}

\end{figure}

In this section, we present the precision-recall curves for four different models.
Specifically, we show results for VGGNet, VGGNet trained with structured SVM objective (VGGNet + StructObj), VGGNet with FGS (VGGNet + FGS), and VGGNet with both (VGGNet + StructObj + FGS) in Figure~\ref{fig:pr-voc2007}.
In general, the improvement from the structured SVM objective is more significant for the high recall range (i.e., recall $\geq0.5$) than the low recall range other than ``sheep'' class.
FGS usually improves the precision for most object categories.

\vfill

\newpage

\section{Localization accuracy on PASCAL VOC 2007}
\label{sup:loc-distr}
\begin{figure}[H]
\centering
\hfill
\includegraphics[width=0.3\textwidth]{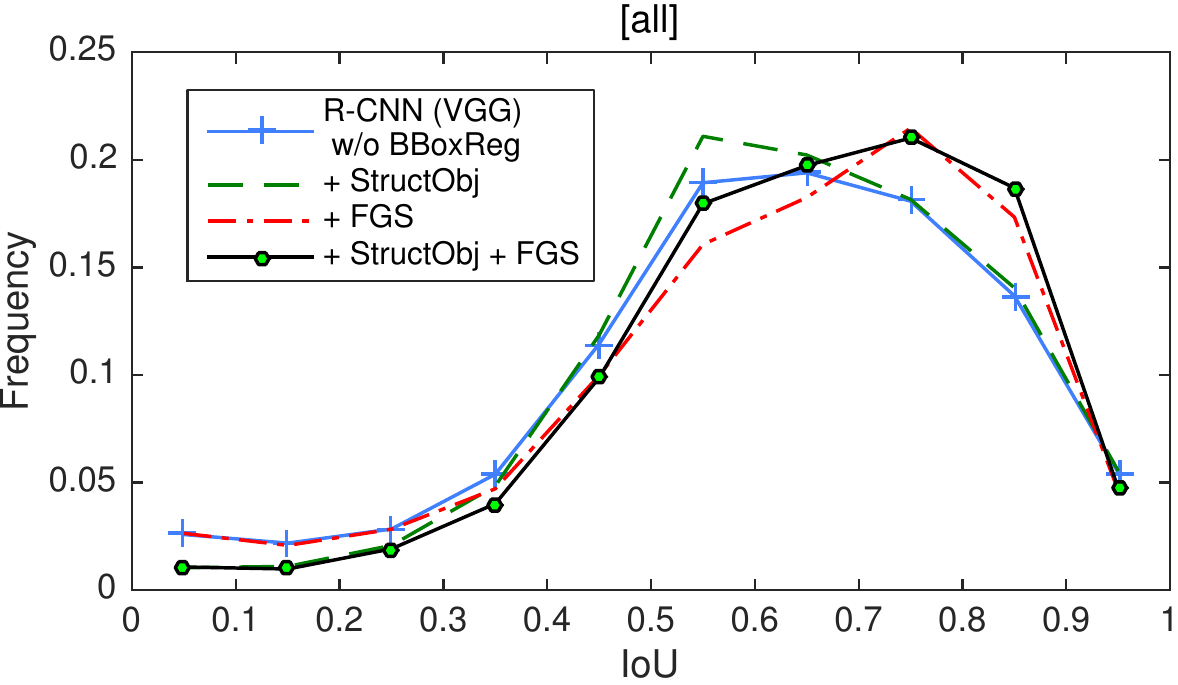}\hfill
\includegraphics[width=0.3\textwidth]{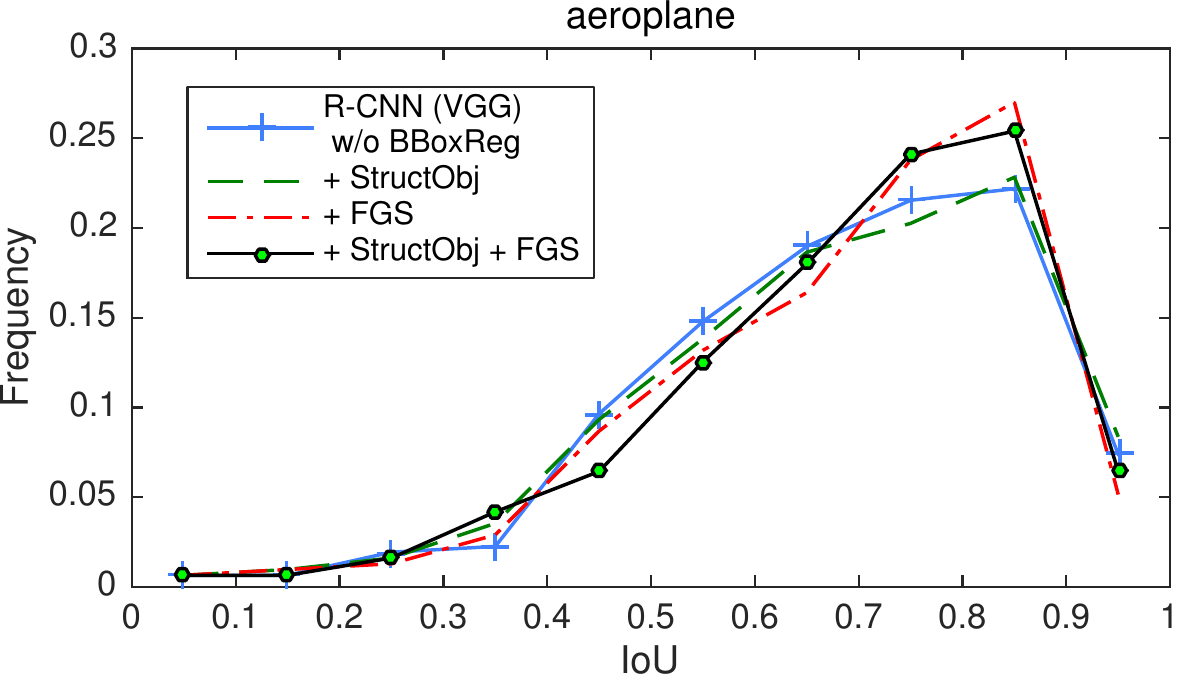}\hfill
\includegraphics[width=0.3\textwidth]{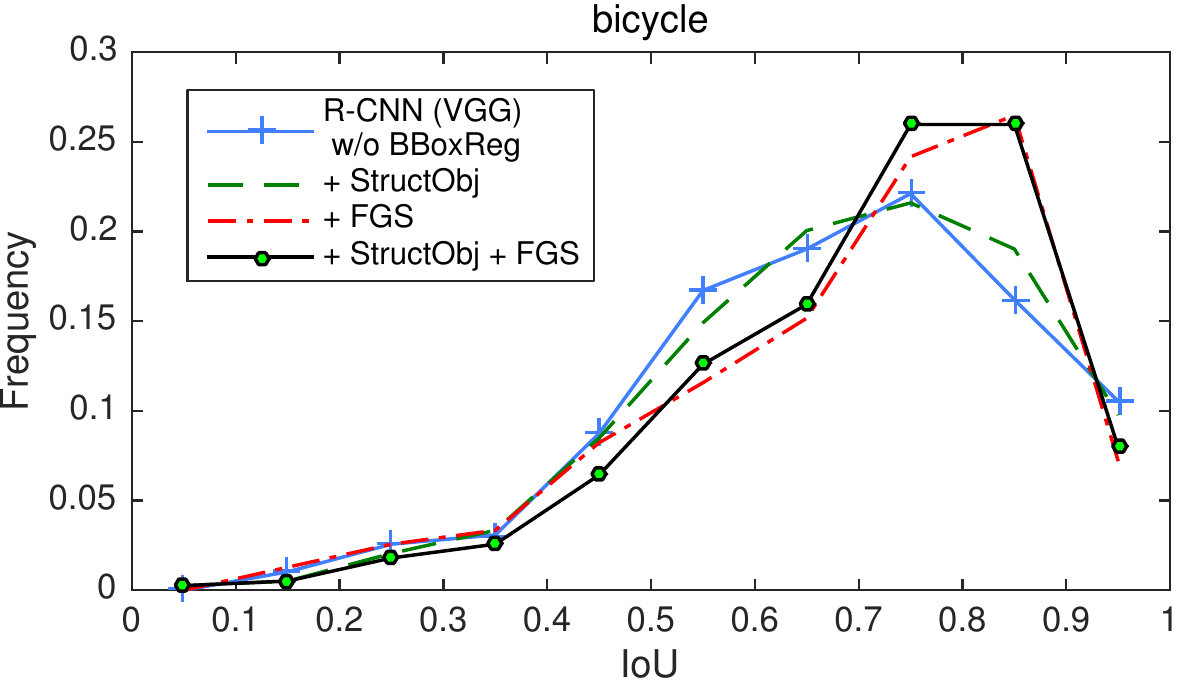}\hfill \\

\hfill
\includegraphics[width=0.3\textwidth]{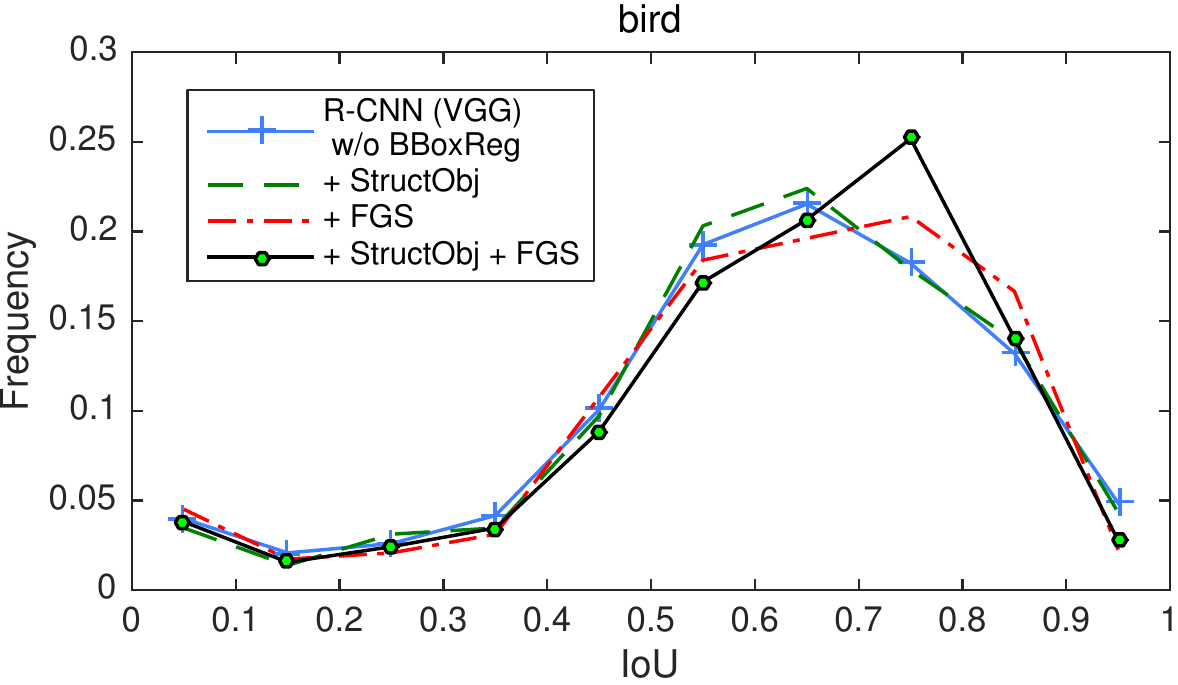}\hfill
\includegraphics[width=0.3\textwidth]{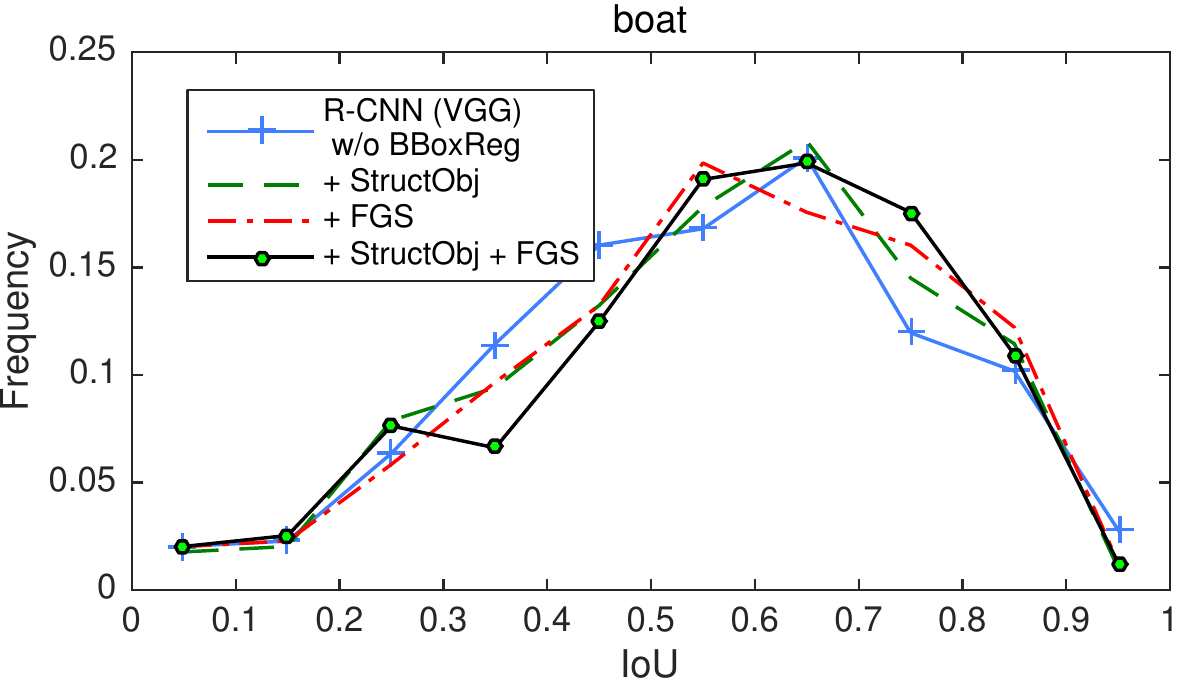}\hfill
\includegraphics[width=0.3\textwidth]{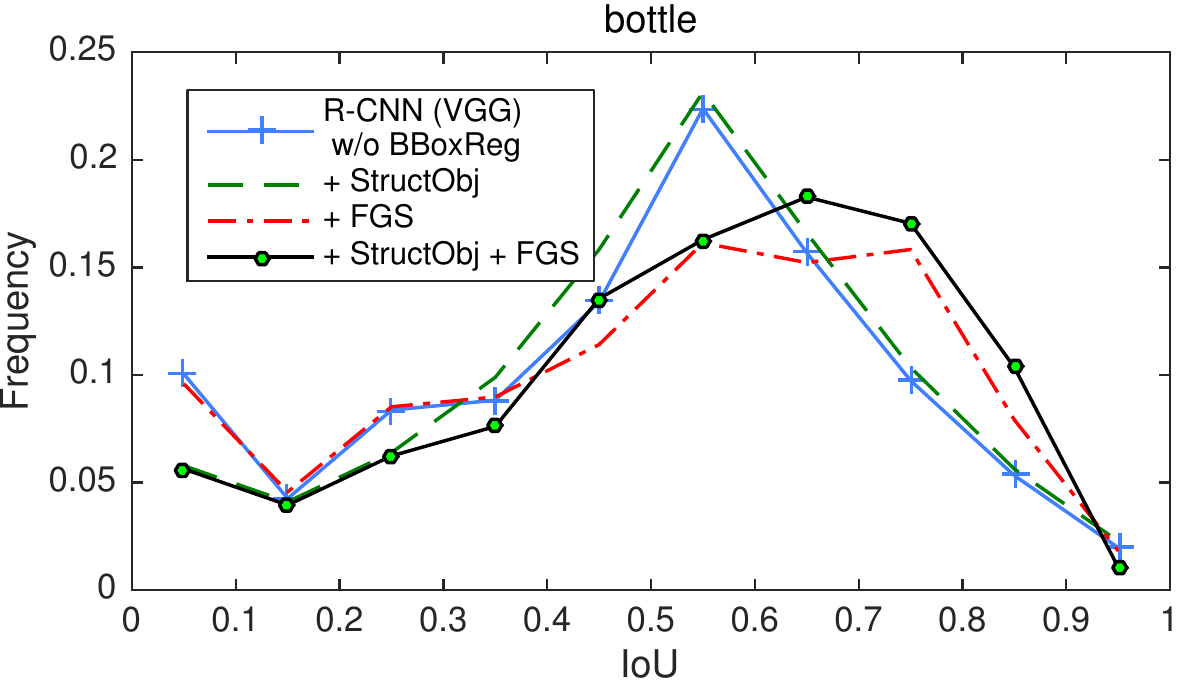}\hfill \\

\hfill
\includegraphics[width=0.3\textwidth]{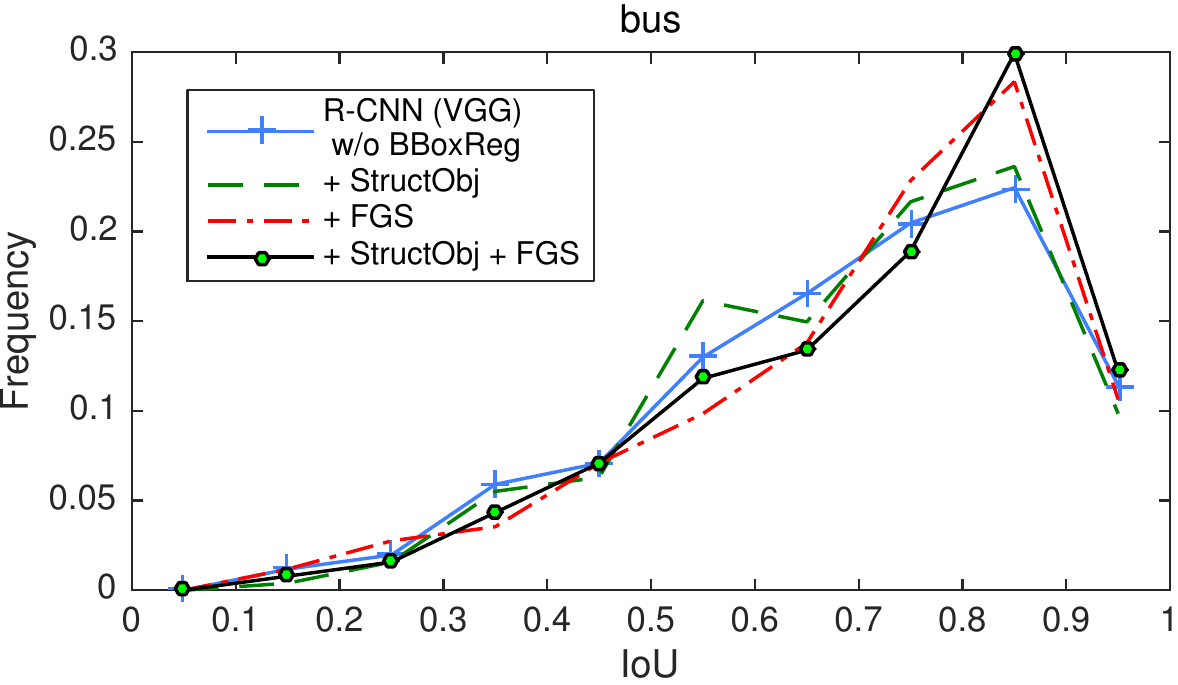}\hfill
\includegraphics[width=0.3\textwidth]{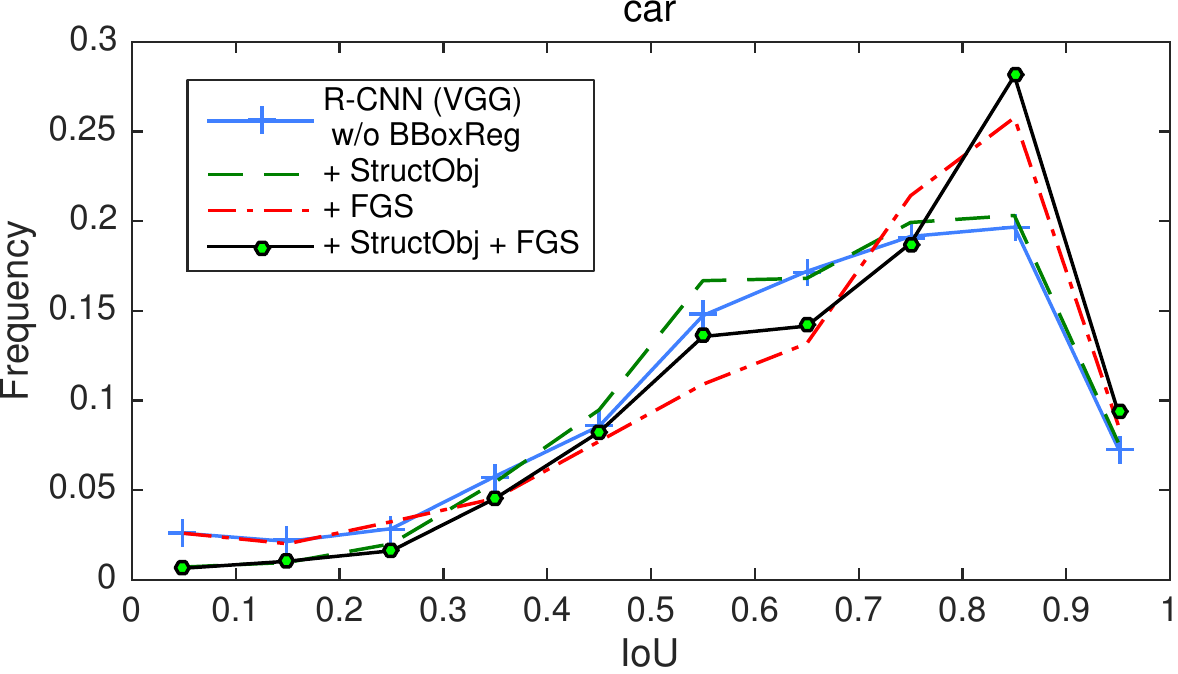}\hfill
\includegraphics[width=0.3\textwidth]{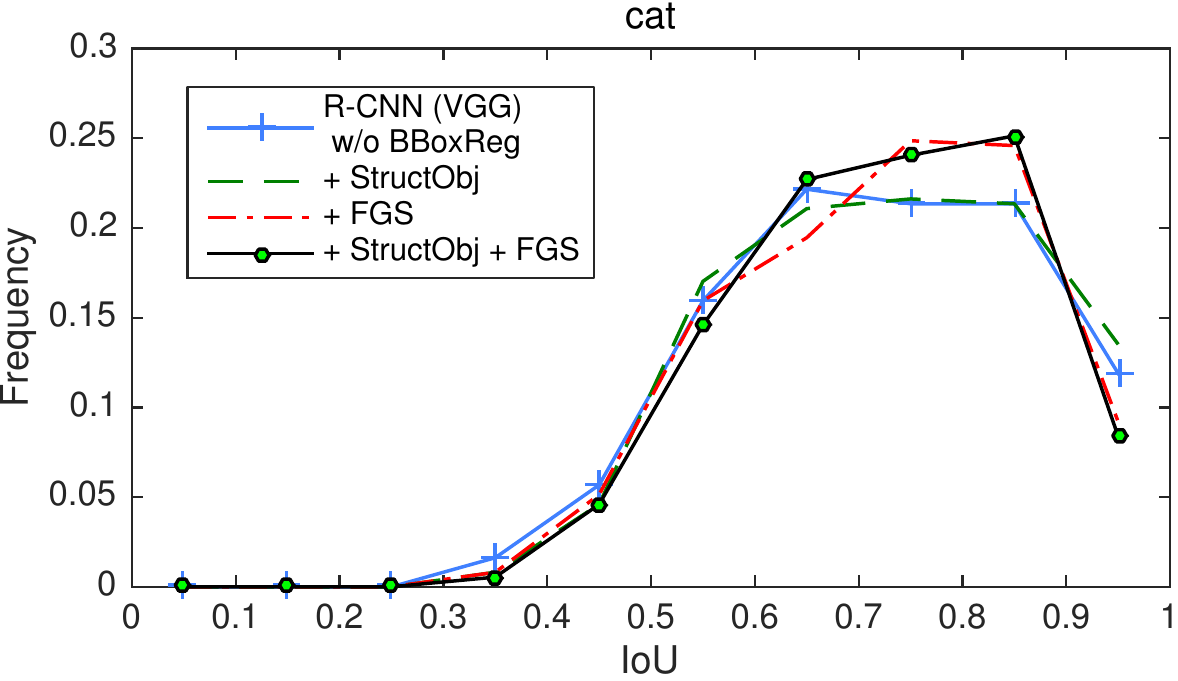}\hfill \\

\hfill
\includegraphics[width=0.3\textwidth]{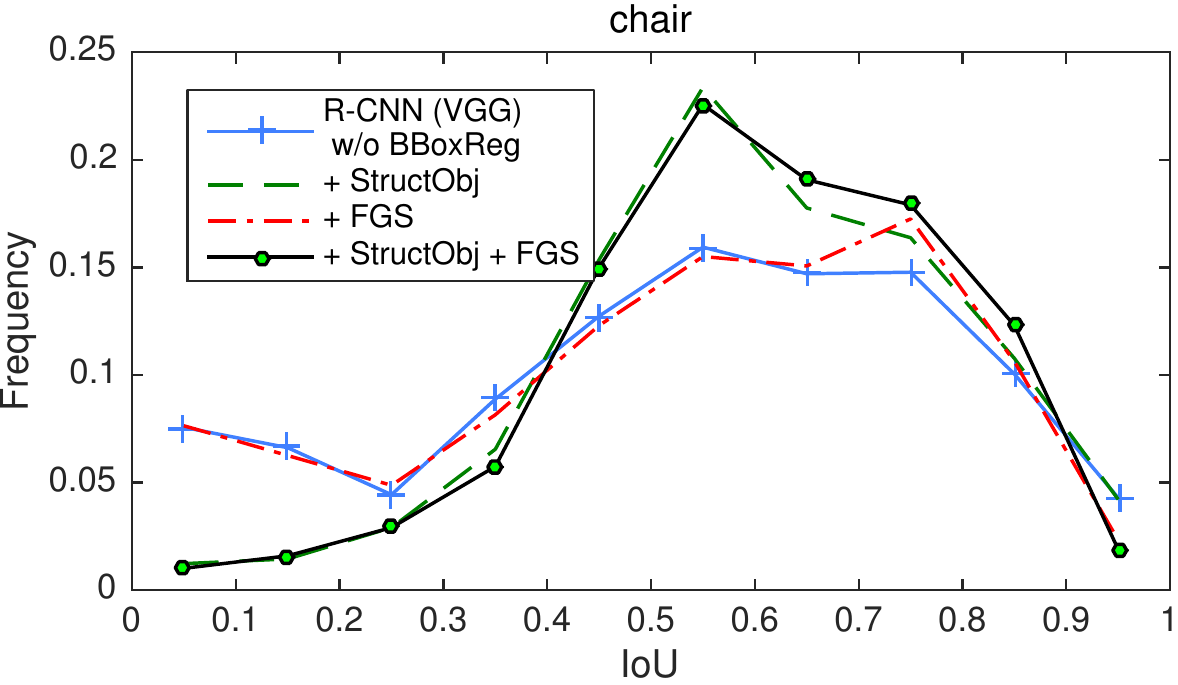}\hfill
\includegraphics[width=0.3\textwidth]{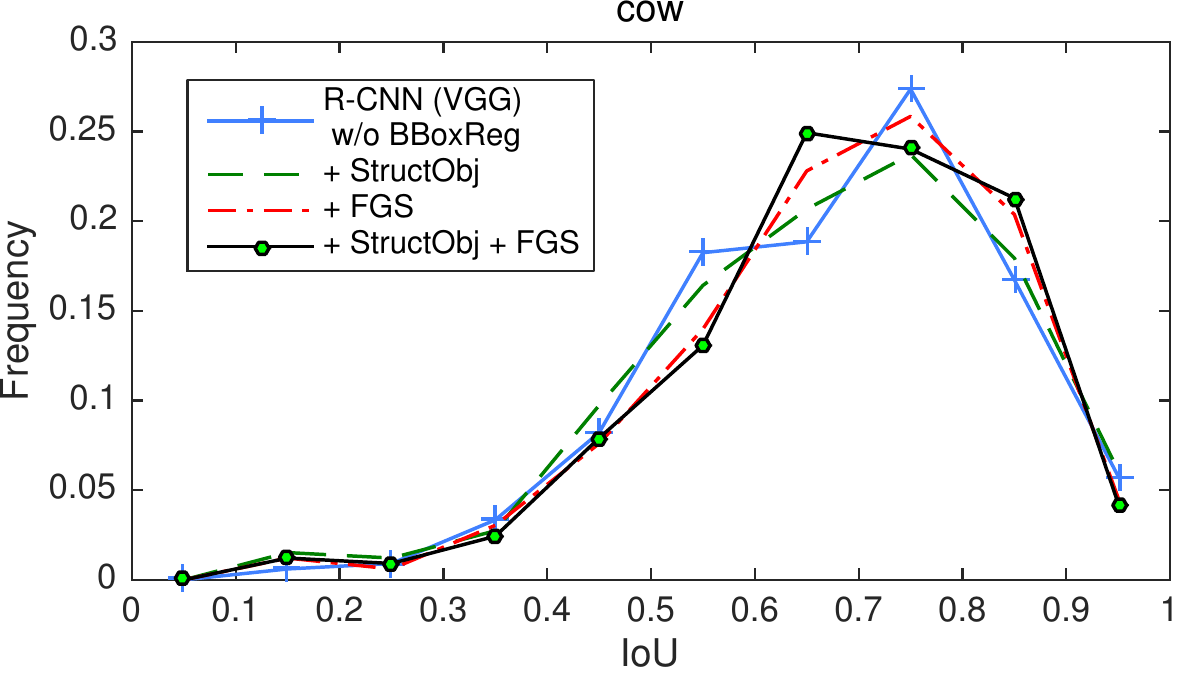}\hfill
\includegraphics[width=0.3\textwidth]{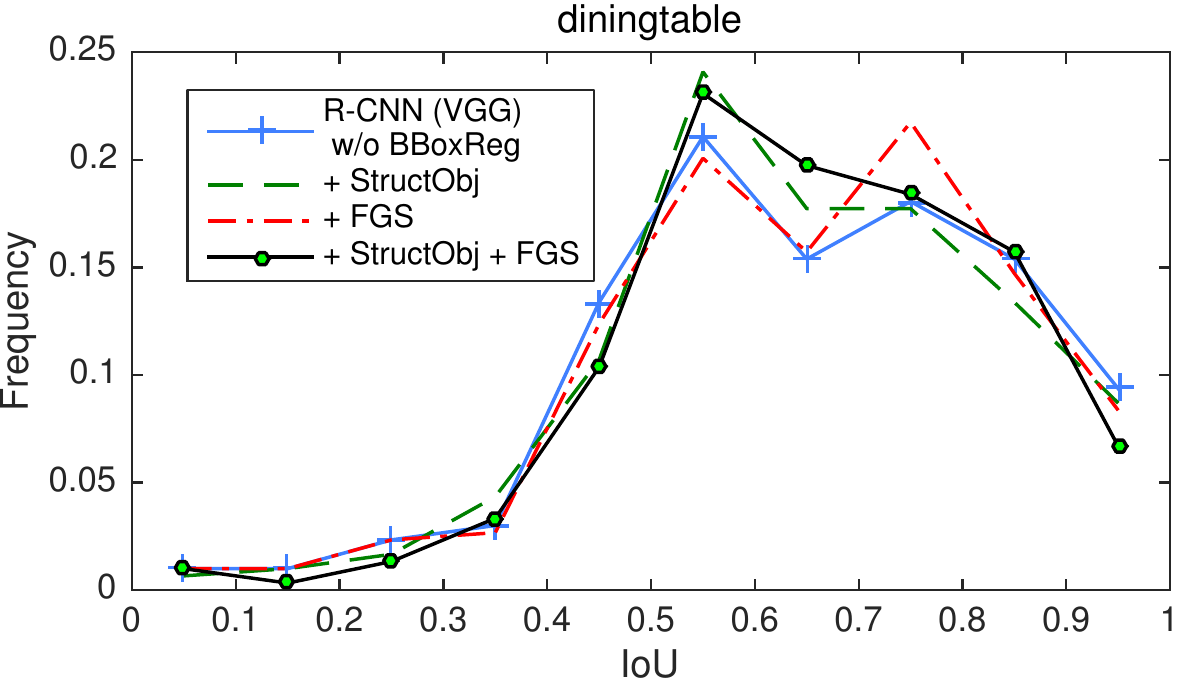}\hfill \\

\hfill
\includegraphics[width=0.3\textwidth]{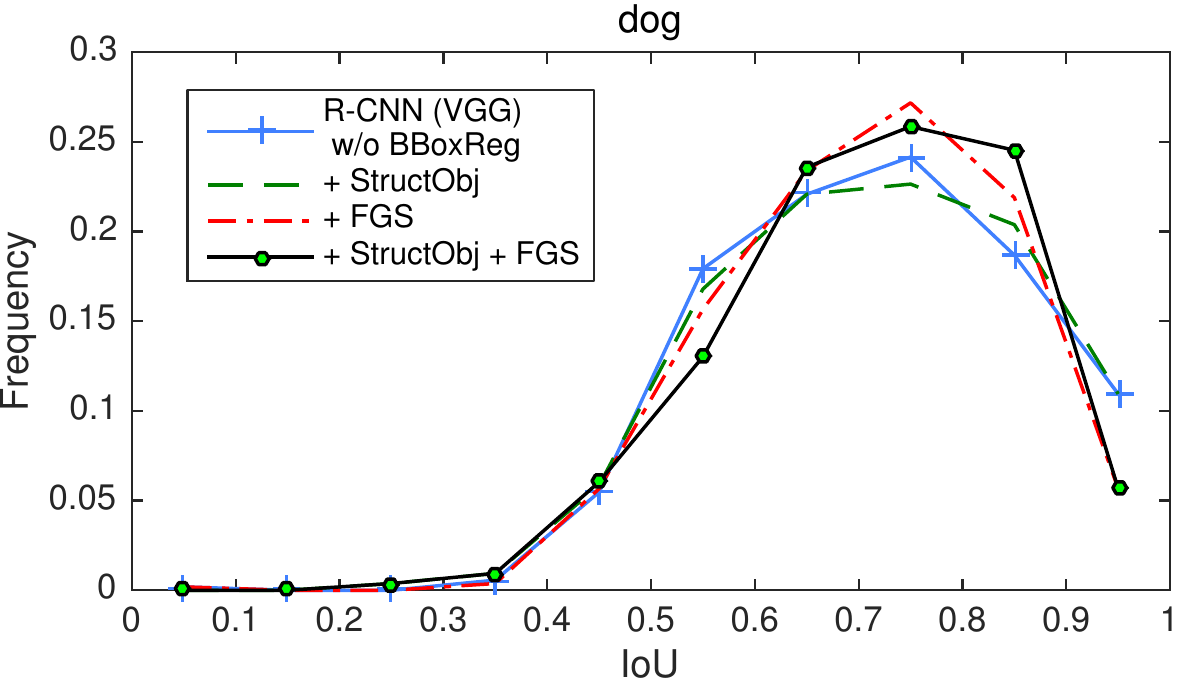}\hfill
\includegraphics[width=0.3\textwidth]{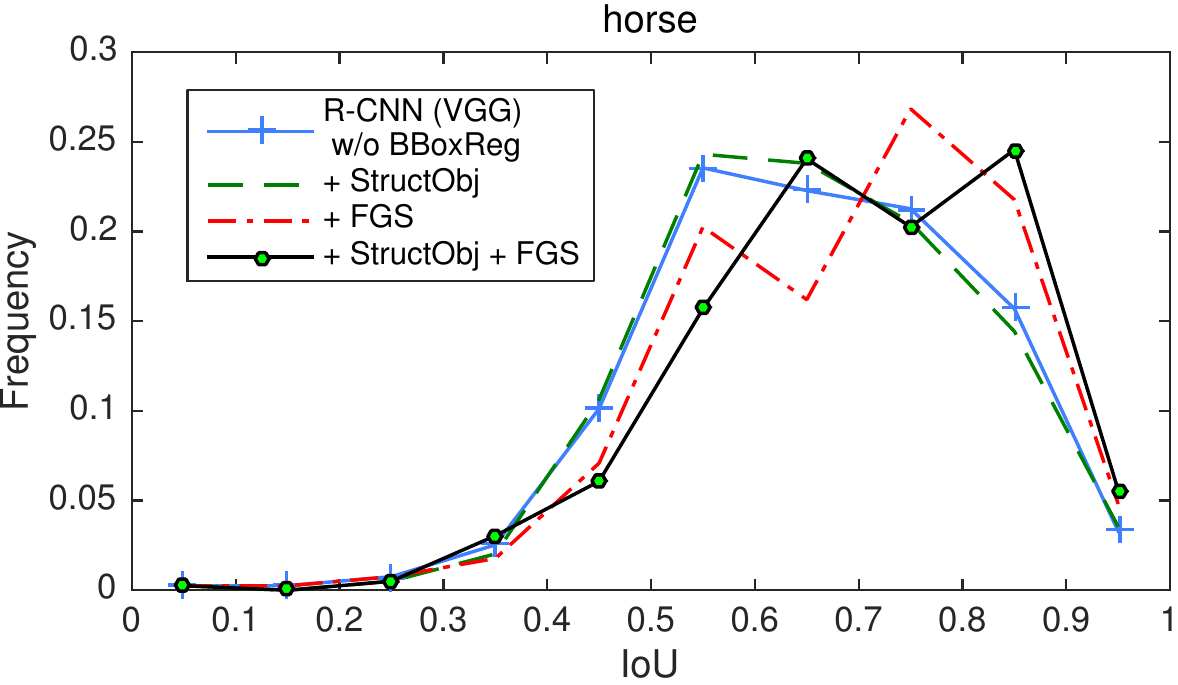}\hfill
\includegraphics[width=0.3\textwidth]{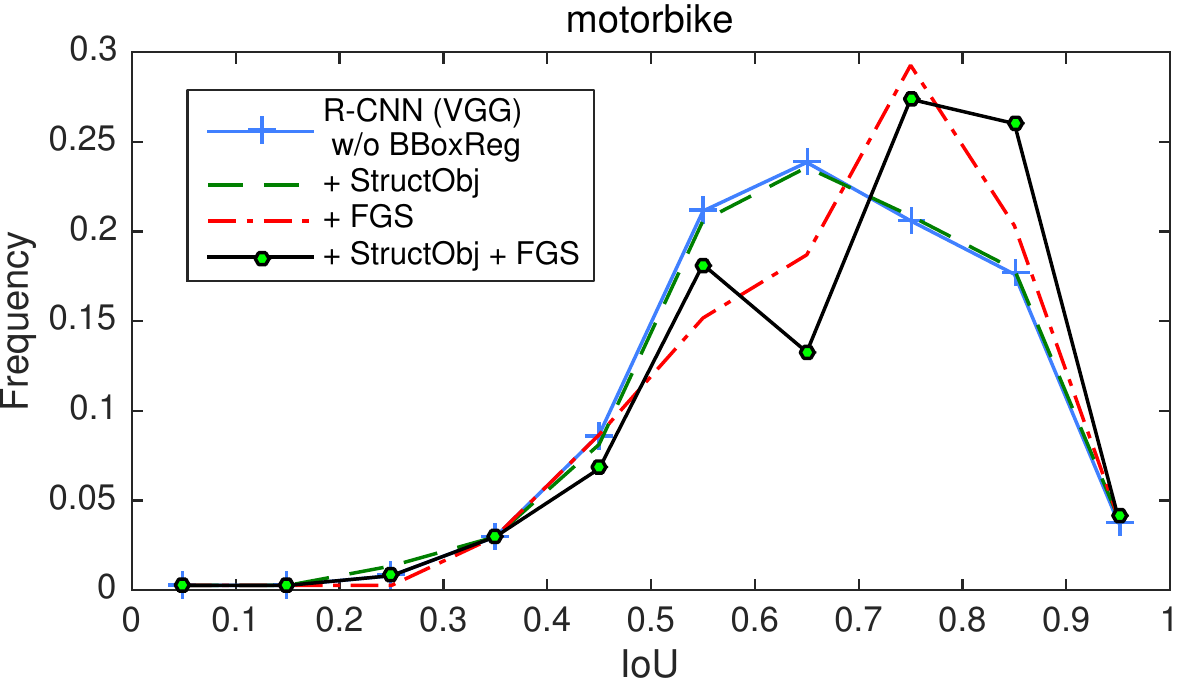}\hfill \\

\hfill
\includegraphics[width=0.3\textwidth]{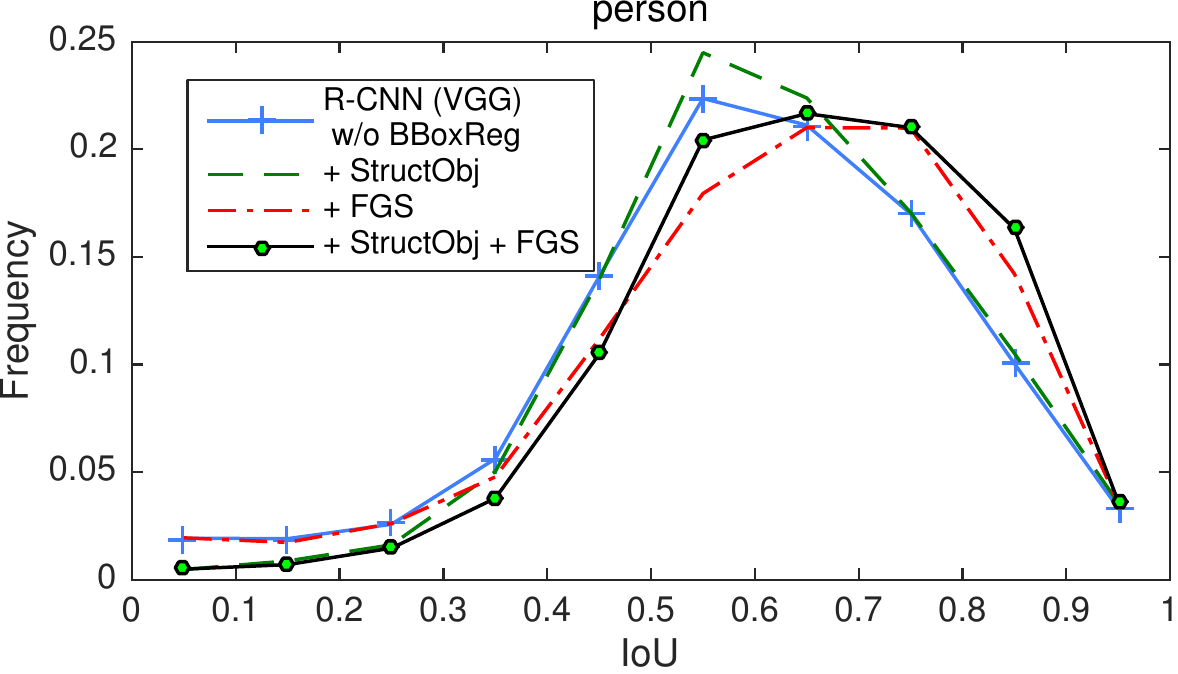}\hfill
\includegraphics[width=0.3\textwidth]{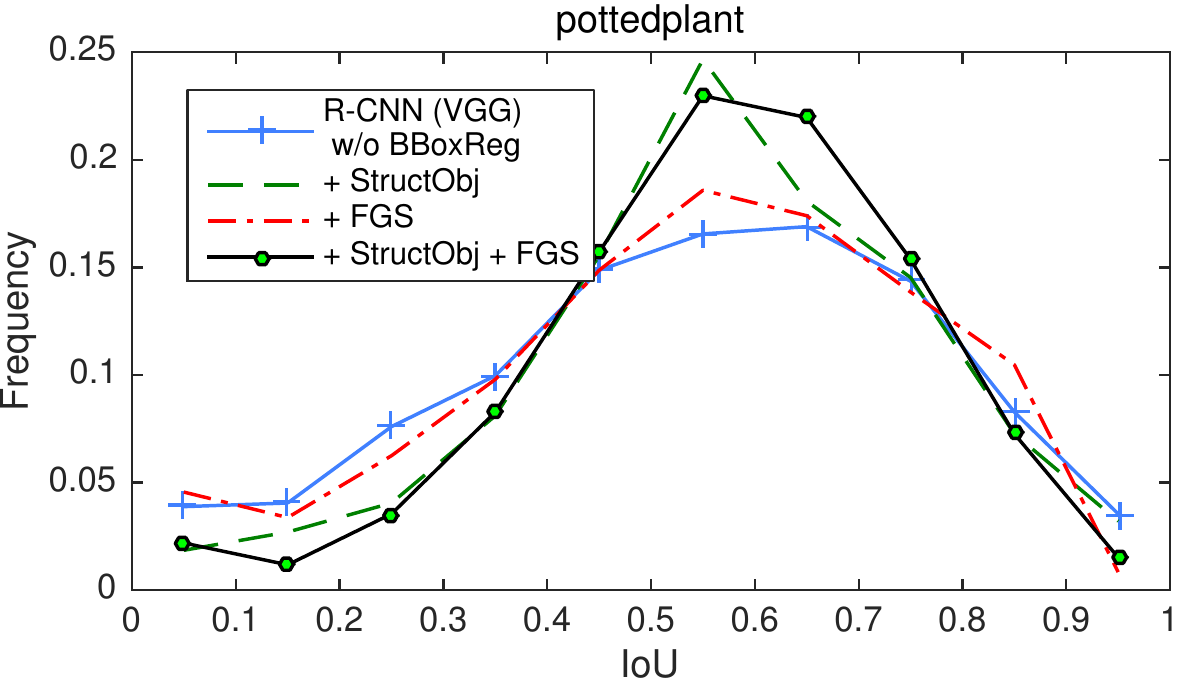}\hfill
\includegraphics[width=0.3\textwidth]{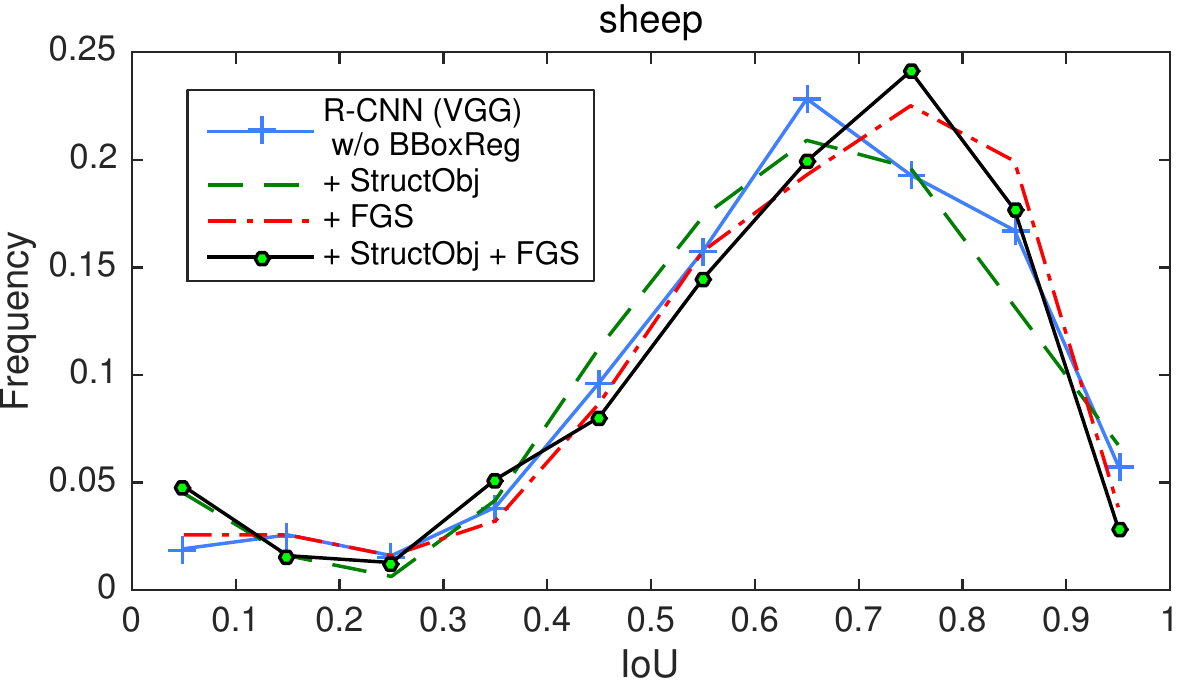}\hfill \\

\hfill
\includegraphics[width=0.3\textwidth]{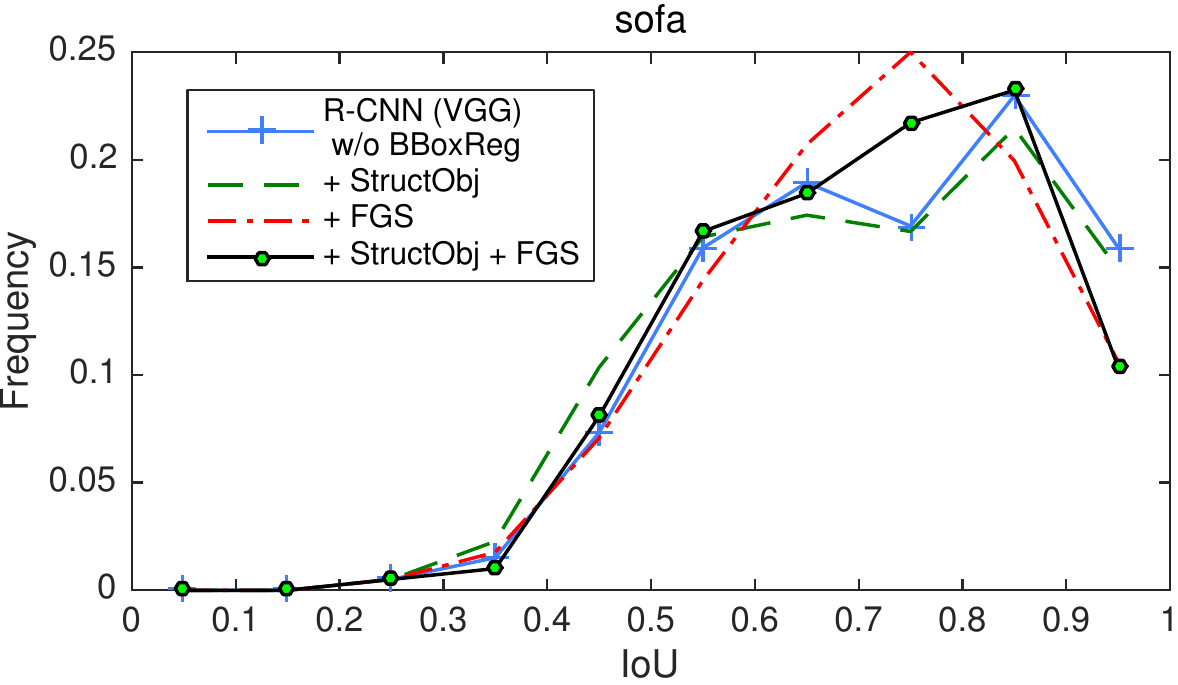}\hfill
\includegraphics[width=0.3\textwidth]{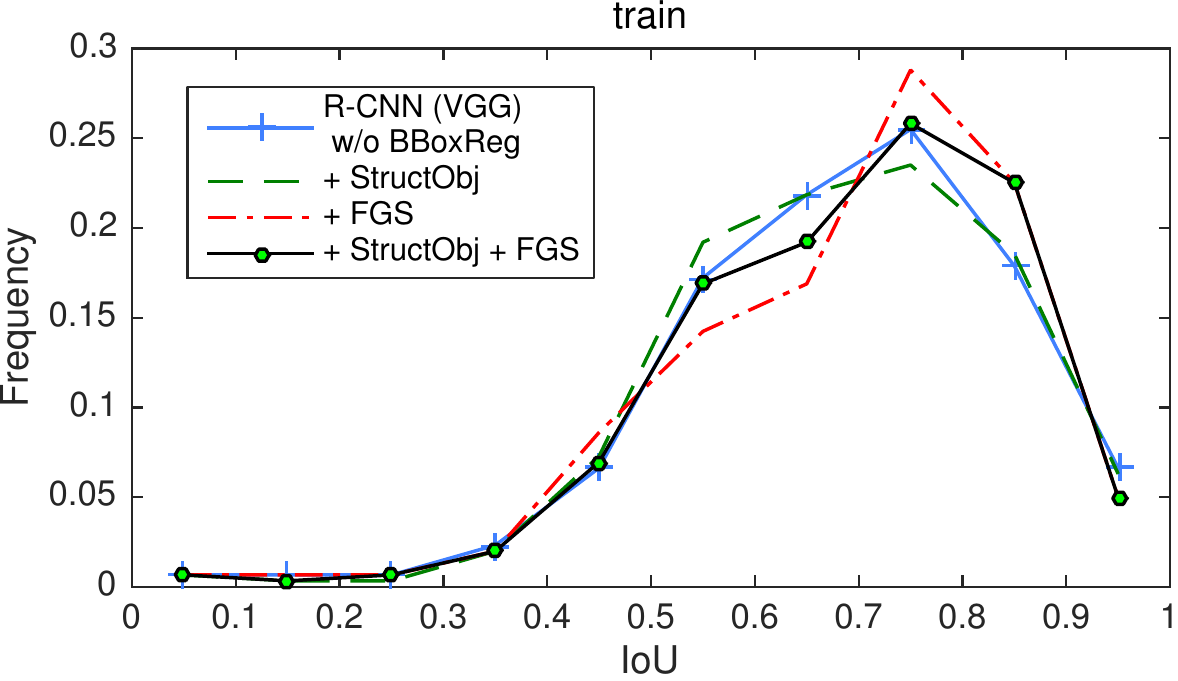}\hfill
\includegraphics[width=0.3\textwidth]{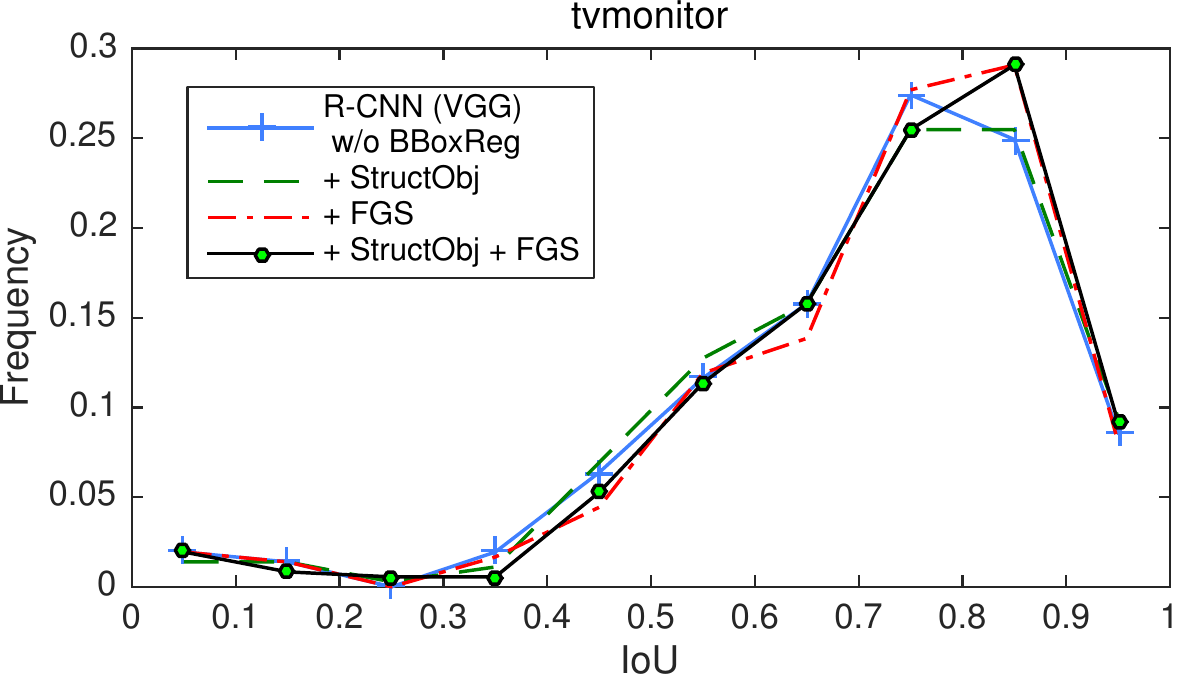}\hfill \\

\caption{Category-wise localization accuracy (in terms of the IoU between a ground truth annotation and its closest detected box) distributions for four different models (VGGNet without BBoxReg, + StructObj, + FGS, + StructObj + FGS)  on PASCAL VOC 2007 datasets. \label{fig:gt-loc-no-bb}}
\end{figure}

\begin{figure}[h]
\centering
\hfill
\includegraphics[width=0.3\textwidth]{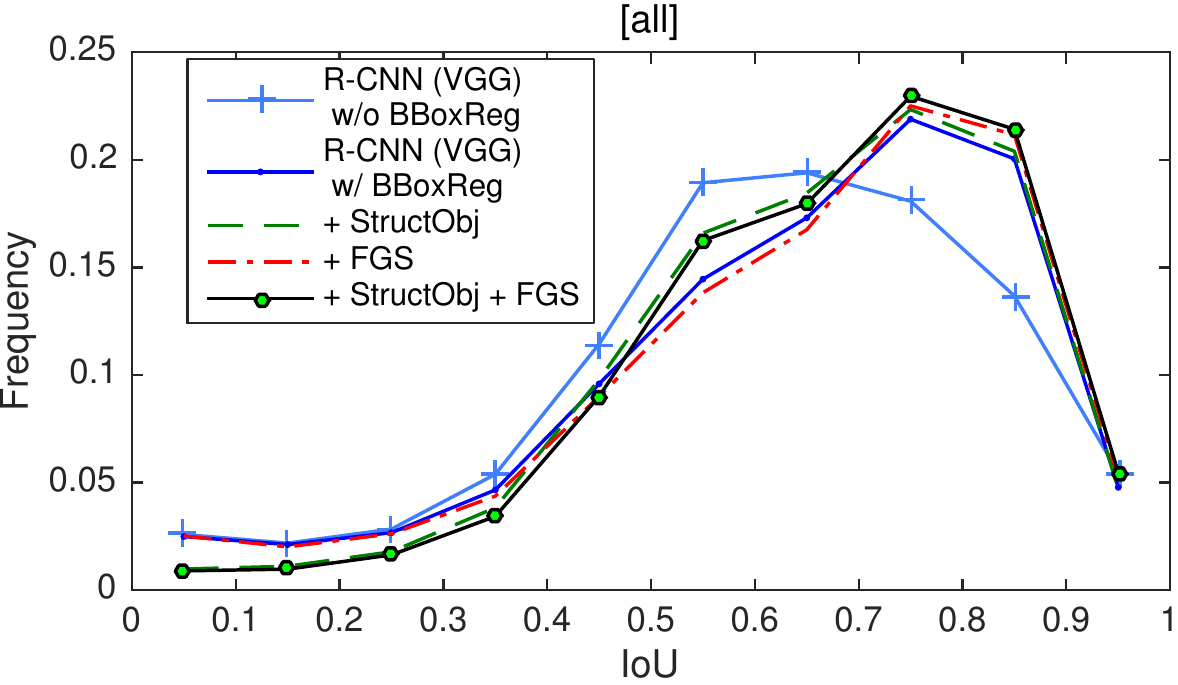}\hfill
\includegraphics[width=0.3\textwidth]{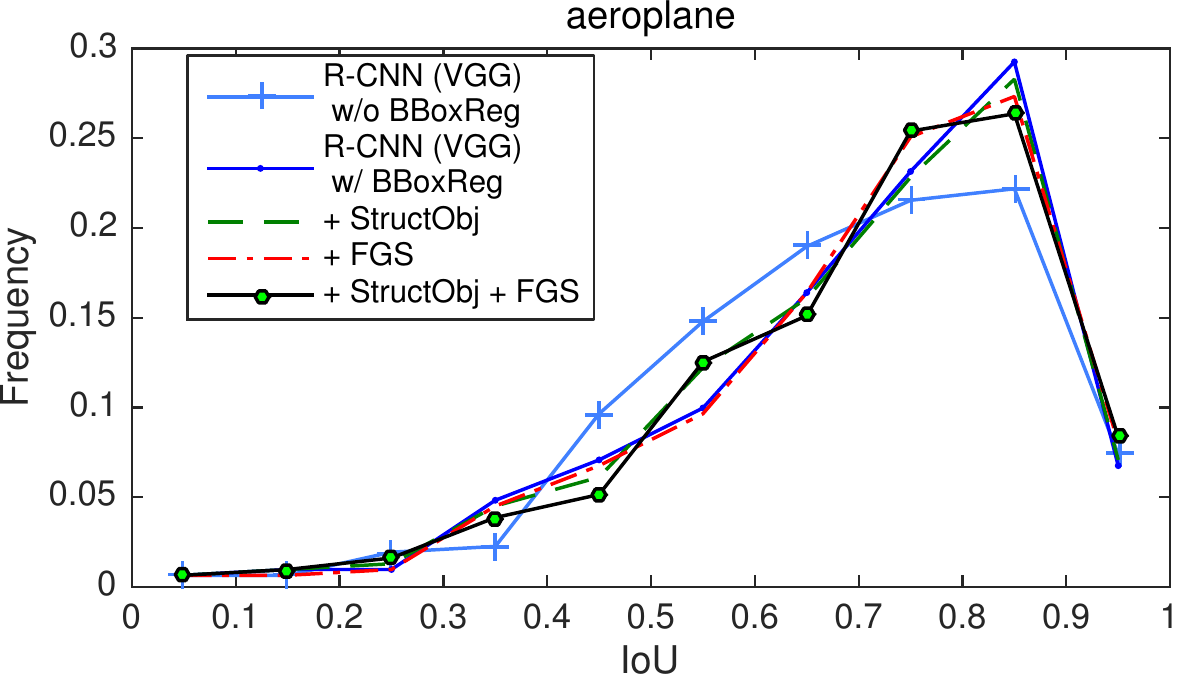}\hfill
\includegraphics[width=0.3\textwidth]{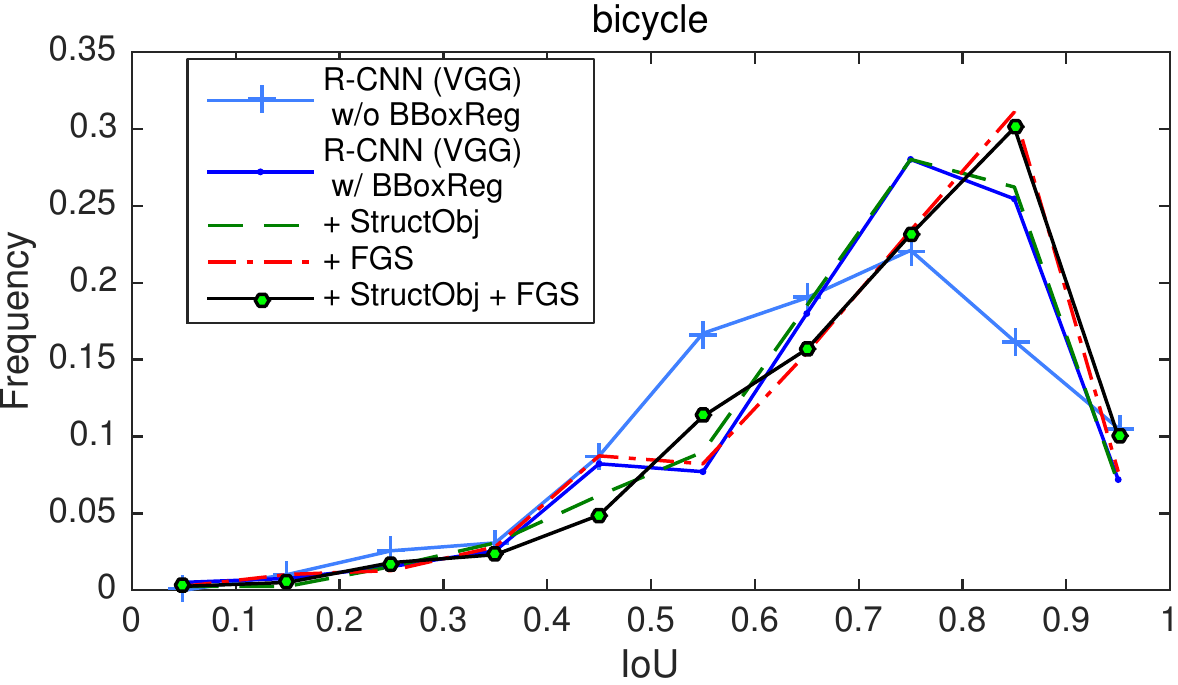}\hfill \\

\hfill
\includegraphics[width=0.3\textwidth]{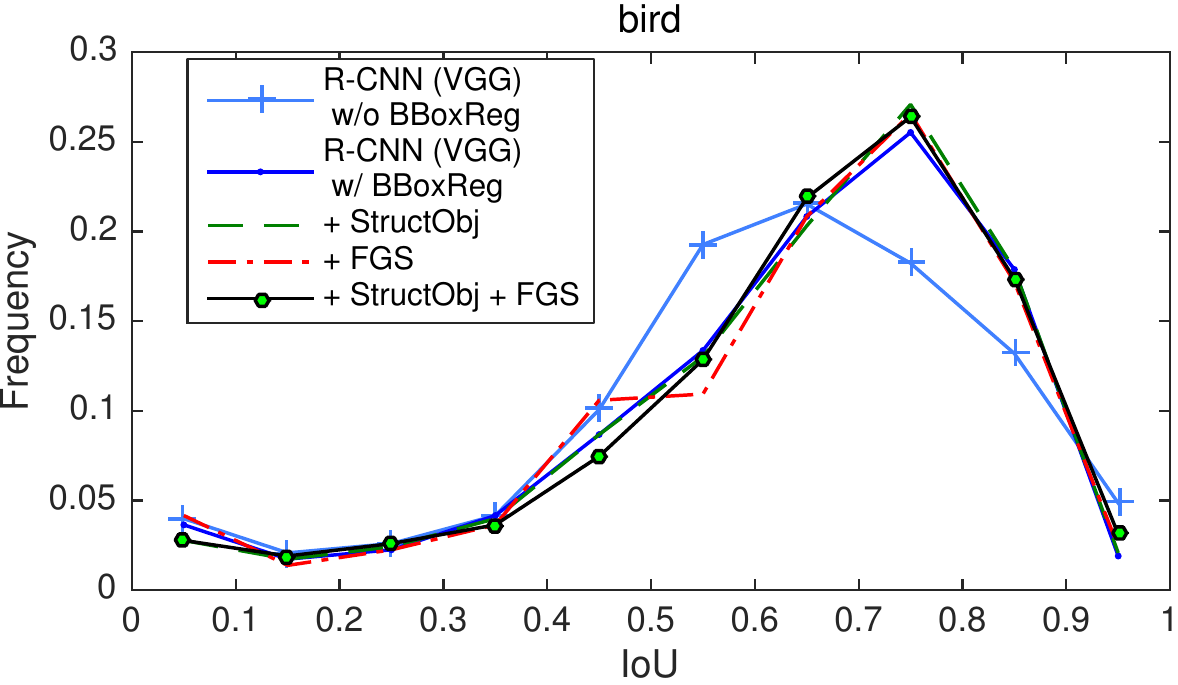}\hfill
\includegraphics[width=0.3\textwidth]{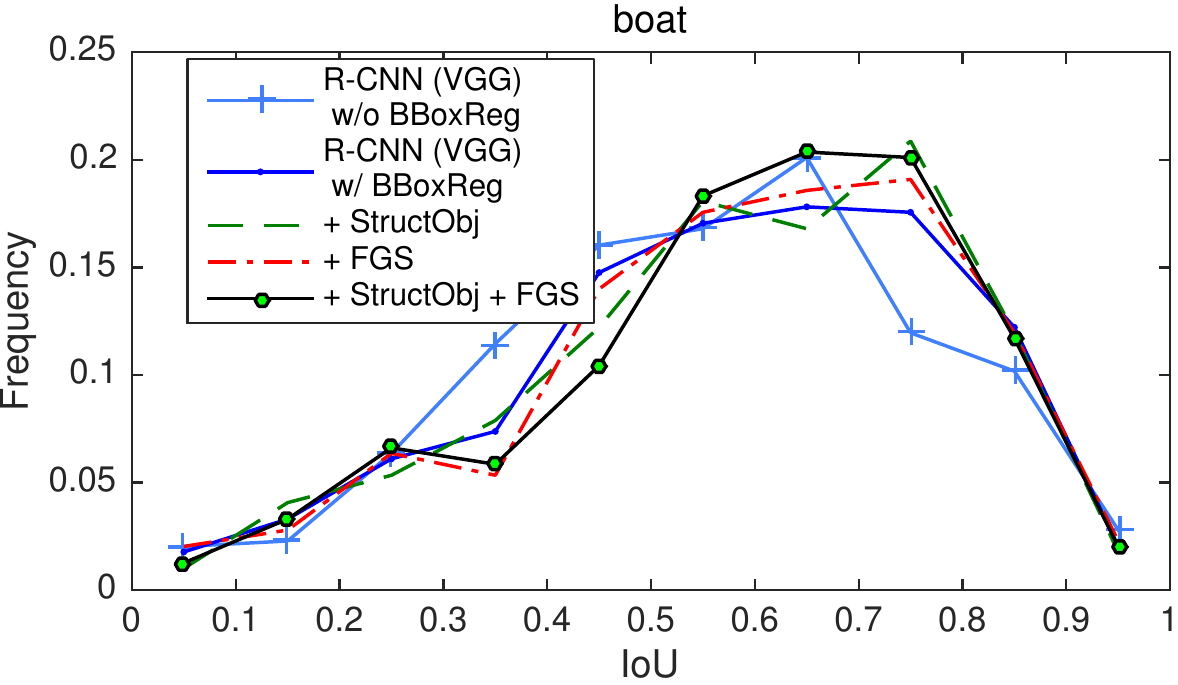}\hfill
\includegraphics[width=0.3\textwidth]{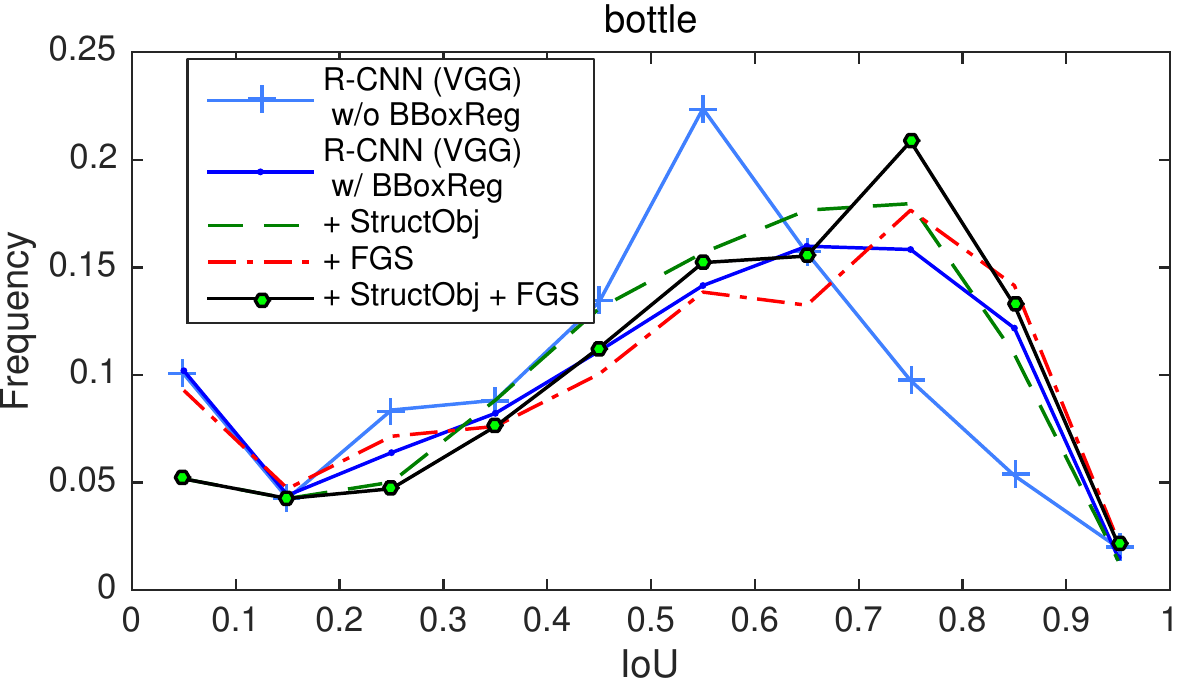}\hfill \\

\hfill
\includegraphics[width=0.3\textwidth]{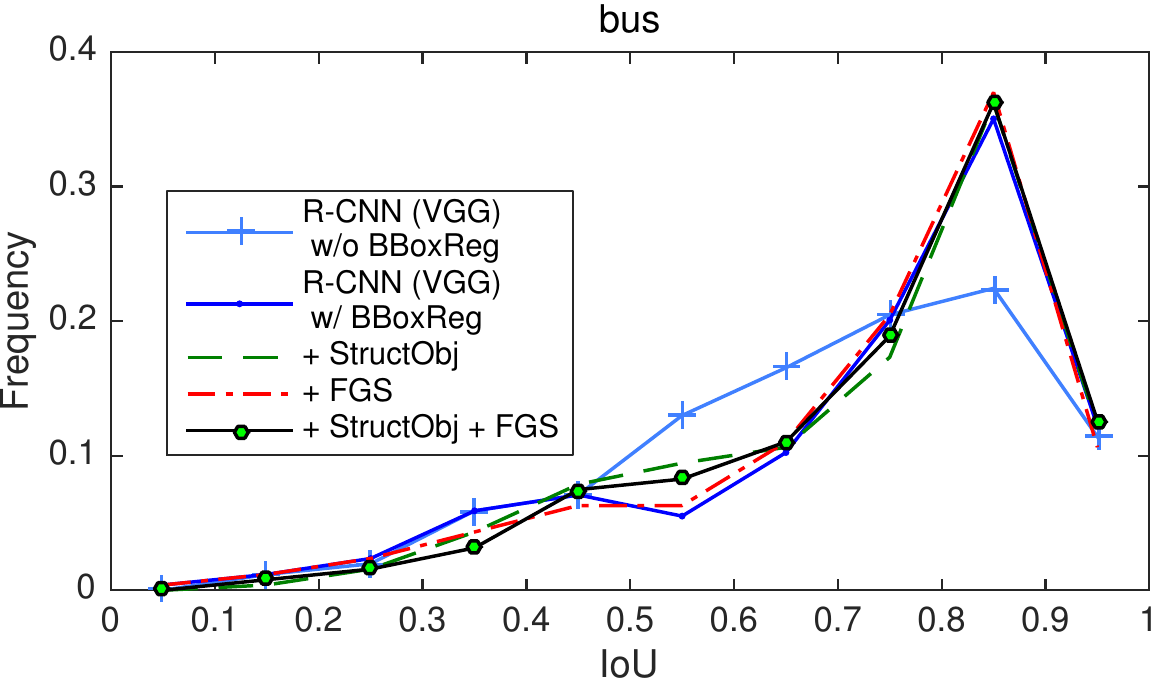}\hfill
\includegraphics[width=0.3\textwidth]{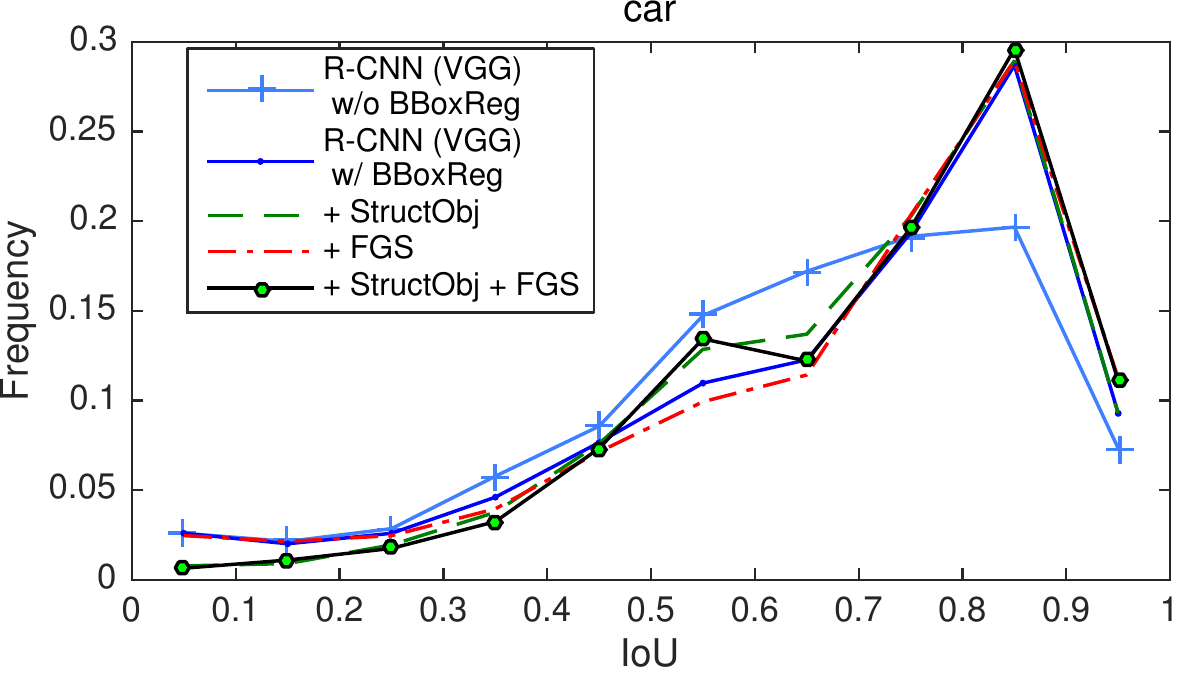}\hfill
\includegraphics[width=0.3\textwidth]{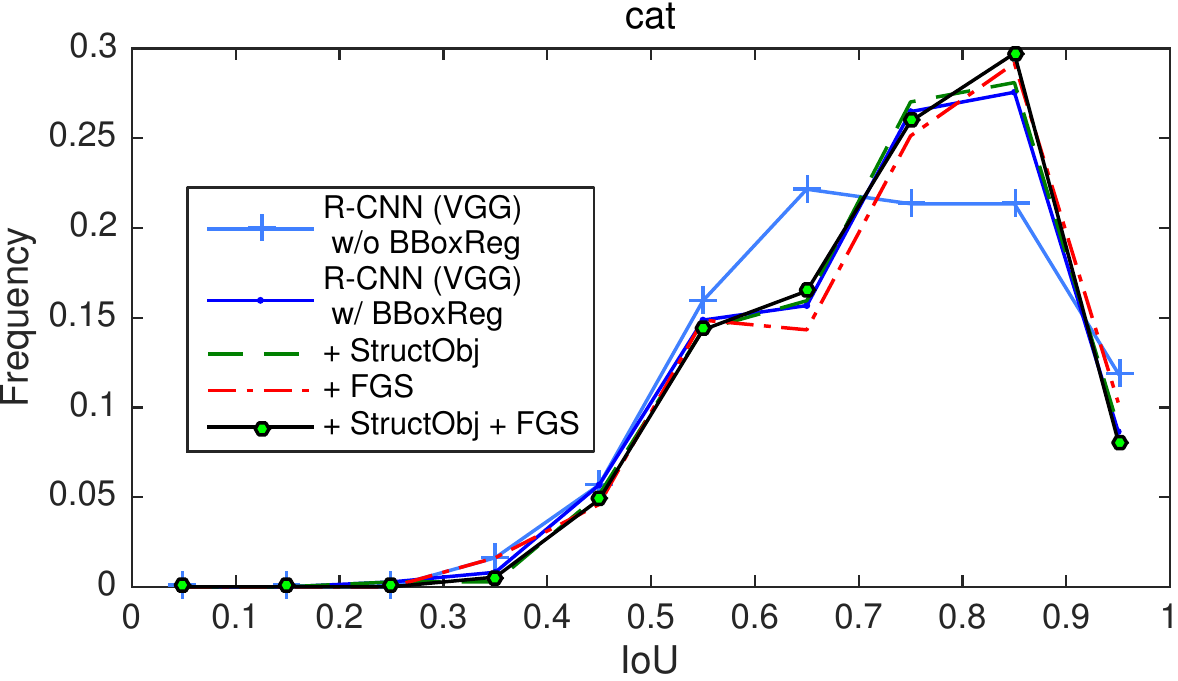}\hfill \\

\hfill
\includegraphics[width=0.3\textwidth]{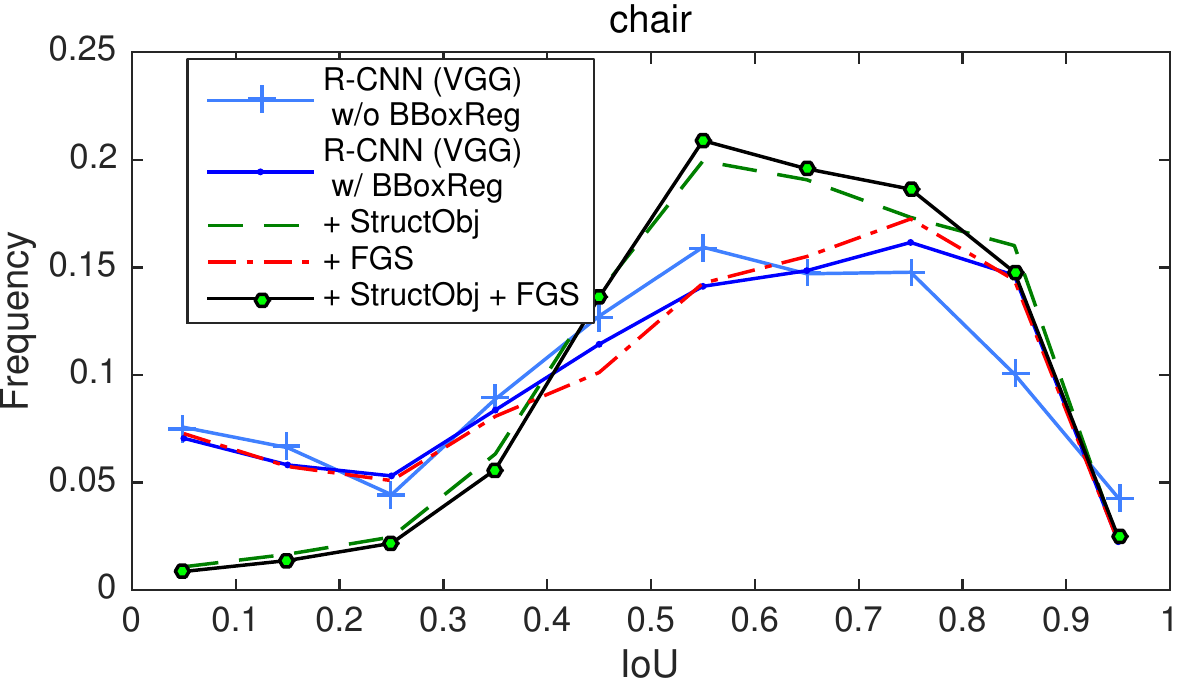}\hfill
\includegraphics[width=0.3\textwidth]{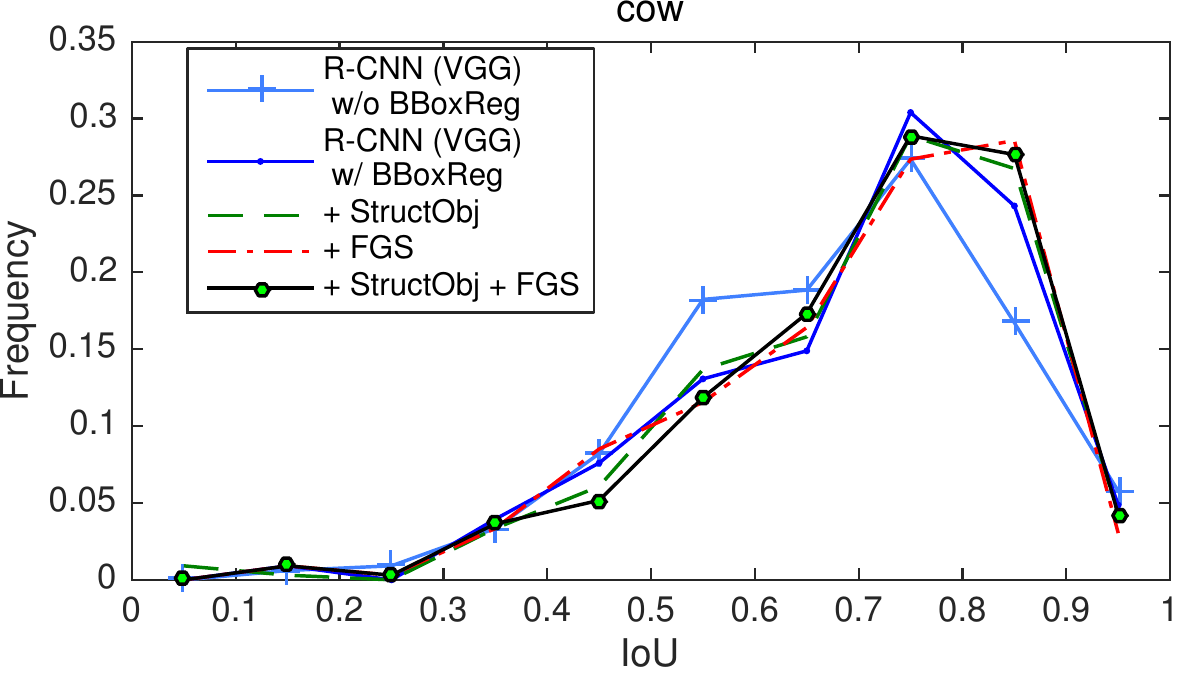}\hfill
\includegraphics[width=0.3\textwidth]{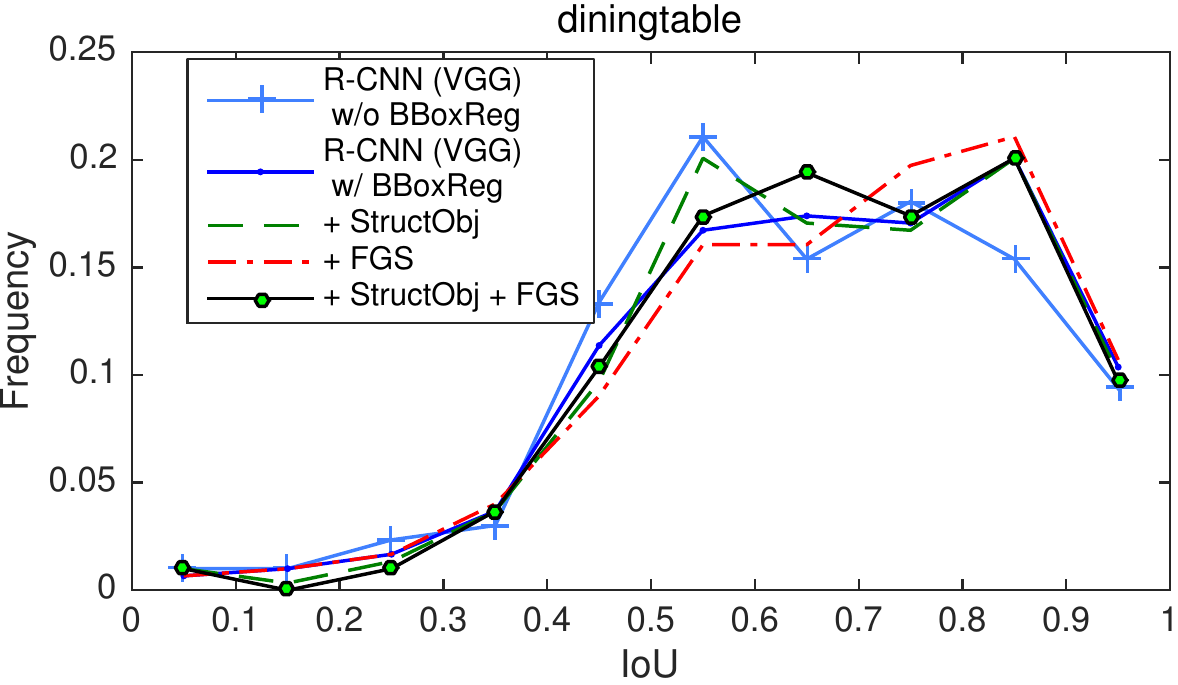}\hfill \\

\hfill
\includegraphics[width=0.3\textwidth]{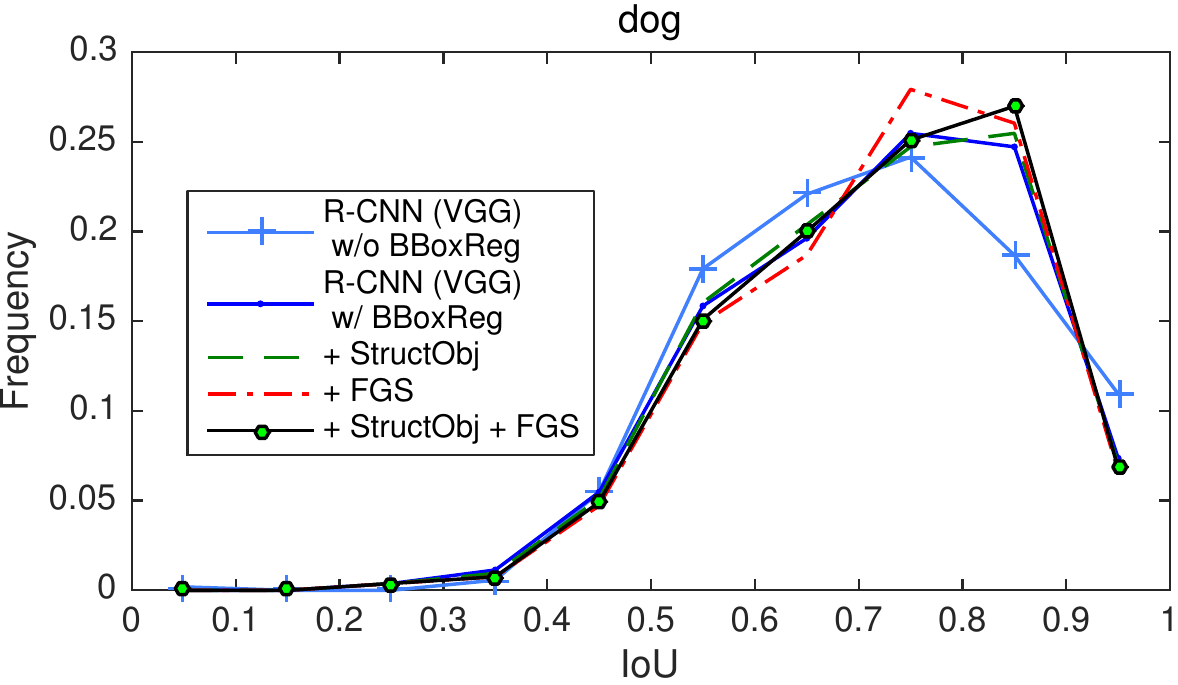}\hfill
\includegraphics[width=0.3\textwidth]{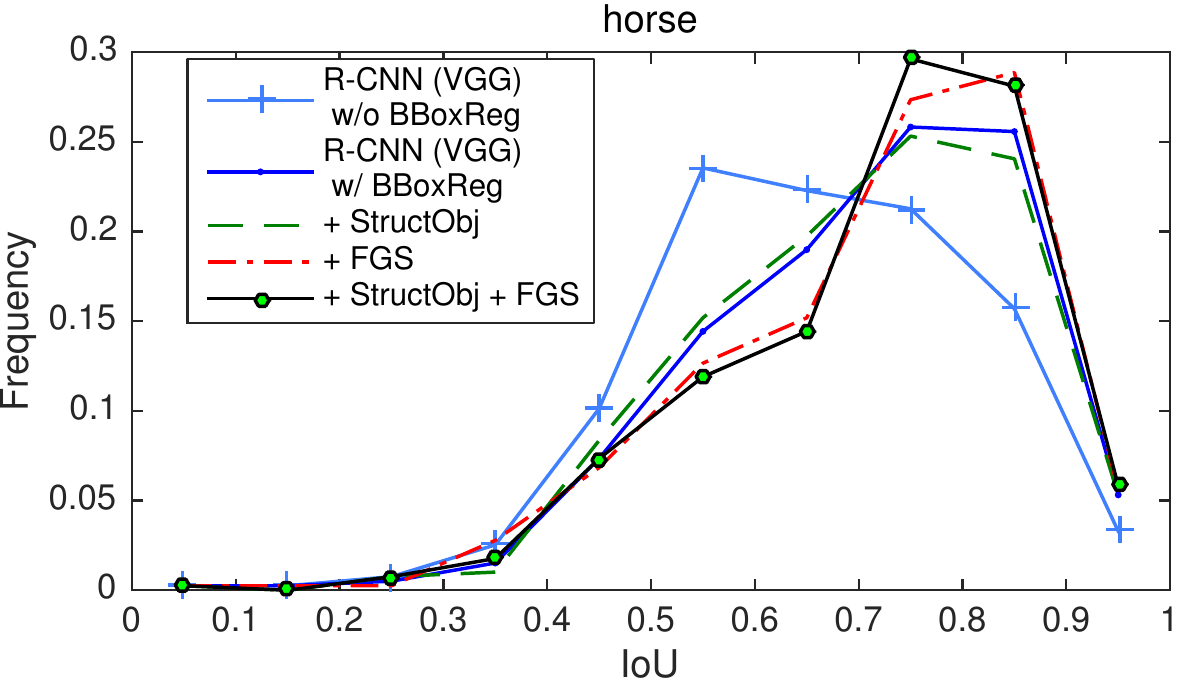}\hfill
\includegraphics[width=0.3\textwidth]{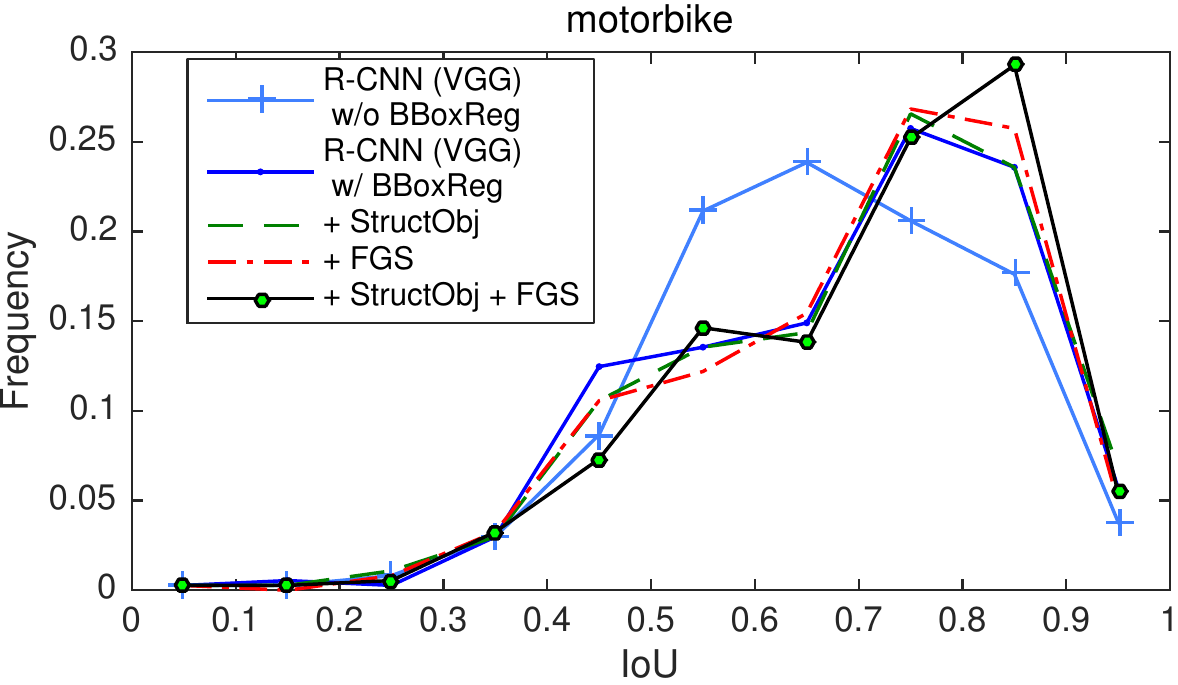}\hfill \\

\hfill
\includegraphics[width=0.3\textwidth]{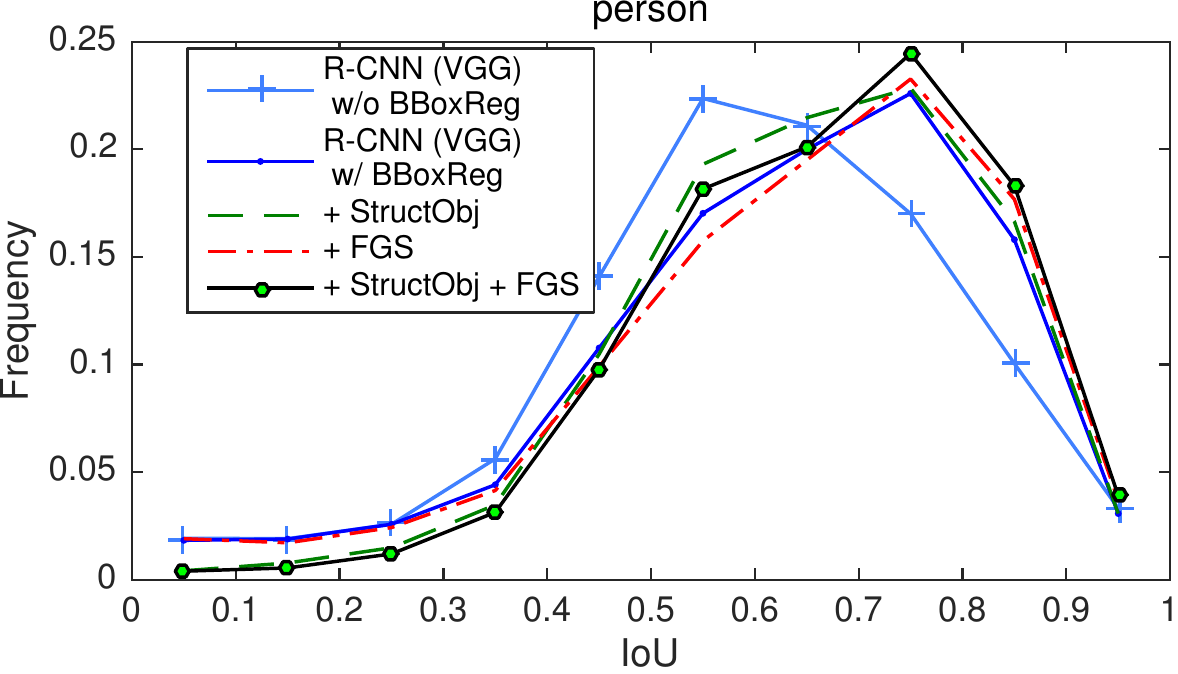}\hfill
\includegraphics[width=0.3\textwidth]{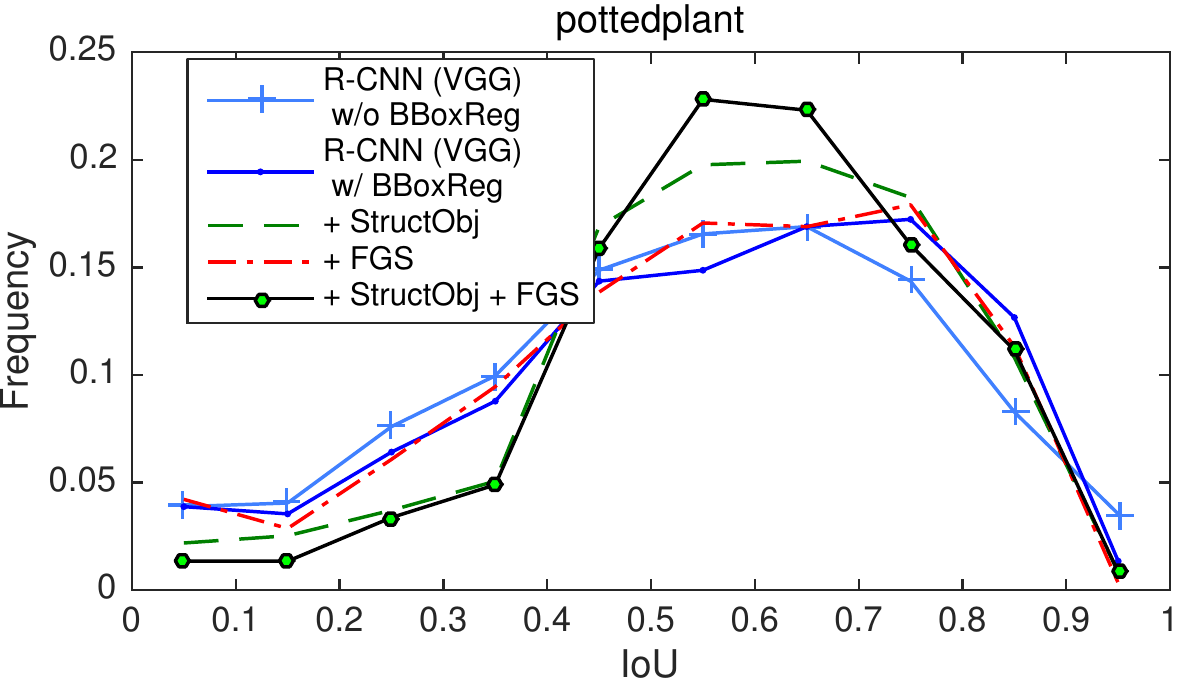}\hfill
\includegraphics[width=0.3\textwidth]{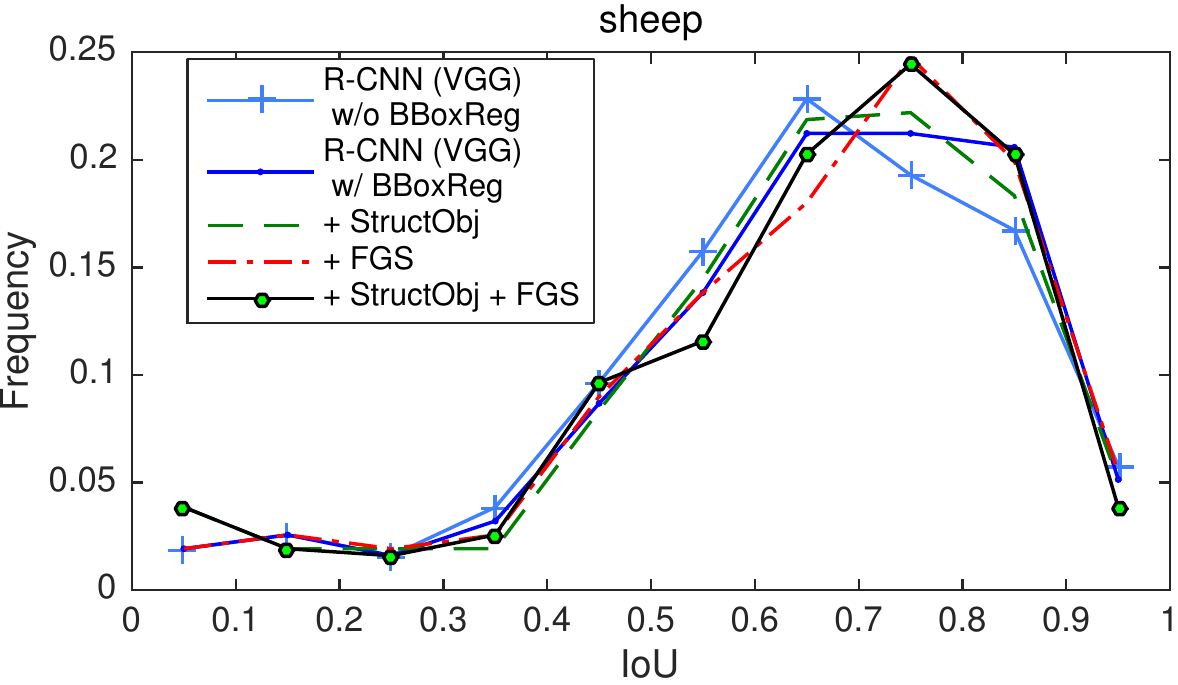}\hfill \\

\hfill
\includegraphics[width=0.3\textwidth]{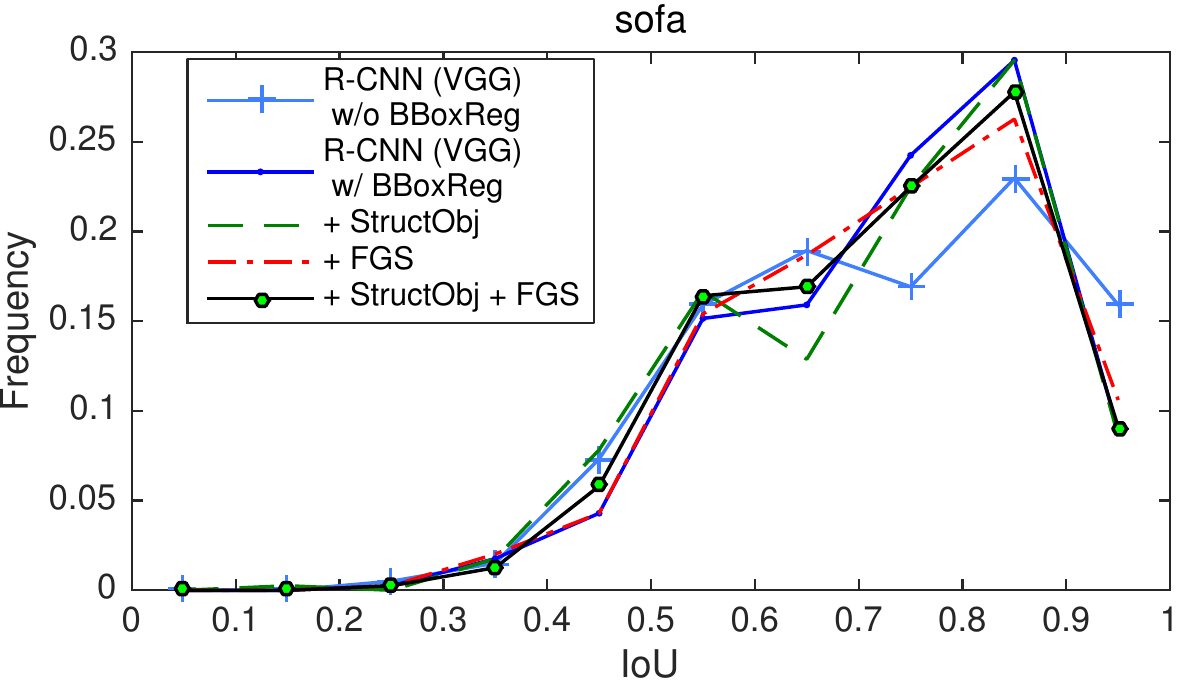}\hfill
\includegraphics[width=0.3\textwidth]{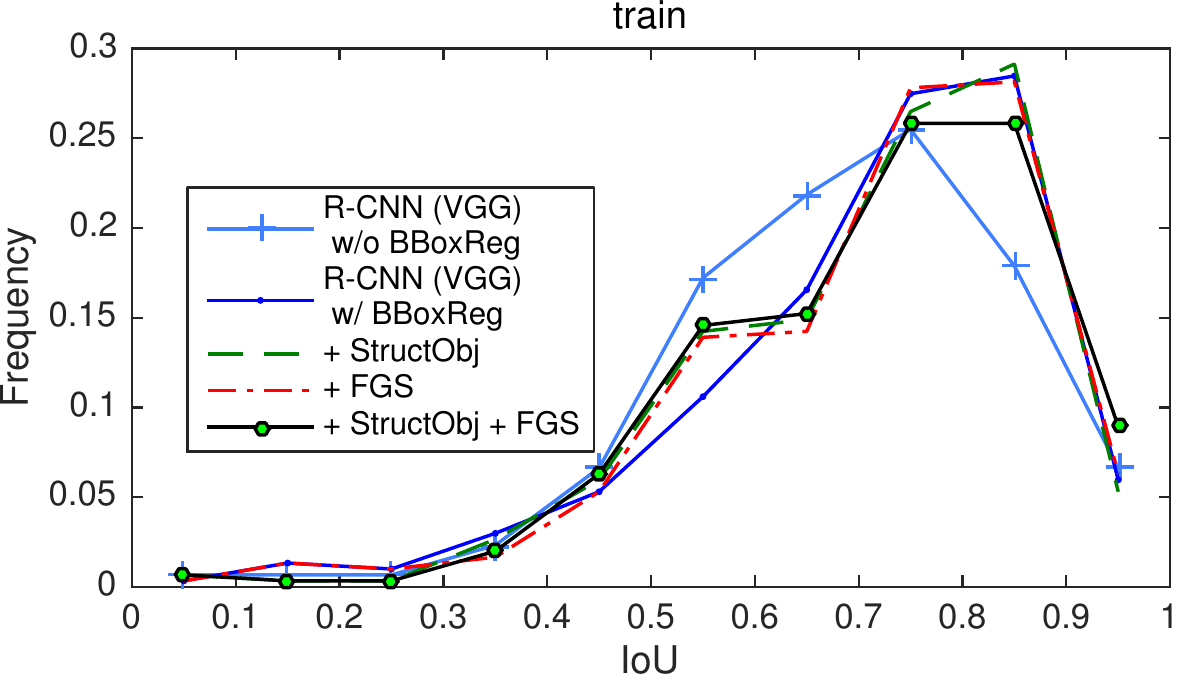}\hfill
\includegraphics[width=0.3\textwidth]{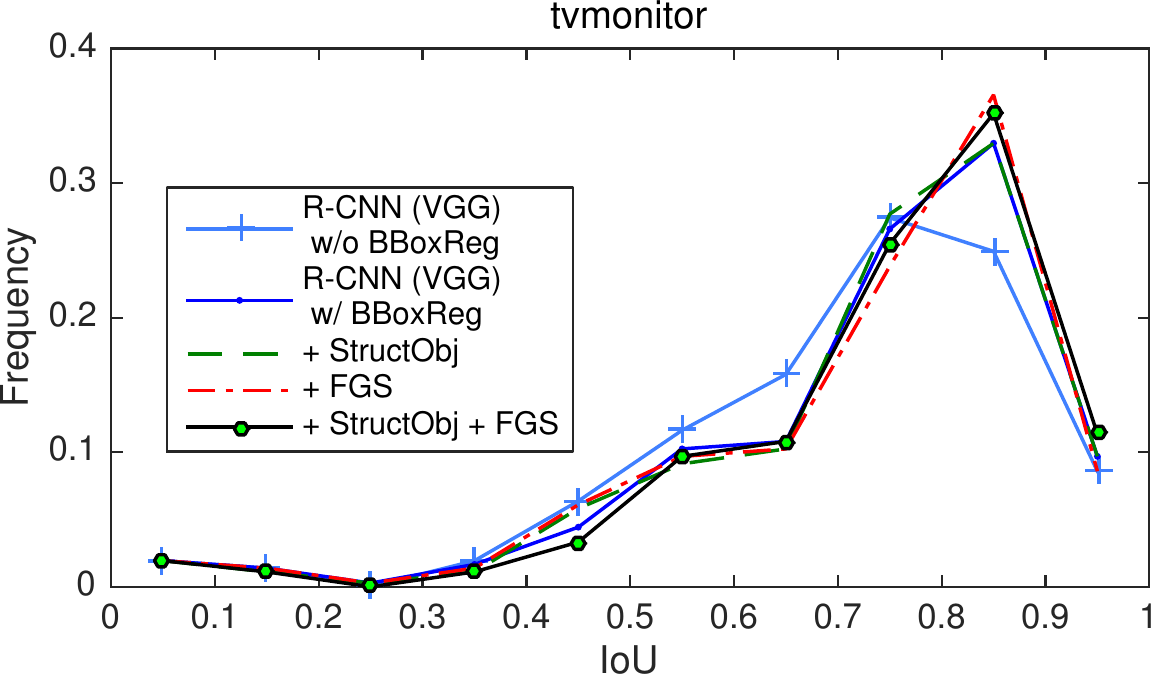}\hfill \\

\caption{Category-wise localization accuracy (in terms of the IoU between a ground truth annotation and its closest detected box) distributions for five different models (VGGNet without BBoxReg; VGGNet with BBoxReg, + StructObj, + FGS, + StructObj + FGS)  on PASCAL VOC 2007 datasets. \label{fig:gt-loc-with-bb-ex}}
\end{figure}

\commenttext{RUBEN: Modified the paragraph below, here is the original if you guys want to put it back to what it was:
In this section, we analyze the localization behavior of different methods. For each ground truth annotation, we find the detected box that is closest to it and compute the overlap (i.e., IoU) between this box and the ground truth, which is the best possible localization predicted by the given method. We estimate the distribution of the overlaps for every category, and compared different methods in Figure~\ref{fig:gt-loc-no-bb} (without BBoxReg) and Figure~\ref{fig:gt-loc-with-bb-ex} BBoxReg). A method with better localization ability should associate with a curve higher at high IoU interval (e.g., $0.6$ to $0.9$) and lower at low IoU interval (e.g., $0$ to $0.4$). The peak of the curve should also lean to the right.}
In this section, we analyze the localization behavior of different methods. We find the predicted bounding boxes that most accurately localize the ground truth from each method by picking the detected box with the highest overlap (i.e., IoU) with respect to each ground truth bounding box. Comparisons between the different methods are performed by estimating the distribution of the overlaps for every category. Our findings are shown in Figure~\ref{fig:gt-loc-no-bb} (without BBoxReg) and Figure~\ref{fig:gt-loc-with-bb-ex} (with BBoxReg + the baseline without BBoxReg). Methods with better localization ability should have higher frequency at higher IoU (i.e., IoU between 0.6 and 0.9) and lower frequency at lower IoU (i.e., IoU between 0 and 0.4). Curve peaks leaning to the right signify better localization.

\commenttext{RUBEN: Modified the paragraph below, here is the original if you guys want to put it back to what it was:
As it is shown in Figure~\ref{fig:gt-loc-no-bb} and \ref{fig:gt-loc-with-bb-ex}, individually applying FGS and StructObj both results in better localization compared to the baseline R-CNN, no matter using bounding box regression or not.}
As shown in Figure~\ref{fig:gt-loc-no-bb} and \ref{fig:gt-loc-with-bb-ex}, individually applying FGS and StructObj results in better localization compared to the baseline R-CNN, regardless of using or not using bounding box regression.
\commenttext{RUBEN: Modified the paragraph below, here is the original if you guys want to put it back to what it was:
However, the performance improvement due to FGS and StructObj is in different manners. 
In general, FGS pushes the distribution peak to the right.
StructObj pulls the distribution peak higher, and pushes the frequencies in the low IoU interval lower. 
It indicates that FGS can help reasonable localization become more accurate, while StructObj can make poor localization become reasonable.}
However, the performance improvement due to FGS and StructObj independently results in different phenomena.
In general, we found that FGS pushes the distribution peak to the right, and StructObj pulls the distribution peak higher while pushing the frequencies in the low IoU interval down. This indicates that FGS can propose more accurately localized boxes if the original set of bounding boxes are reasonably well localized, and StructObj can make detection scores more accurate (i.e., give low detection scores to boxes with low overlap and high detection scores to boxes with high overlap) based on the overlap of the proposed bounding boxes with respect to the ground truth.
Combining FGS and StructObj together capitalizes on the advantages of both, and leads to the best localization accuracy.  

\section{Examples with the largest improvement on PASCAL VOC 2007 test set}
\label{sup:example-improvement}

In this section, we show examples with the largest improvement in localization accuracy using our best proposed method (VGGNet + StructObj + FGS) over the baseline detection method (VGGNet) from PASCAL VOC 2007 test set. For each example in Figure~\ref{fig:best-voc2007}, we show the category of interest on the left-bottom corner of the image, and draw the detection of our best proposed method (in yellow box) that is best matched with the particular ground truth (in green box) and the best matched detection of the baseline (in red box). The number on the top right of the detected bounding box denotes the IoU with the ground truth. 

\section{Top-ranked false positives on PASCAL VOC 2007 test set}
\label{sup:example-fp}

In this section, we show examples of the top-ranked false positive detections\footnote{The top-ranked false positives are selected among false positive bounding boxes with the highest detection scores.} (in green box) of our best proposed method (VGGNet + StructObj + FGS) from PASCAL VOC 2007 test set (Figure~\ref{fig:fp-voc2007}).
We categorize the false positives into four categories as in~\cite{voc-error}: 
\begin{itemize}
\item{loc: poor localization,}
\item{sim: confusion with similar objects,}
\item{oth: confusion with other objects,}
\item{bg: confusion with background or unlabeled objects.}
\end{itemize}
The overlap (ov) measured by the IoU between the false positive detection and its best matching ground truth bounding box is provided.
For ``loc'' examples, the closest bounding box annotated with the same object category is provided as a ground truth (in yellow box).
For ``sim'' or ``oth'' examples, the closest bounding box annotated with any object category is provided as a ground truth (in red box).

\section{Random detection examples on PASCAL VOC 2007 test set}
\label{sup:example-rand}

Finally, we show randomly selected detection examples of our best proposed method (VGGNet + StructObj + FGS) from PASCAL VOC 2007 test set.
In Figure~\ref{fig:random-voc2007}, we use bounding boxes with different colors for different categories.
The category label with detection score is displayed on the top-left corner of each bounding box. 
Detections with low scores are ignored. 

\newpage{}

\clearpage

\begin{center}
{\begin{tabular}{ c }
\includegraphics[height=1.0in]{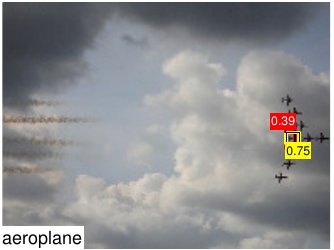}\hspace{0.05in}
\includegraphics[height=1.0in]{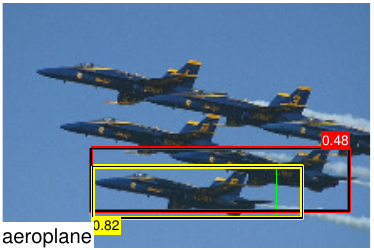}\hspace{0.05in}
\includegraphics[height=1.0in]{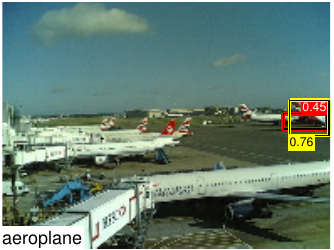}\hspace{0.05in}
\includegraphics[height=1.0in]{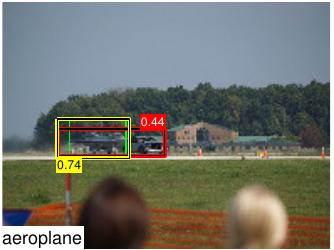}\\
aeroplane (examples with the largest improvement) \\
\end{tabular}}
\end{center}\vspace{-0.7em}

\begin{center}
{\begin{tabular}{ c }
\includegraphics[height=1.0in]{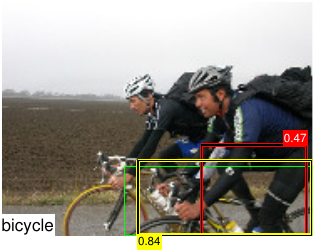}\hspace{0.05in}
\includegraphics[height=1.0in]{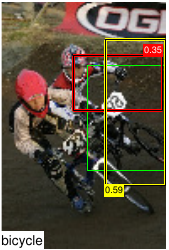}\hspace{0.05in}
\includegraphics[height=1.0in]{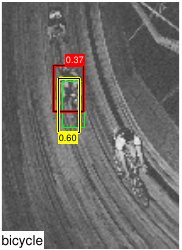}\hspace{0.05in}
\includegraphics[height=1.0in]{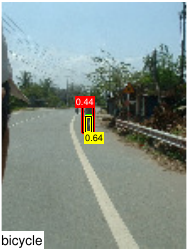}\\
bicycle (examples with the largest improvement) \\
\end{tabular}}
\end{center}\vspace{-0.7em}

\begin{center}
{\begin{tabular}{ c }
\includegraphics[height=1.0in]{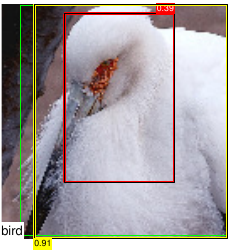}\hspace{0.05in}
\includegraphics[height=1.0in]{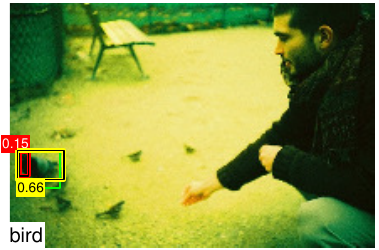}\hspace{0.05in}
\includegraphics[height=1.0in]{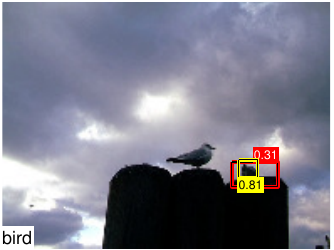}\hspace{0.05in}
\includegraphics[height=1.0in]{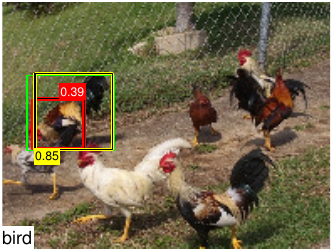}\hspace{0.05in}
\includegraphics[height=1.0in]{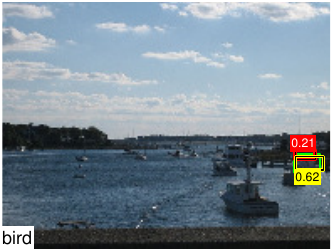}\\ bird (examples with the largest improvement) \\
\end{tabular}}
\end{center}\vspace{-0.7em}

\begin{center}
{\begin{tabular}{ c }
\includegraphics[height=1.0in]{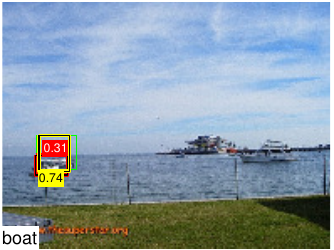}\hspace{0.05in}
\includegraphics[height=1.0in]{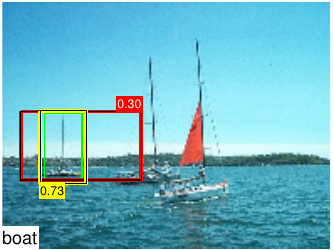}\hspace{0.05in}
\includegraphics[height=1.0in]{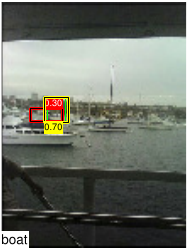}\hspace{0.05in}
\includegraphics[height=1.0in]{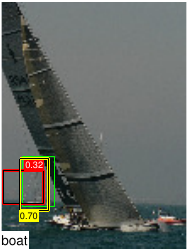}\hspace{0.05in}
\includegraphics[height=1.0in]{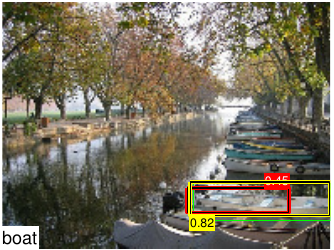}\\ boat (examples with the largest improvement) \\
\end{tabular}}
\end{center}\vspace{-0.7em}

\begin{center}
{\begin{tabular}{ c }
\includegraphics[height=1.0in]{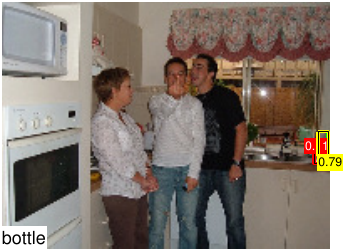}\hspace{0.05in}
\includegraphics[height=1.0in]{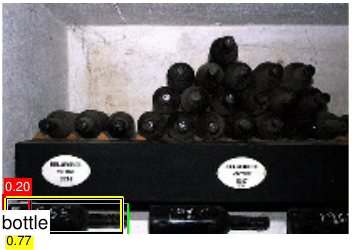}\hspace{0.05in}
\includegraphics[height=1.0in]{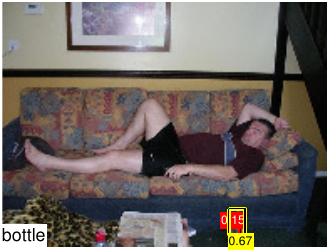}\hspace{0.05in}
\includegraphics[height=1.0in]{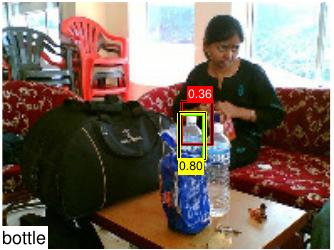}\\
bottle (examples with the largest improvement) \\
\end{tabular}}
\end{center}\vspace{-0.7em}

\begin{center}
{\begin{tabular}{ c }
\includegraphics[height=1.0in]{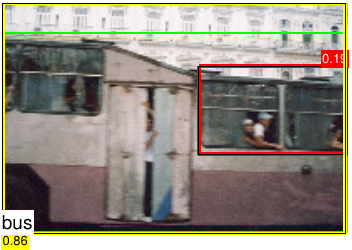}\hspace{0.05in}
\includegraphics[height=1.0in]{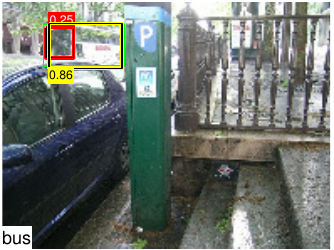}\hspace{0.05in}
\includegraphics[height=1.0in]{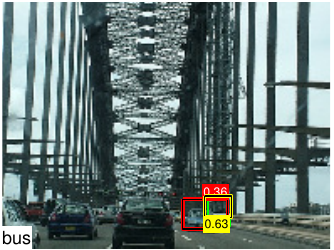}\hspace{0.05in}
\includegraphics[height=1.0in]{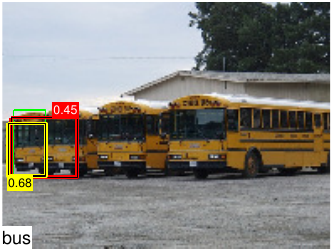}\\
bus (examples with the largest improvement) \\
\end{tabular}}
\end{center}\vspace{-0.7em}

\begin{center}
{\begin{tabular}{ c }
\includegraphics[height=1.0in]{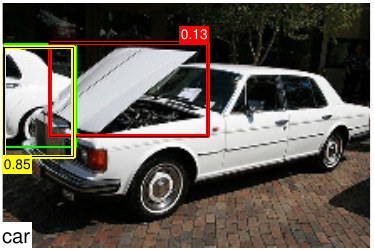}\hspace{0.05in}
\includegraphics[height=1.0in]{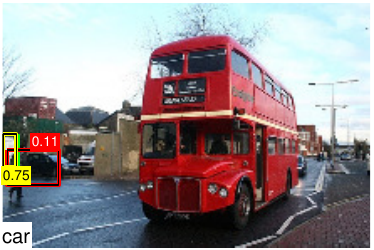}\hspace{0.05in}
\includegraphics[height=1.0in]{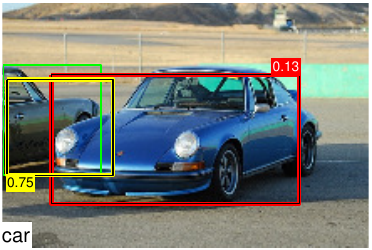}\hspace{0.05in}
\includegraphics[height=1.0in]{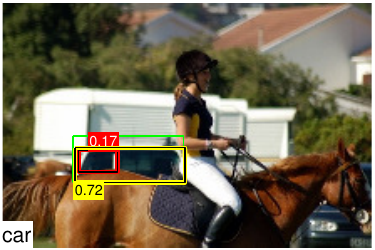}\\ car (examples with the largest improvement) \\
\end{tabular}}
\end{center}\vspace{-0.7em}

\begin{center}
{\begin{tabular}{ c }
\includegraphics[height=1.0in]{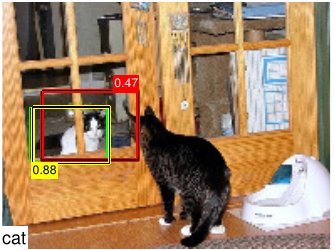}\hspace{0.05in}
\includegraphics[height=1.0in]{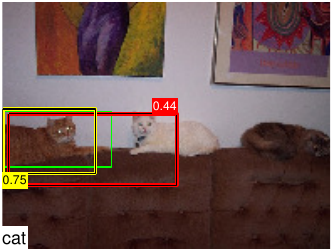}\hspace{0.05in}
\includegraphics[height=1.0in]{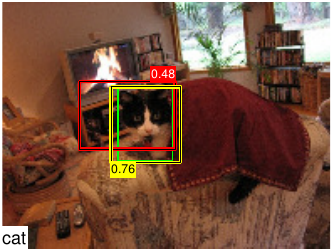}\hspace{0.05in}
\includegraphics[height=1.0in]{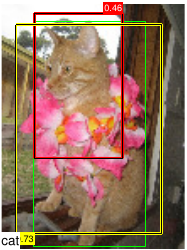}\\
cat (examples with the largest improvement) \\
\end{tabular}}
\end{center}\vspace{-0.7em}

\begin{center}
{\begin{tabular}{ c }
\includegraphics[height=1.0in]{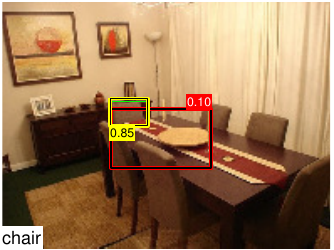}\hspace{0.05in}
\includegraphics[height=1.0in]{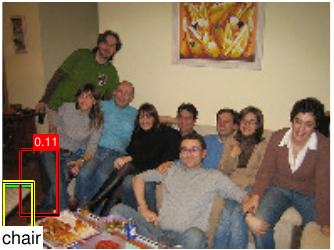}\hspace{0.05in}
\includegraphics[height=1.0in]{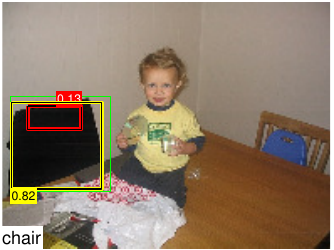}\hspace{0.05in}
\includegraphics[height=1.0in]{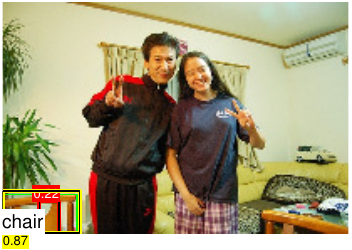}\hspace{0.05in}
\includegraphics[height=1.0in]{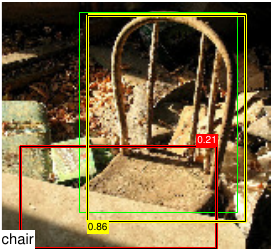}\\ chair (examples with the largest improvement) \\
\end{tabular}}
\end{center}\vspace{-0.7em}

\begin{center}
{\begin{tabular}{ c }
\includegraphics[height=1.0in]{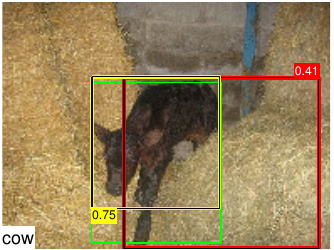}\hspace{0.05in}
\includegraphics[height=1.0in]{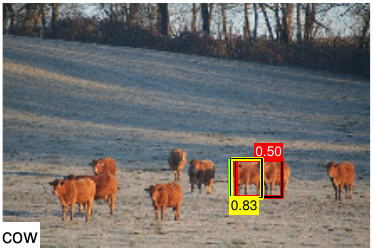}\hspace{0.05in}
\includegraphics[height=1.0in]{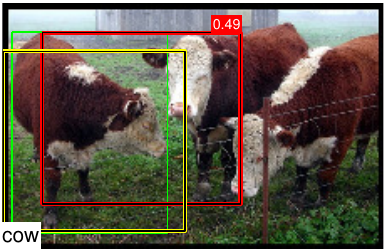}\hspace{0.05in}
\includegraphics[height=1.0in]{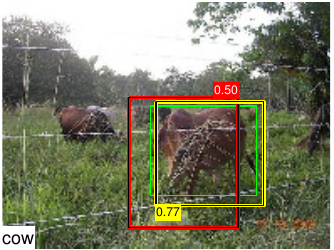}\\ cow (examples with the largest improvement) \\
\end{tabular}}
\end{center}\vspace{-0.7em}

\begin{center}
{\begin{tabular}{ c }
\includegraphics[height=1.0in]{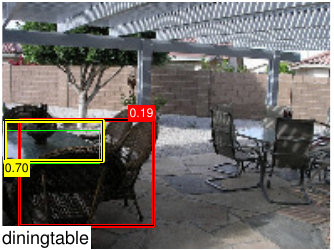}\hspace{0.05in}
\includegraphics[height=1.0in]{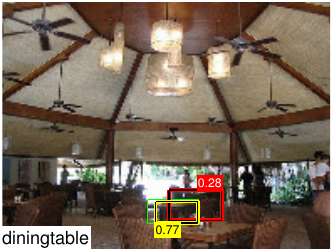}\hspace{0.05in}
\includegraphics[height=1.0in]{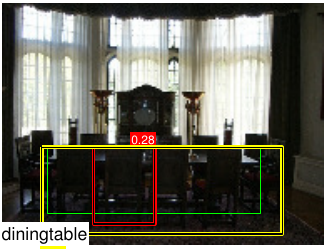}\hspace{0.05in}
\includegraphics[height=1.0in]{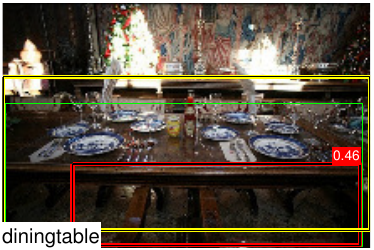}\hspace{0.05in}
\includegraphics[height=1.0in]{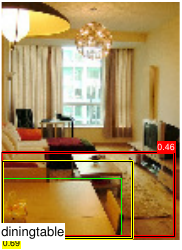}\\
diningtable (examples with the largest improvement) \\
\end{tabular}}
\end{center}\vspace{-0.7em}

\begin{center}
{\begin{tabular}{ c }
\includegraphics[height=1.0in]{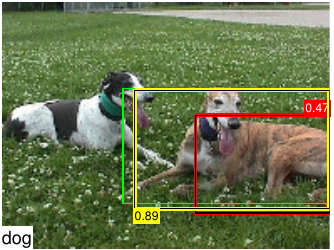}\hspace{0.05in}
\includegraphics[height=1.0in]{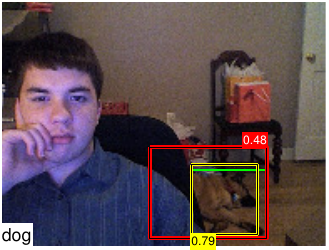}\hspace{0.05in}
\includegraphics[height=1.0in]{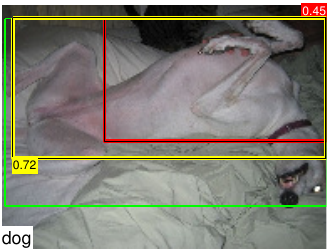}\hspace{0.05in}
\includegraphics[height=1.0in]{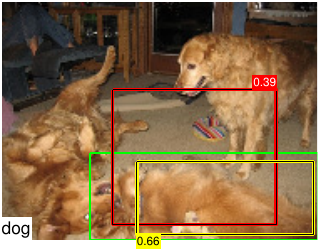}\hspace{0.05in}
\includegraphics[height=1.0in]{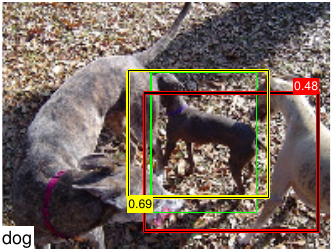}\\ dog (examples with the largest improvement) \\
\end{tabular}}
\end{center}\vspace{-0.7em}

\begin{center}
{\begin{tabular}{ c }
\includegraphics[height=1.0in]{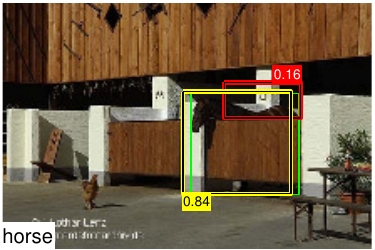}\hspace{0.05in}
\includegraphics[height=1.0in]{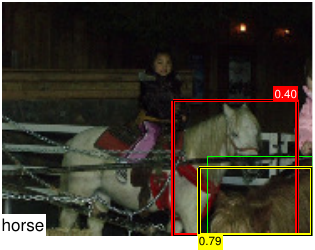}\hspace{0.05in}
\includegraphics[height=1.0in]{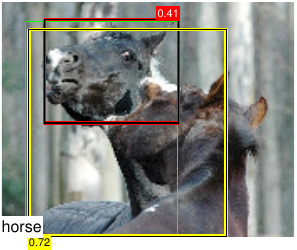}\hspace{0.05in}
\includegraphics[height=1.0in]{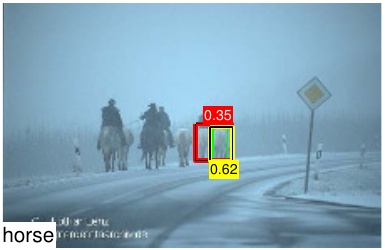}\\ horse (examples with the largest improvement) \\
\end{tabular}}
\end{center}\vspace{-0.7em}

\begin{center}
{\begin{tabular}{ c }
\includegraphics[height=1.0in]{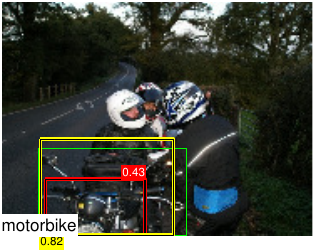}\hspace{0.05in}
\includegraphics[height=1.0in]{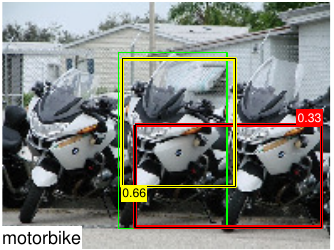}\hspace{0.05in}
\includegraphics[height=1.0in]{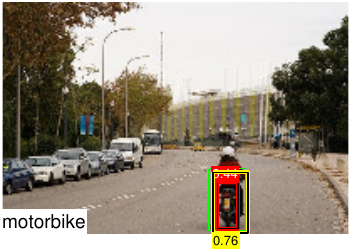}\hspace{0.05in}
\includegraphics[height=1.0in]{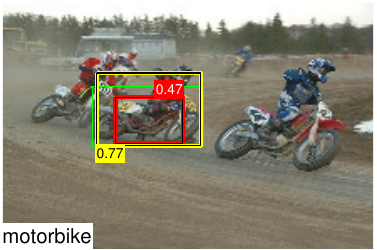}\\ motorbike (examples with the largest improvement) \\
\end{tabular}}
\end{center}\vspace{-0.7em}

\begin{center}
{\begin{tabular}{ c }
\includegraphics[height=1.0in]{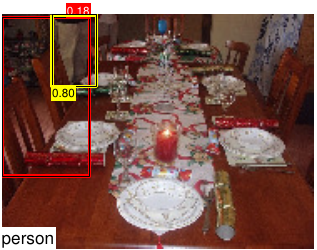}\hspace{0.05in}
\includegraphics[height=1.0in]{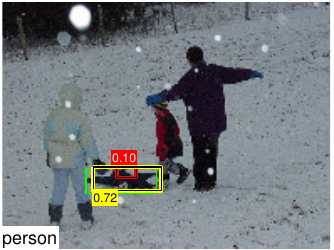}\hspace{0.05in}
\includegraphics[height=1.0in]{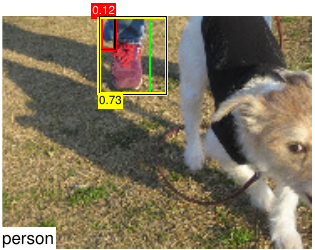}\hspace{0.05in}
\includegraphics[height=1.0in]{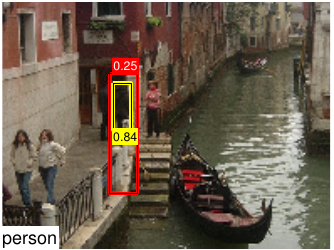}\\ person (examples with the largest improvement) \\
\end{tabular}}
\end{center}\vspace{-0.7em}

\begin{center}
{\begin{tabular}{ c }
\includegraphics[height=1.0in]{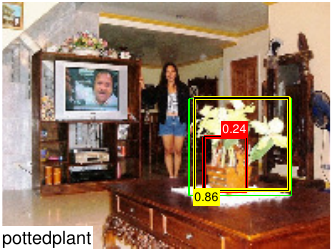}\hspace{0.05in}
\includegraphics[height=1.0in]{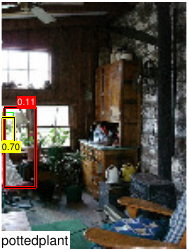}\hspace{0.05in}
\includegraphics[height=1.0in]{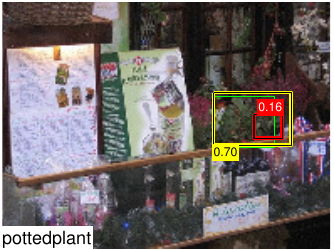}\hspace{0.05in}
\includegraphics[height=1.0in]{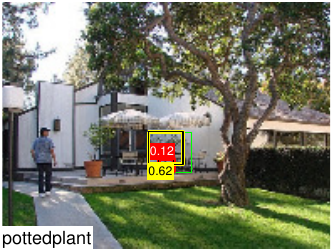}\hspace{0.05in}
\includegraphics[height=1.0in]{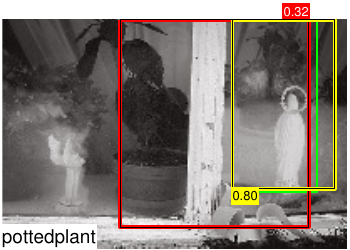}\\ pottedplant (examples with the largest improvement) \\
\end{tabular}}
\end{center}\vspace{-0.7em}

\begin{center}
{\begin{tabular}{ c }
\includegraphics[height=1.0in]{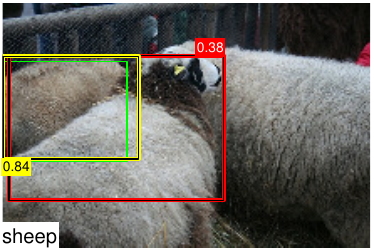}\hspace{0.05in}
\includegraphics[height=1.0in]{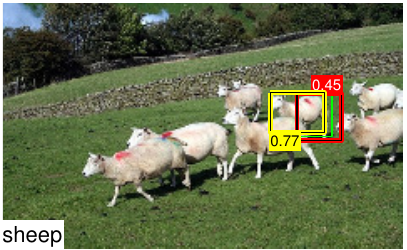}\hspace{0.05in}
\includegraphics[height=1.0in]{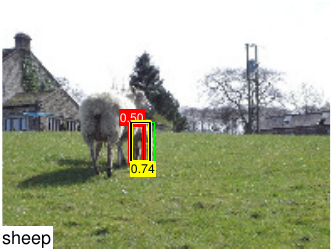}\\
sheep (examples with the largest improvement) \\
\end{tabular}}
\end{center}\vspace{-0.7em}

\begin{center}
{\begin{tabular}{ c }
\includegraphics[height=1.0in]{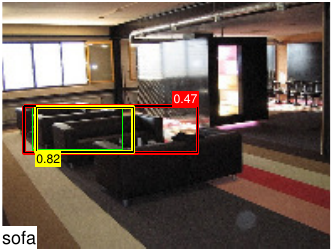}\hspace{0.05in}
\includegraphics[height=1.0in]{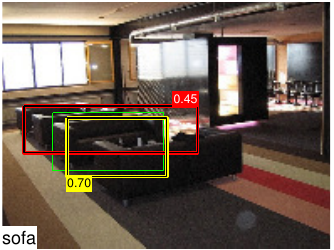}\\
sofa (examples with the largest improvement) \\
\end{tabular}}
\end{center}\vspace{-0.7em}

\begin{center}
{\begin{tabular}{ c }
\includegraphics[height=1.0in]{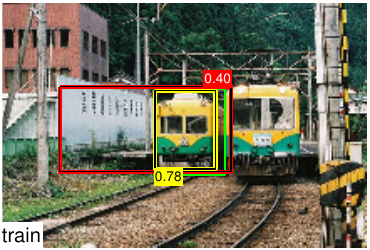}\hspace{0.05in}
\includegraphics[height=1.0in]{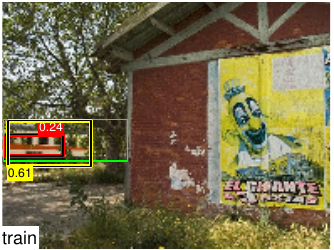}\hspace{0.05in}
\includegraphics[height=1.0in]{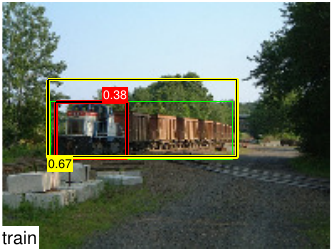}\hspace{0.05in}
\includegraphics[height=1.0in]{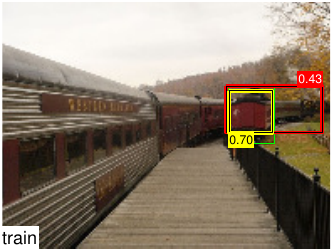}\\ train (examples with the largest improvement) \\
\end{tabular}}
\end{center}\vspace{-0.7em}

\begin{center}
{\begin{tabular}{ c }
\includegraphics[height=1.0in]{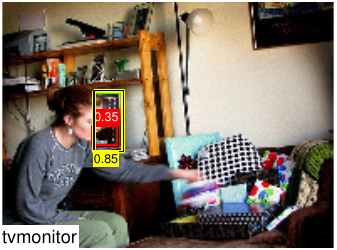}\hspace{0.05in}
\includegraphics[height=1.0in]{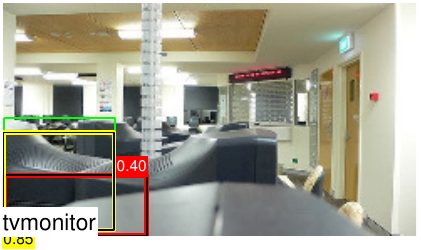}\hspace{0.05in}
\includegraphics[height=1.0in]{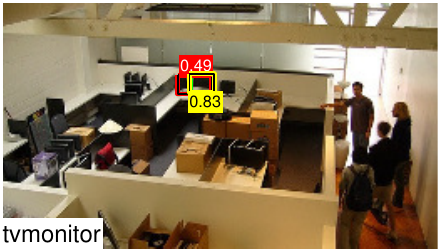}\hspace{0.05in}
\includegraphics[height=1.0in]{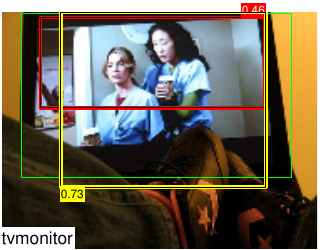}\\ tvmonitor (examples with the largest improvement) \\
\end{tabular}}
\end{center}\vspace{-0.7em}
 
\vspace{-1em}
\begin{figure}[H]
\caption{Examples with the largest improvement with regards to the baseline method on PASCAL VOC 2007 test set. Refer to the text for more detail.}
\label{fig:best-voc2007}
\end{figure}

\clearpage

\noindent \begin{center} {\begin{tabular}{ c }
\includegraphics[height=1.0in]{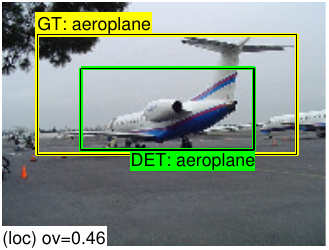} \hspace{0.05in}
\includegraphics[height=1.0in]{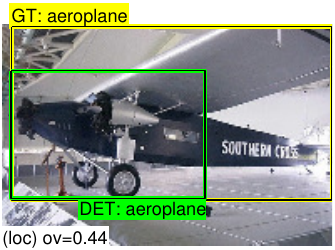} \hspace{0.05in}
\includegraphics[height=1.0in]{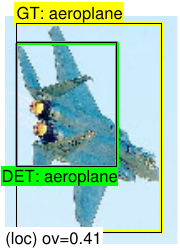} \hspace{0.05in}
\includegraphics[height=1.0in]{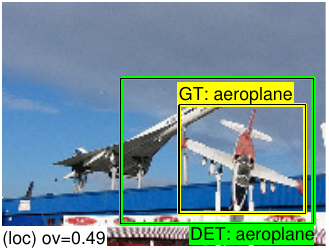} \hspace{0.05in}
\includegraphics[height=1.0in]{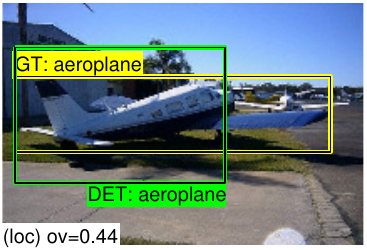} \\
aeroplane (top-ranked false positive) \\
\end{tabular}}\par\end{center}

\vspace{-0.1in}

\noindent \begin{center} {\begin{tabular}{c}
\includegraphics[height=1.0in]{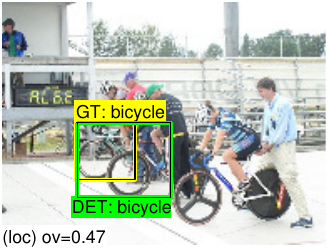} \hspace{0.05in}
\includegraphics[height=1.0in]{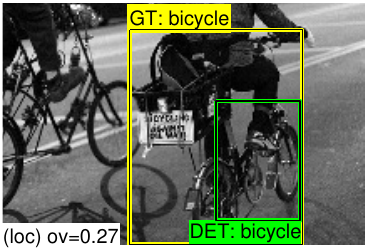} \hspace{0.05in}
\includegraphics[height=1.0in]{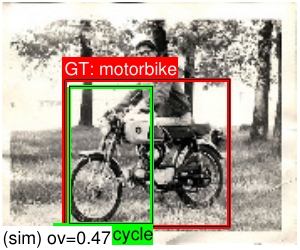} \hspace{0.05in}
\includegraphics[height=1.0in]{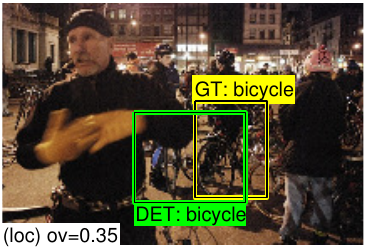} \\
bicycle (top-ranked false positive) \\
\end{tabular}}\par\end{center}

\vspace{-0.1in}

\noindent \begin{center} {\begin{tabular}{c}
\includegraphics[height=1.0in]{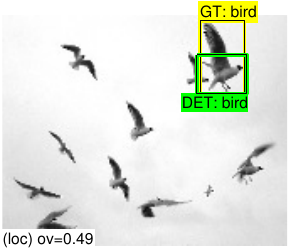} \hspace{0.05in}
\includegraphics[height=1.0in]{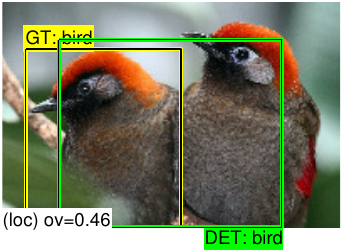} \hspace{0.05in}
\includegraphics[height=1.0in]{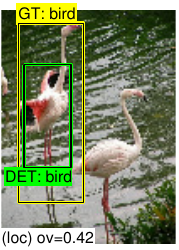} \hspace{0.05in}
\includegraphics[height=1.0in]{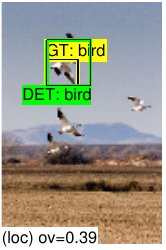} \hspace{0.05in}
\includegraphics[height=1.0in]{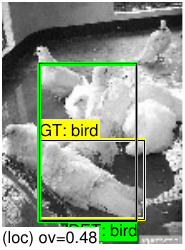} \\
bird (top-ranked false positive) \\
\end{tabular}}\par\end{center}

\vspace{-0.1in}

\noindent \begin{center} {\begin{tabular}{c}
\includegraphics[height=1.0in]{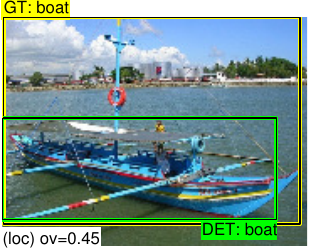} \hspace{0.05in}
\includegraphics[height=1.0in]{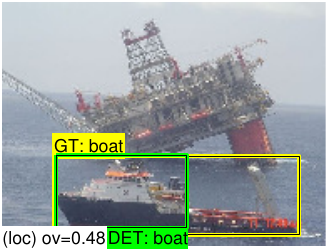} \hspace{0.05in}
\includegraphics[height=1.0in]{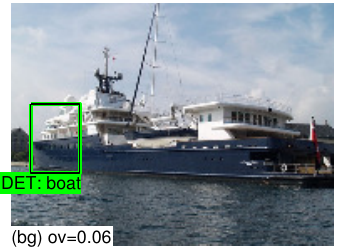} \hspace{0.05in}
\includegraphics[height=1.0in]{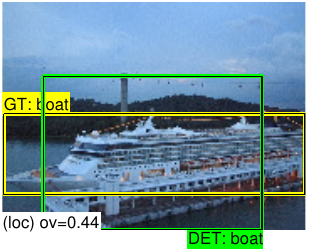} \hspace{0.05in}
\includegraphics[height=1.0in]{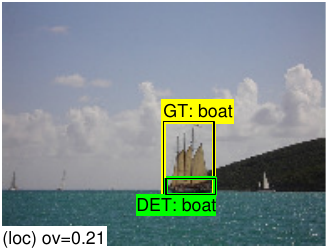} \\
boat (top-ranked false positive) \\
\end{tabular}}\par\end{center}

\vspace{-0.1in}

\noindent \begin{center} {\begin{tabular}{c}
\includegraphics[height=1.0in]{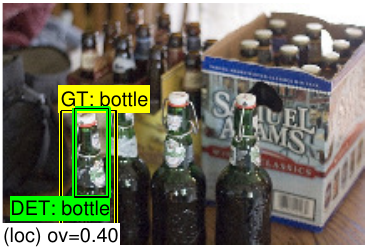} \hspace{0.05in}
\includegraphics[height=1.0in]{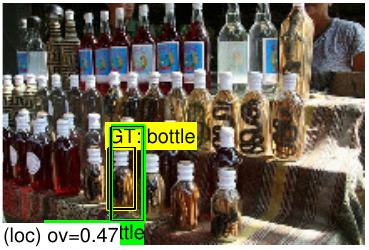} \hspace{0.05in}
\includegraphics[height=1.0in]{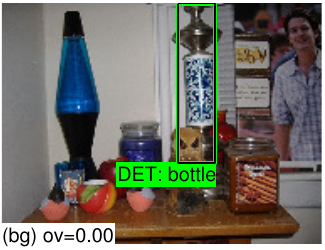} \hspace{0.05in}
\includegraphics[height=1.0in]{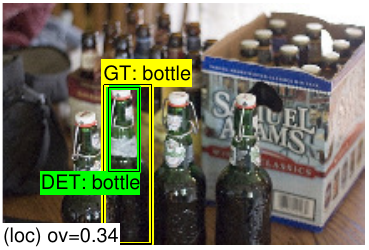} \\ bottle (top-ranked false positive) \\
\end{tabular}}\par\end{center}

\vspace{-0.1in}

\noindent \begin{center} {\begin{tabular}{c}
\includegraphics[height=1.0in]{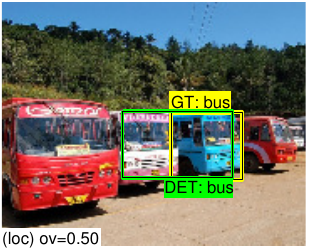} \hspace{0.05in}
\includegraphics[height=1.0in]{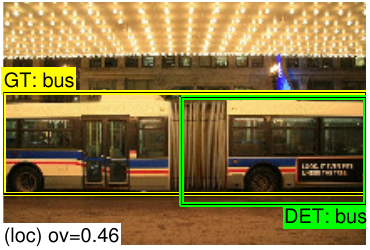} \hspace{0.05in}
\includegraphics[height=1.0in]{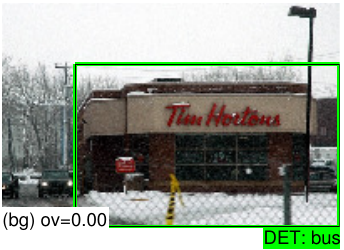} \hspace{0.05in}
\includegraphics[height=1.0in]{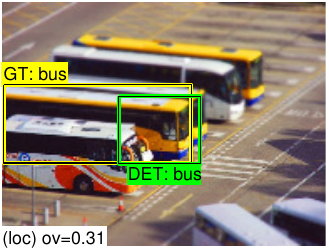} \\
bus (top-ranked false positive) \\
\end{tabular}}\par\end{center}

\vspace{-0.1in}

\noindent \begin{center} {\begin{tabular}{c}
\includegraphics[height=1.0in]{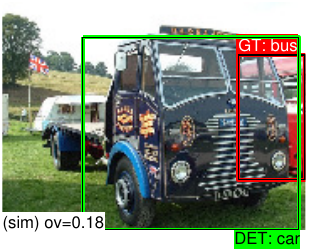} \hspace{0.05in}
\includegraphics[height=1.0in]{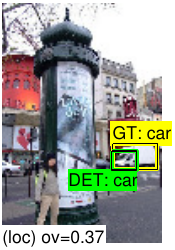} \hspace{0.05in}
\includegraphics[height=1.0in]{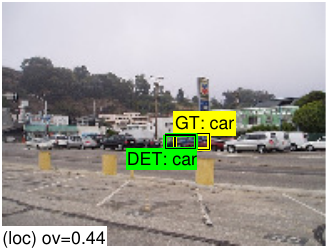} \hspace{0.05in}
\includegraphics[height=1.0in]{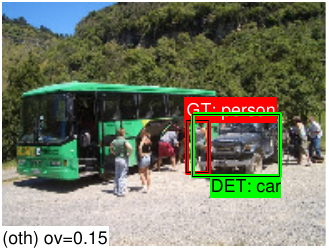} \hspace{0.05in}
\includegraphics[height=1.0in]{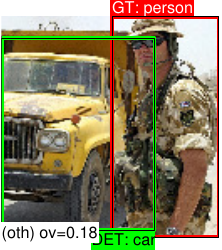} \\
car (top-ranked false positive) \\
\end{tabular}}\par\end{center}

\vspace{-0.1in}

\noindent \begin{center} {\begin{tabular}{c}
\includegraphics[height=1.0in]{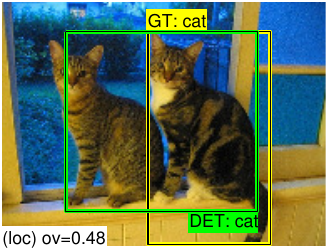} \hspace{0.05in}
\includegraphics[height=1.0in]{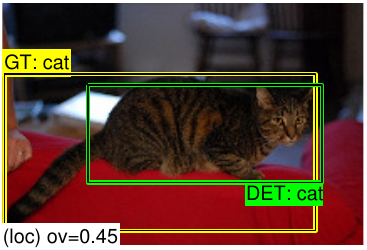} \hspace{0.05in}
\includegraphics[height=1.0in]{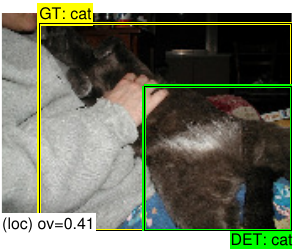} \hspace{0.05in}
\includegraphics[height=1.0in]{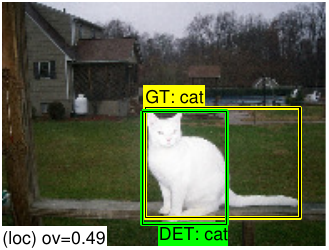} \\ cat (top-ranked false positive) \\
\end{tabular}}\par\end{center}

\vspace{-0.1in}

\noindent \begin{center} {\begin{tabular}{c}
\includegraphics[height=1.0in]{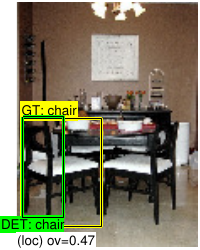} \hspace{0.05in}
\includegraphics[height=1.0in]{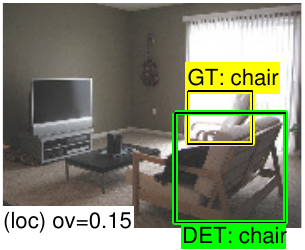} \hspace{0.05in}
\includegraphics[height=1.0in]{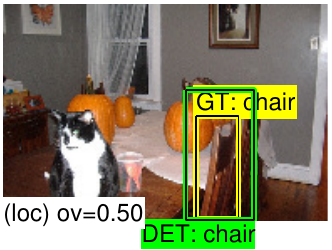} \hspace{0.05in}
\includegraphics[height=1.0in]{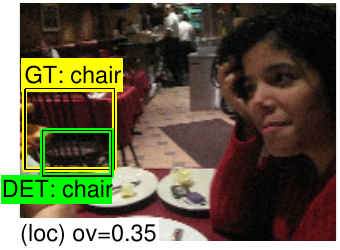} \hspace{0.05in}
\includegraphics[height=1.0in]{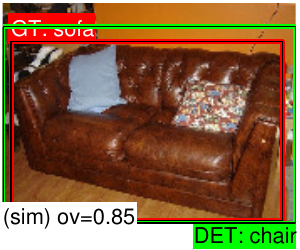} \\
chair (top-ranked false positive) \\
\end{tabular}}\par\end{center}

\vspace{-0.1in}

\noindent \begin{center} {\begin{tabular}{c}
\includegraphics[height=1.0in]{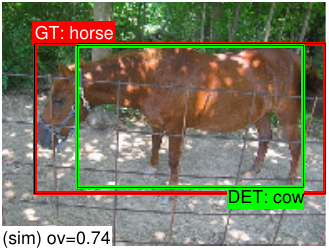} \hspace{0.05in}
\includegraphics[height=1.0in]{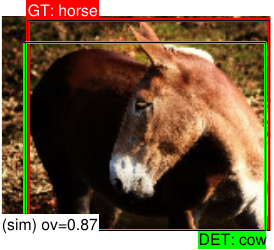} \hspace{0.05in}
\includegraphics[height=1.0in]{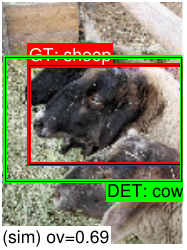} \hspace{0.05in}
\includegraphics[height=1.0in]{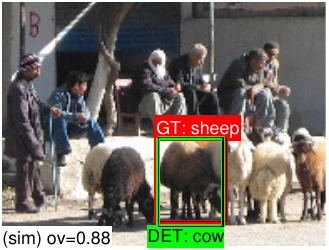} \hspace{0.05in}
\includegraphics[height=1.0in]{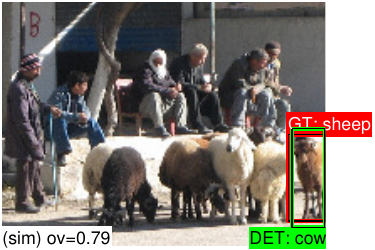} \\
cow (top-ranked false positive) \\
\end{tabular}}\par\end{center}

\vspace{-0.1in}

\noindent \begin{center} {\begin{tabular}{c}
\includegraphics[height=1.0in]{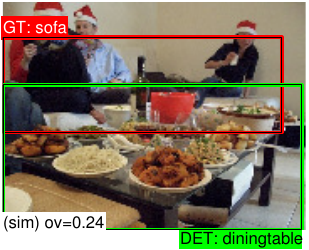} \hspace{0.05in}
\includegraphics[height=1.0in]{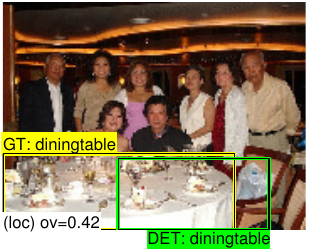} \hspace{0.05in}
\includegraphics[height=1.0in]{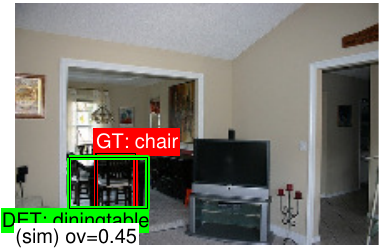} \hspace{0.05in}
\includegraphics[height=1.0in]{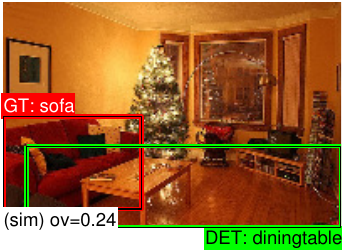} \\ diningtable (top-ranked false positive) \\
\end{tabular}}\par\end{center}

\vspace{-0.1in}

\noindent \begin{center} {\begin{tabular}{c}
\includegraphics[height=1.0in]{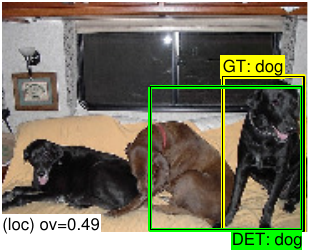} \hspace{0.05in}
\includegraphics[height=1.0in]{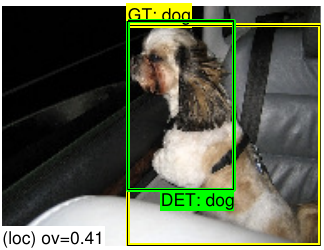} \hspace{0.05in}
\includegraphics[height=1.0in]{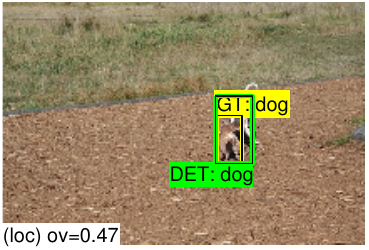} \hspace{0.05in}
\includegraphics[height=1.0in]{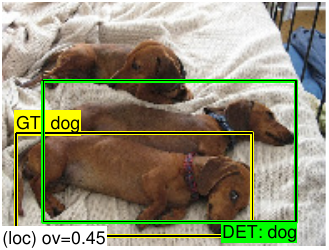} \\ dog (top-ranked false positive) \\
\end{tabular}}\par\end{center}

\vspace{-0.1in}

\noindent \begin{center} {\begin{tabular}{c}
\includegraphics[height=1.0in]{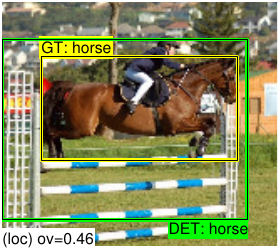} \hspace{0.05in}
\includegraphics[height=1.0in]{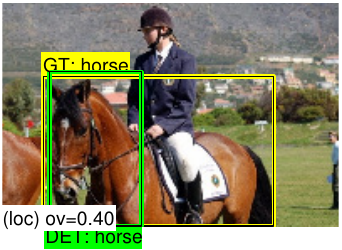} \hspace{0.05in}
\includegraphics[height=1.0in]{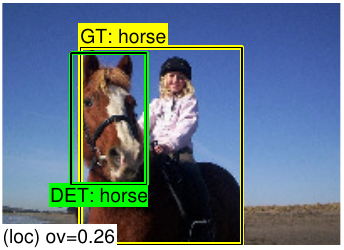} \hspace{0.05in}
\includegraphics[height=1.0in]{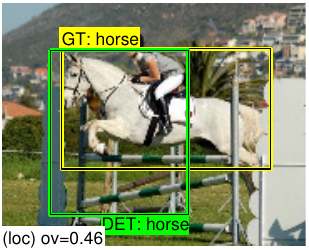} \hspace{0.05in}
\includegraphics[height=1.0in]{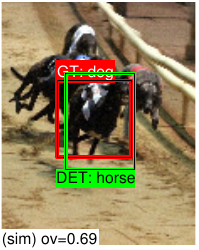} \\
horse (top-ranked false positive) \\
\end{tabular}}\par\end{center}

\vspace{-0.1in}

\noindent \begin{center} {\begin{tabular}{c}
\includegraphics[height=1.0in]{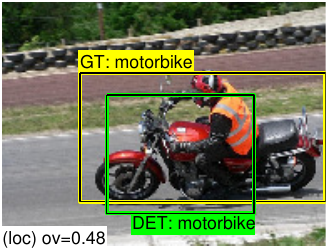} \hspace{0.05in}
\includegraphics[height=1.0in]{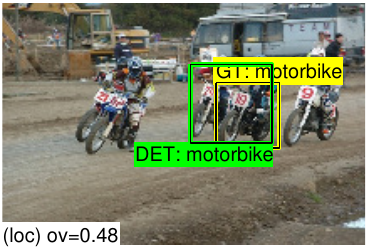} \hspace{0.05in}
\includegraphics[height=1.0in]{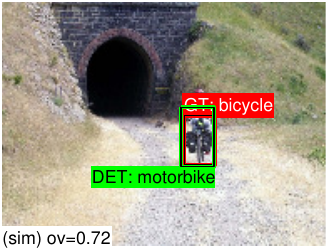} \hspace{0.05in}
\includegraphics[height=1.0in]{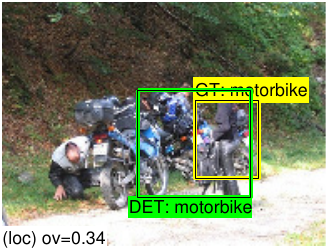} \\ motorbike (top-ranked false positive) \\
\end{tabular}}\par\end{center}

\vspace{-0.1in}

\noindent \begin{center} {\begin{tabular}{c}
\includegraphics[height=1.0in]{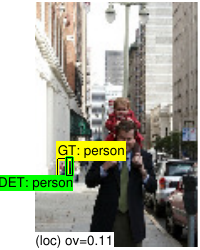} \hspace{0.05in}
\includegraphics[height=1.0in]{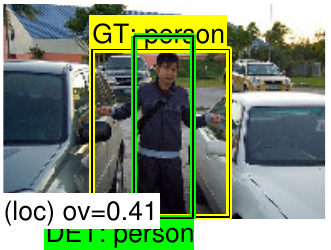} \hspace{0.05in}
\includegraphics[height=1.0in]{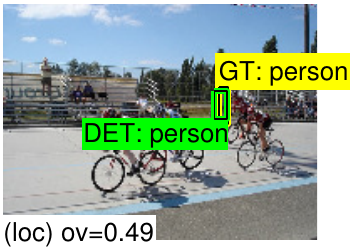} \hspace{0.05in}
\includegraphics[height=1.0in]{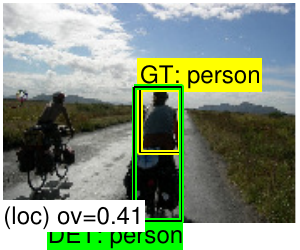} \hspace{0.05in}
\includegraphics[height=1.0in]{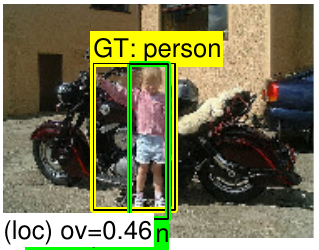} \\
person (top-ranked false positive) \\
\end{tabular}}\par\end{center}

\vspace{-0.1in}

\noindent \begin{center} {\begin{tabular}{c}
\includegraphics[height=1.0in]{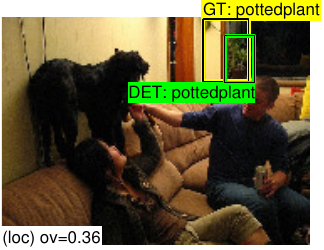} \hspace{0.05in}
\includegraphics[height=1.0in]{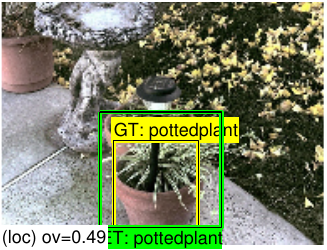} \hspace{0.05in}
\includegraphics[height=1.0in]{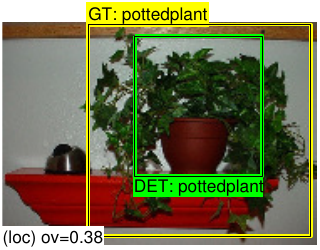} \hspace{0.05in}
\includegraphics[height=1.0in]{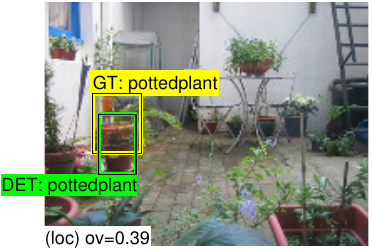} \hspace{0.05in}
\includegraphics[height=1.0in]{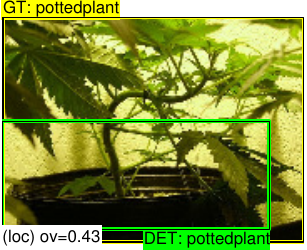} \\
pottedplant (top-ranked false positive) \\
\end{tabular}}\par\end{center}

\vspace{-0.1in}

\noindent \begin{center} {\begin{tabular}{c}
\includegraphics[height=1.0in]{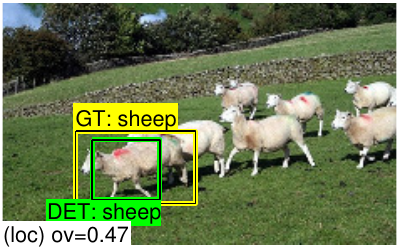} \hspace{0.05in}
\includegraphics[height=1.0in]{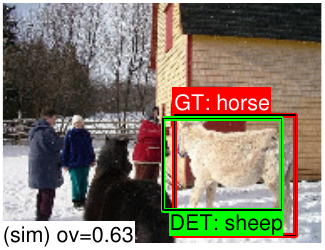} \hspace{0.05in}
\includegraphics[height=1.0in]{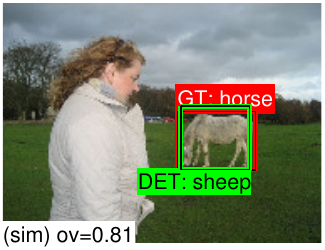} \hspace{0.05in}
\includegraphics[height=1.0in]{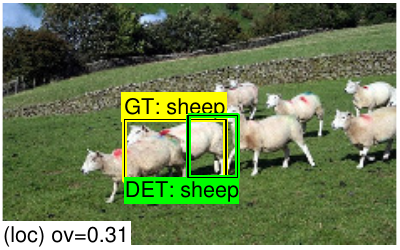} \\ sheep (top-ranked false positive) \\
\end{tabular}}\par\end{center}

\vspace{-0.1in}

\noindent \begin{center} {\begin{tabular}{c}
\includegraphics[height=1.0in]{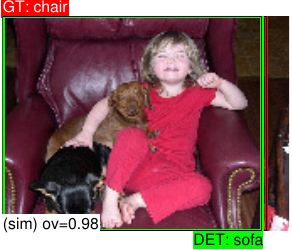} \hspace{0.05in}
\includegraphics[height=1.0in]{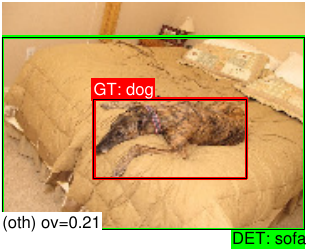} \hspace{0.05in}
\includegraphics[height=1.0in]{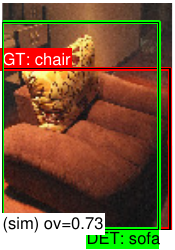} \hspace{0.05in}
\includegraphics[height=1.0in]{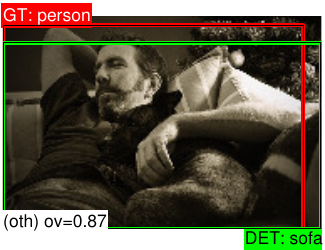} \hspace{0.05in}
\includegraphics[height=1.0in]{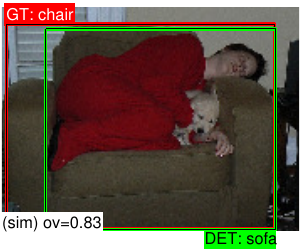} \\
sofa (top-ranked false positive) \\
\end{tabular}}\par\end{center}

\vspace{-0.1in}

\noindent \begin{center} {\begin{tabular}{c}
\includegraphics[height=1.0in]{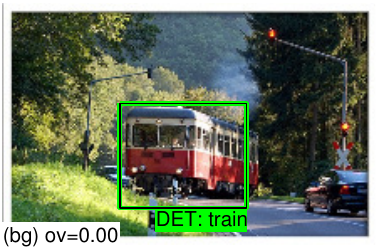} \hspace{0.05in}
\includegraphics[height=1.0in]{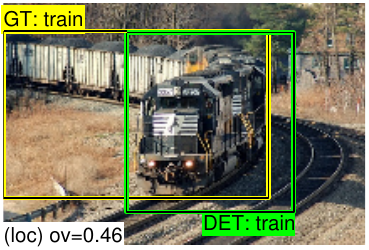} \hspace{0.05in}
\includegraphics[height=1.0in]{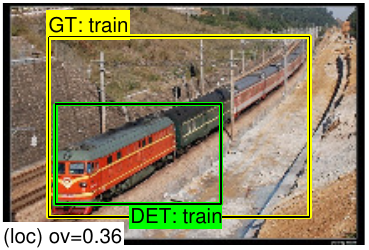} \hspace{0.05in}
\includegraphics[height=1.0in]{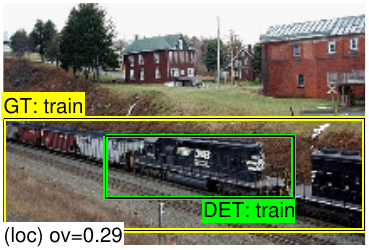} \\
train (top-ranked false positive) \\
\end{tabular}}\par\end{center}

\vspace{-0.1in}

\noindent \begin{center} {\begin{tabular}{c}
\includegraphics[height=1.0in]{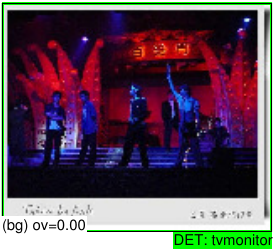} \hspace{0.05in}
\includegraphics[height=1.0in]{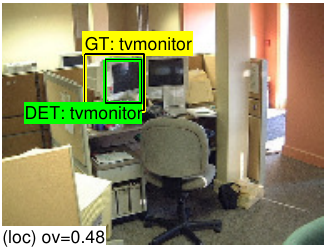} \hspace{0.05in}
\includegraphics[height=1.0in]{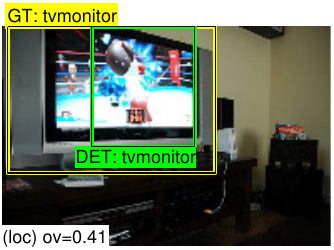} \hspace{0.05in}
\includegraphics[height=1.0in]{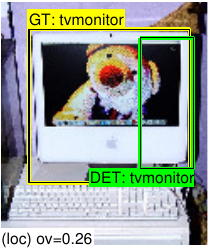} \hspace{0.05in}
\includegraphics[height=1.0in]{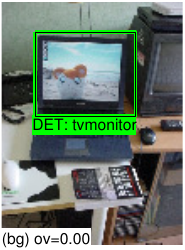} \\
tvmonitor (top-ranked false positive) \\
\end{tabular}}\par\end{center}

\vspace{-1em}
\begin{figure}[H]
\caption{Top-ranked false positives of ``VGGNet + StructObj + FGS'' on PASCAL VOC 2007 test set.}
\label{fig:fp-voc2007}
\end{figure}

\clearpage

\begin{centering}\includegraphics[height=1.2in]{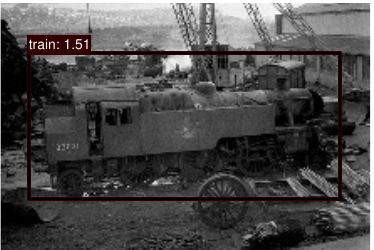}
\includegraphics[height=1.2in]{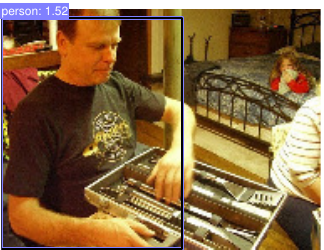}
\includegraphics[height=1.2in]{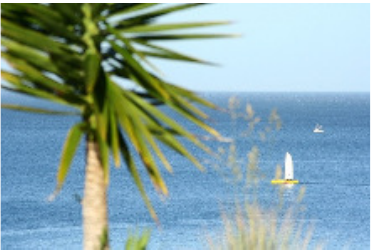}
\includegraphics[height=1.2in]{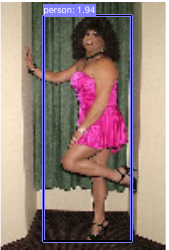}
\includegraphics[height=1.2in]{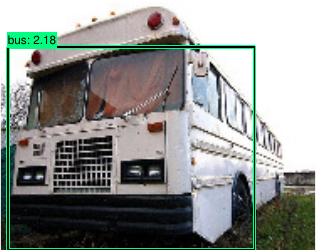}
\includegraphics[height=1.2in]{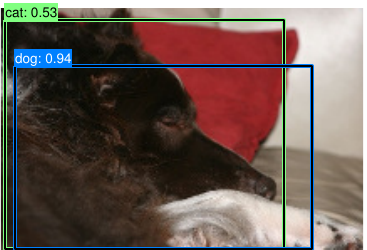}
\includegraphics[height=1.2in]{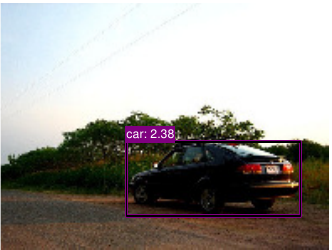}
\includegraphics[height=1.2in]{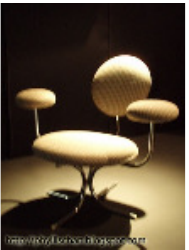}
\includegraphics[height=1.2in]{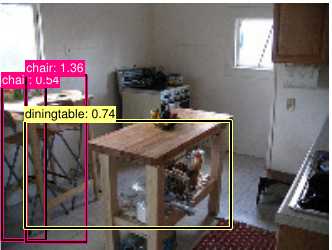}
\includegraphics[height=1.2in]{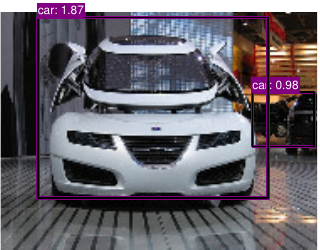}
\includegraphics[height=1.2in]{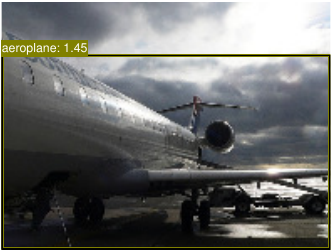}
\includegraphics[height=1.2in]{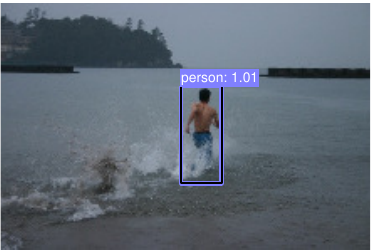}
\includegraphics[height=1.2in]{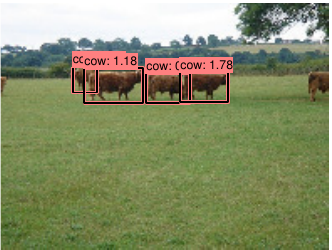}
\includegraphics[height=1.2in]{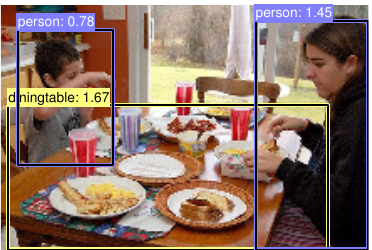}
\includegraphics[height=1.2in]{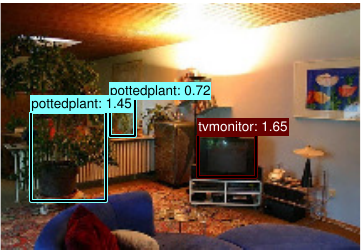}
\includegraphics[height=1.2in]{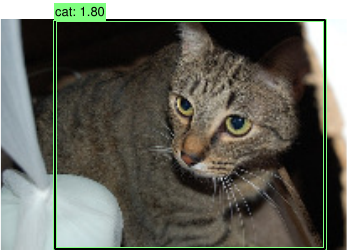}
\includegraphics[height=1.2in]{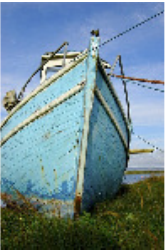}
\includegraphics[height=1.2in]{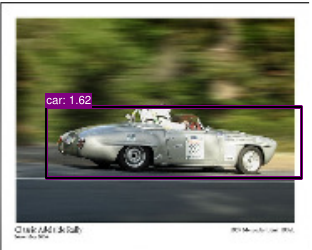}
\includegraphics[height=1.2in]{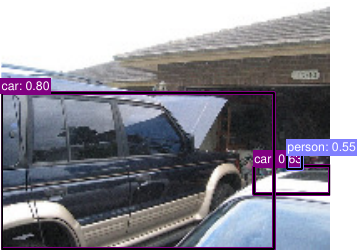}
\includegraphics[height=1.2in]{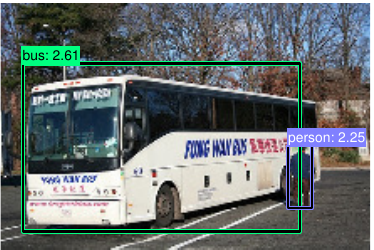}
\includegraphics[height=1.2in]{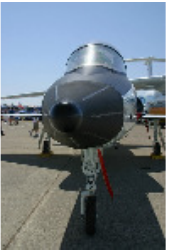}
\includegraphics[height=1.2in]{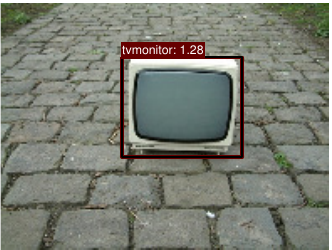}
\includegraphics[height=1.2in]{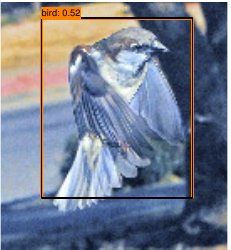}
\includegraphics[height=1.2in]{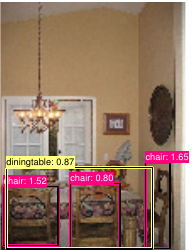}
\includegraphics[height=1.2in]{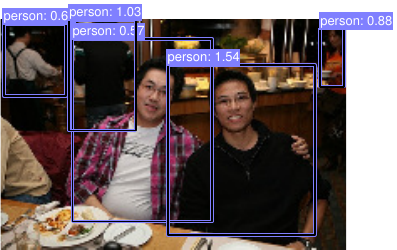}
\includegraphics[height=1.2in]{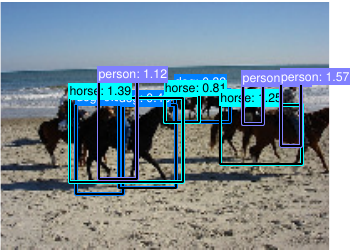}
\includegraphics[height=1.2in]{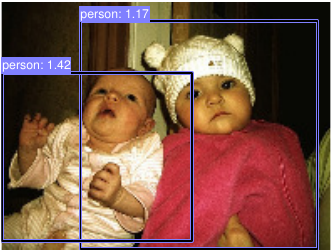}
\includegraphics[height=1.2in]{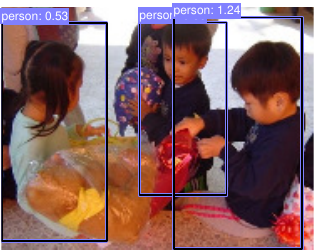}
\includegraphics[height=1.2in]{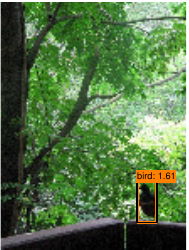}
\includegraphics[height=1.2in]{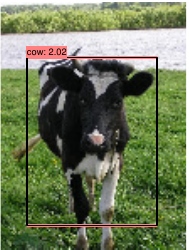}
\includegraphics[height=1.2in]{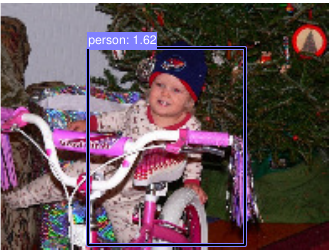}
\includegraphics[height=1.2in]{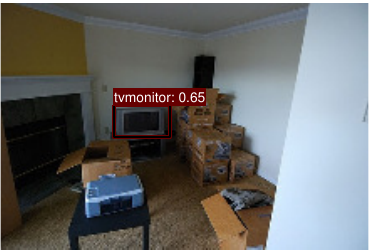}
\includegraphics[height=1.2in]{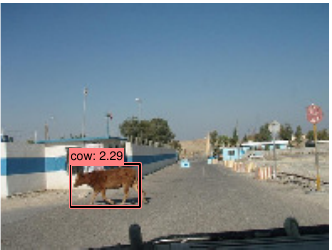}
\includegraphics[height=1.2in]{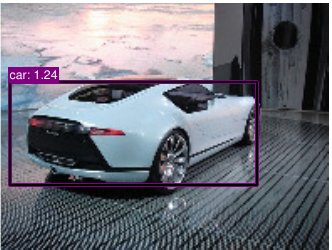}
\includegraphics[height=1.2in]{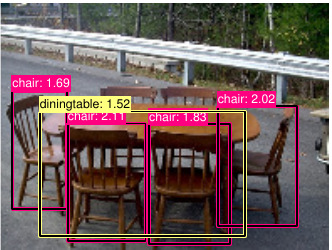}
\includegraphics[height=1.2in]{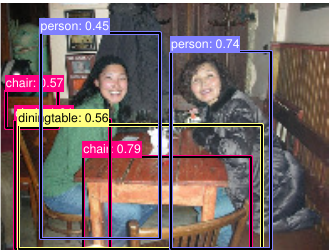}
\includegraphics[height=1.2in]{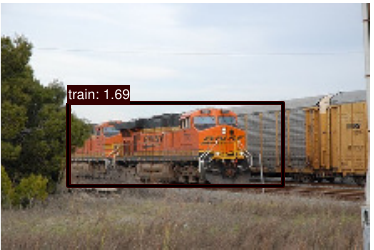}
\includegraphics[height=1.2in]{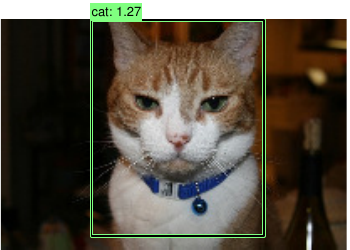}
\includegraphics[height=1.2in]{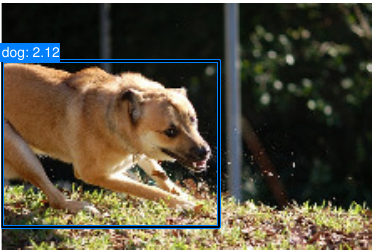}
\includegraphics[height=1.2in]{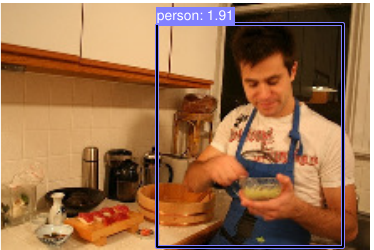}
\includegraphics[height=1.2in]{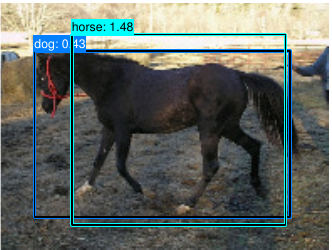}
\includegraphics[height=1.2in]{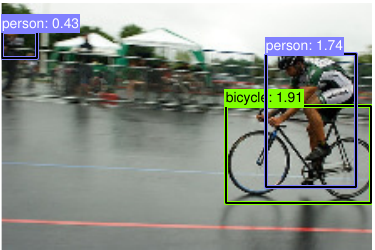}
\includegraphics[height=1.2in]{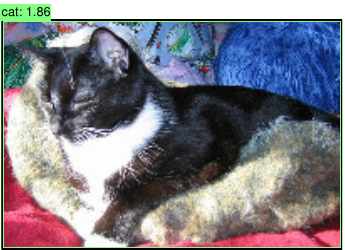}
\includegraphics[height=1.2in]{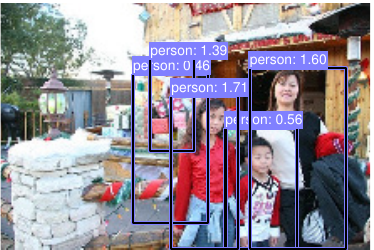}
\includegraphics[height=1.2in]{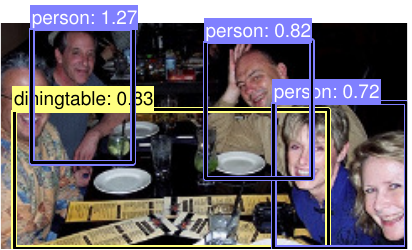}
\includegraphics[height=1.2in]{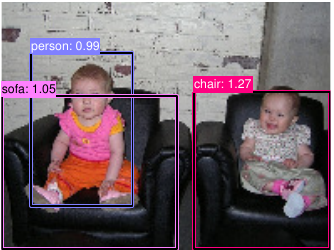}
\includegraphics[height=1.2in]{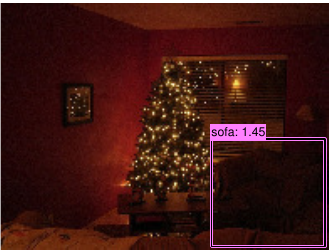}
\includegraphics[height=1.2in]{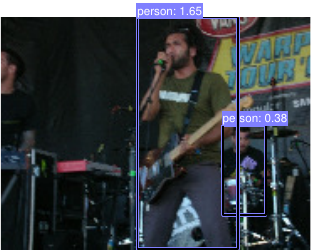}
\includegraphics[height=1.2in]{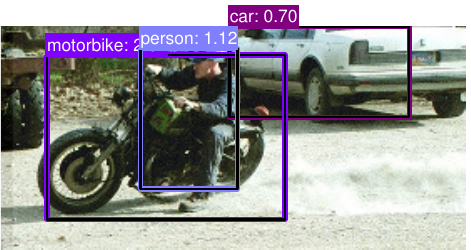}
\includegraphics[height=1.2in]{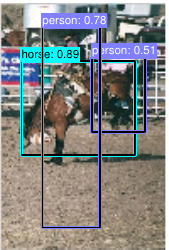}
\includegraphics[height=1.2in]{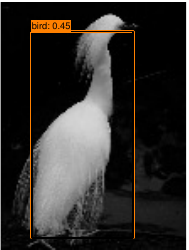}
\includegraphics[height=1.2in]{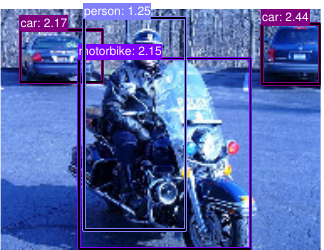}
\includegraphics[height=1.2in]{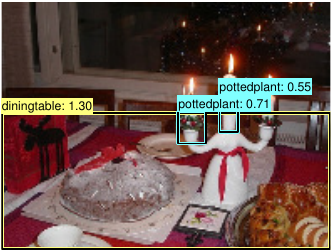}
\includegraphics[height=1.2in]{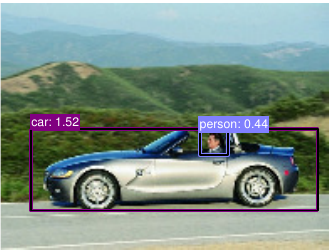}
\includegraphics[height=1.2in]{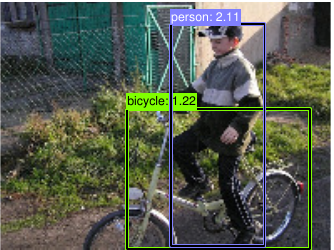}
\includegraphics[height=1.2in]{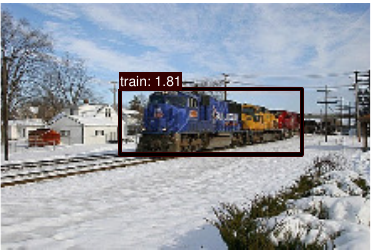}
\includegraphics[height=1.2in]{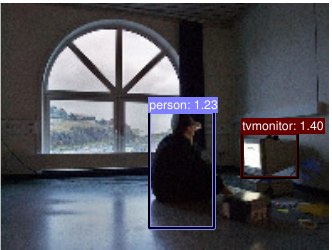}
\includegraphics[height=1.2in]{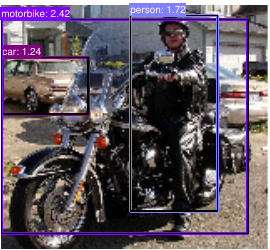}
\includegraphics[height=1.2in]{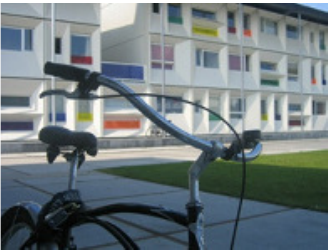}
\includegraphics[height=1.2in]{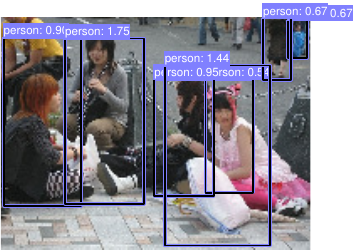}
\includegraphics[height=1.2in]{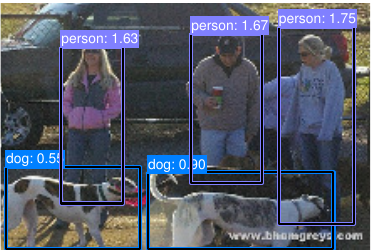}
\includegraphics[height=1.2in]{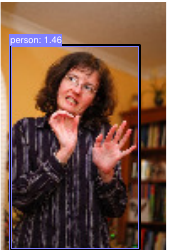}
\includegraphics[height=1.2in]{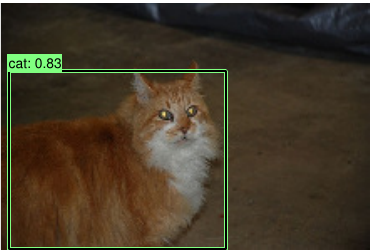}
\includegraphics[height=1.2in]{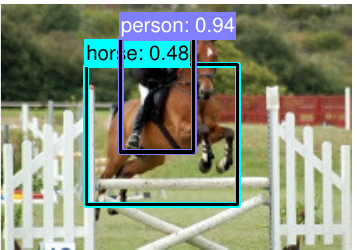}
\includegraphics[height=1.2in]{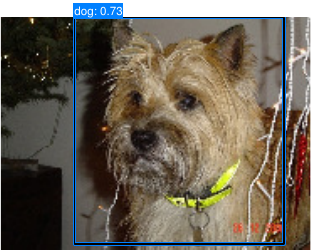}
\includegraphics[height=1.2in]{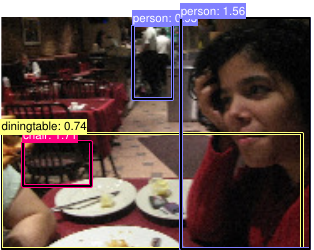}
\includegraphics[height=1.2in]{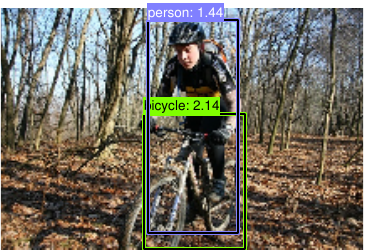}
\includegraphics[height=1.2in]{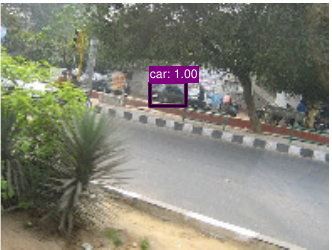}
\includegraphics[height=1.2in]{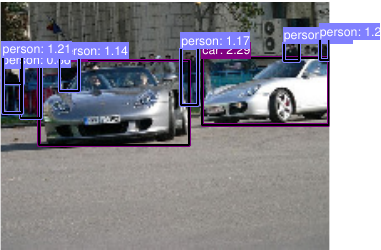}
\includegraphics[height=1.2in]{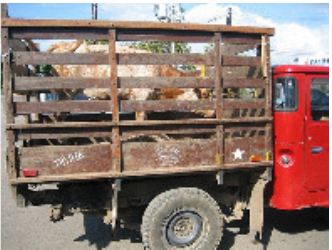}
\includegraphics[height=1.2in]{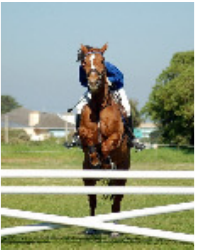}
\includegraphics[height=1.2in]{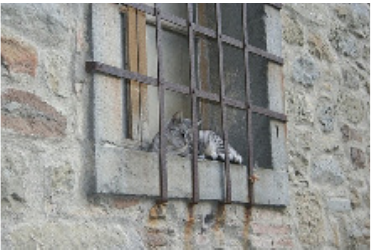}
\includegraphics[height=1.2in]{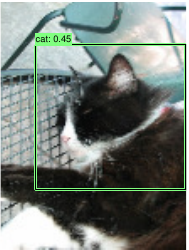}
\includegraphics[height=1.2in]{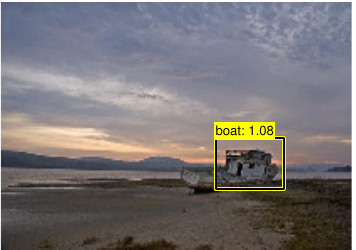}
\includegraphics[height=1.2in]{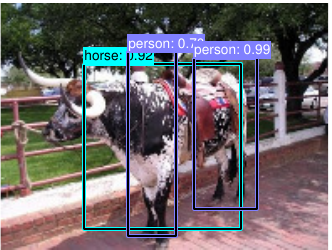}
\includegraphics[height=1.2in]{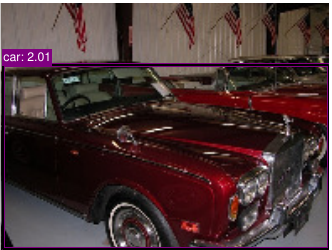}
\includegraphics[height=1.2in]{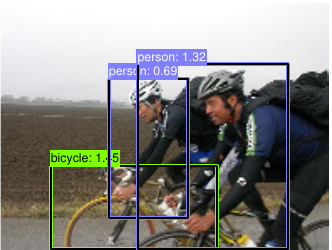}
\includegraphics[height=1.2in]{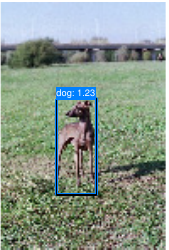}
\includegraphics[height=1.2in]{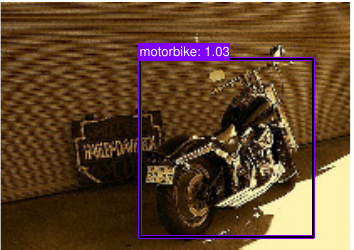}
\includegraphics[height=1.2in]{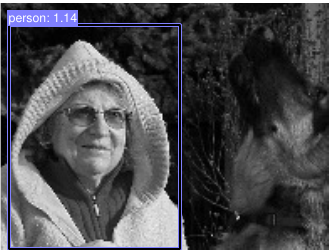}
\includegraphics[height=1.2in]{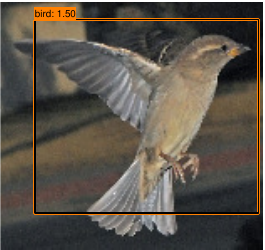}
\includegraphics[height=1.2in]{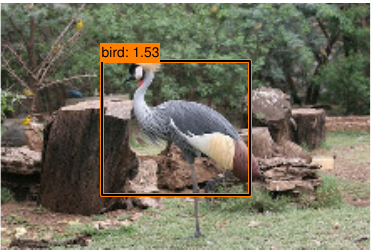}
\includegraphics[height=1.2in]{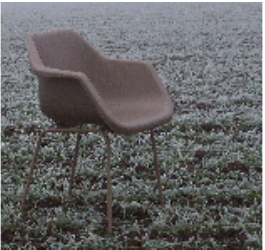}
\includegraphics[height=1.2in]{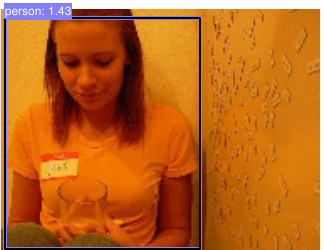}
\includegraphics[height=1.2in]{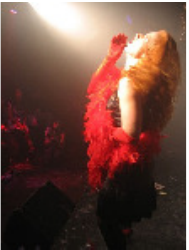}
\includegraphics[height=1.2in]{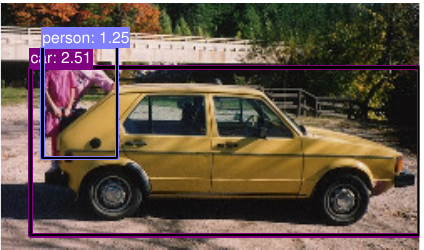}
\includegraphics[height=1.2in]{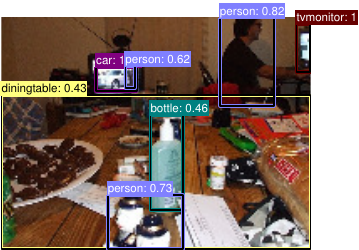}
\includegraphics[height=1.2in]{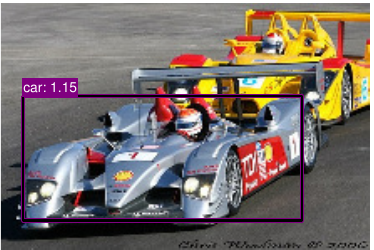}
\includegraphics[height=1.2in]{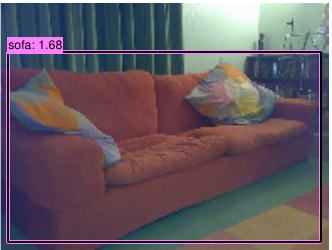}
\includegraphics[height=1.2in]{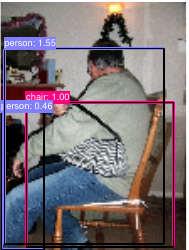}
\includegraphics[height=1.2in]{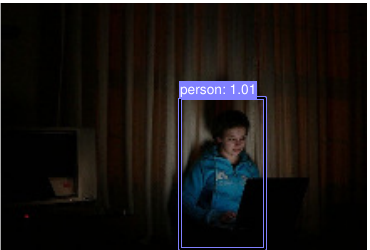}
\includegraphics[height=1.2in]{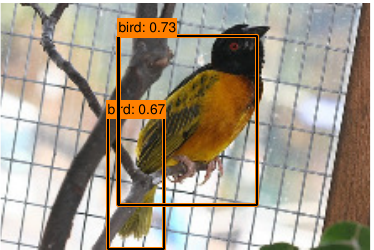}
\includegraphics[height=1.2in]{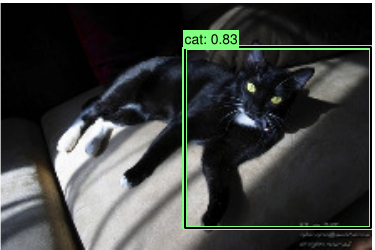}
\includegraphics[height=1.2in]{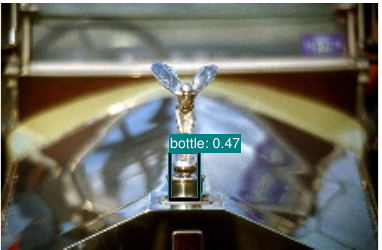}
\includegraphics[height=1.2in]{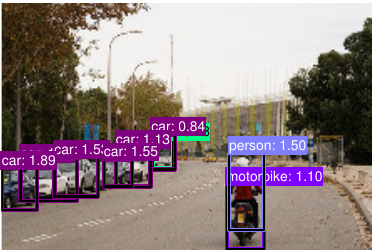}
\includegraphics[height=1.2in]{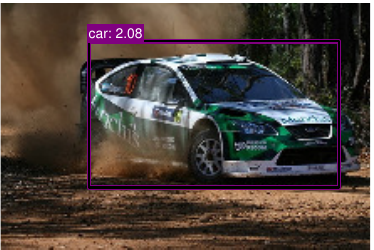}
\includegraphics[height=1.2in]{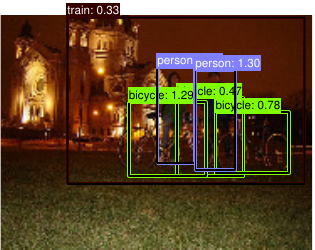}
\includegraphics[height=1.2in]{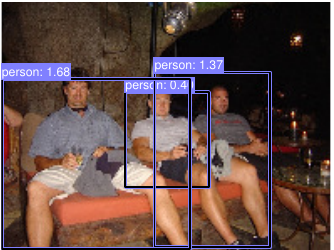}
\includegraphics[height=1.2in]{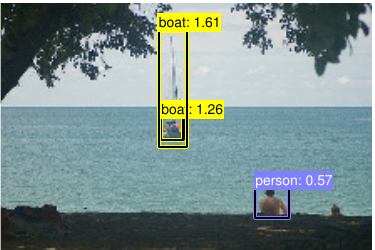}

\end{centering}
 
\vspace{-1em}
\begin{figure}[H]
\caption{Random detection examples of ``VGGNet + StructObj + FGS'' on PASCAL VOC 2007 test set.}
\label{fig:random-voc2007}
\end{figure}

  \end{document}